\documentclass[twoside,11pt]{article}
\usepackage{jair, theapa, rawfonts}

\usepackage{subfig}

\usepackage[utf8]{inputenc} 
\usepackage[T1]{fontenc}   
\usepackage{url}            
\usepackage{booktabs}       
\usepackage{amsmath}       
\usepackage{amsfonts}       
\usepackage{amssymb}
\usepackage{nicefrac}       
\usepackage{microtype}      
\usepackage[dvipsnames]{xcolor}
\usepackage{graphicx}
\usepackage{scalefnt}
\usepackage{rotating}
\usepackage{svg}
\usepackage{soul}
\usepackage{wrapfig}
\sethlcolor{lightgray}
\usepackage{framed}

\usepackage{listings}
\jairheading{1}{1993}{1-15}{6/91}{9/91}
\ShortHeadings{The SocialAI School}{Kova\v c, Portelas, Dominey, \& Oudeyer}
\firstpageno{1}

\newcommand{\socialai}{\textit{SocialAI}\xspace}

\newcommand{\mytextsc}[1]{{\small\textsc{#1}}}

\usepackage{adjustbox}
\usepackage{array}
\usepackage{xspace}

\usepackage{enumitem}
\setitemize{noitemsep,topsep=2pt,parsep=0pt,partopsep=0pt}

\newcolumntype{R}[2]{%
    >{\adjustbox{angle=#1,lap=\width-(#2)}\bgroup}%
    l%
    <{\egroup}%
}
\newcommand*\rot{\multicolumn{1}{R{45}{1em}}}

\begin{document}

\title{The SocialAI School: Insights from Developmental Psychology Towards  Artificial Socio-Cultural Agents}

\author{\name Grgur Kova\v c \email grgur.kovac@inria.fr \\
    \addr Flowers Team, Inria (FR) \\
    \AND
    \name Rémy Portelas \email remy.portelas@ubisoft.com \\
    \addr Ubisoft La Forge (FR) \\
    \addr Flowers Team, Inria (FR) \\
    \AND
    \name Peter Ford Dominey \email peter-ford.dominey@u-bourgogne.fr \\
    \addr INSERM UMR1093-CAPS, Université Bourgogne (FR) \\
    \addr Robot Cognition Laboratory, Institute Marey (FR) \\
    \AND
    \name Pierre-Yves Oudeyer \email pierre-yves.oudeyer@inria.fr \\
    \addr Flowers Team, Inria (FR) \\
}



\maketitle

\begin{abstract}

Developmental psychologists have long-established socio-cognitive abilities as fundamental to human intelligence and development. 
These abilities enable individuals to enter, learn from, and contribute to a surrounding culture. This drives the process of cumulative cultural evolution, which is responsible for humanity's most remarkable achievements.
AI research on social interactive agents mostly concerns the \textit{emergence} of culture in a multi-agent setting (often without a strong grounding in developmental psychology).
We argue that AI research should be informed by psychology and study socio-cognitive abilities enabling to \textit{enter} a culture as well.
We draw inspiration from the work of Michael Tomasello and Jerome Bruner, who studied socio-cognitive development and emphasized the influence of a cultural environment on intelligence.
We outline a broader set of concepts than those currently studied in AI to provide a foundation for research in artificial social intelligence.
Those concepts include social cognition (joint attention, perspective taking), communication, social learning, formats, and scaffolding.
To facilitate research in this domain, we present The SocialAI school - a tool that offers a customizable parameterized suite of procedurally generated environments.
This tool simplifies experimentation with the introduced concepts.
Additionally, these environments can be used both with multimodal RL agents, or with pure-text Large Language Models (LLMs) as interactive agents.
Through a series of case studies, we demonstrate the versatility of the SocialAI school for studying both RL and LLM-based agents.
Our motivation is to engage the AI community around social intelligence informed by developmental psychology, and to provide a user-friendly resource and tool for initial investigations in this direction.
Refer to the project website for code and additional information: \small{\url{https://sites.google.com/view/socialai-school}}.
\end{abstract}

\section{Introduction}
\label{sec:intro}


Our everyday life is immersed in a sociocultural world, which we navigate using a set of sophisticated socio-cognitive abilities.
Although at first it might seem that this sociocultural world is just another downstream product of our cognition, decades of research in developmental psychology suggest the opposite.
Our socio-cultural world, cultural knowledge, and our socio-cognitive abilities are the foundation of our development and both our social and asocial intelligence \shortcite{vygotsky_1978,bruner1990acts,tomasello2019becoming}. 

For Vygotsky, a main driver for “higher-level” cognition are socio-cultural interactions \shortcite{vygotsky_1978}. For him, many high-level cognitive functions first appear at the social level and \textit{then} develop at the individual level. This leap from interpersonal processes to intrapersonal processes is referred to as \textit{internalization}. A typical example of this process is learning to count. Children first learn to count out loud, i.e. with language and social guidance, which is an interpersonal process. As the child improves, it will learn to count in its head, no longer requiring any external guidance: counting became internalized, and will be a first step towards other more complex forms of abstract thinking. Vygotsky's theories influenced multiple works within cognitive science \shortcite{clark-being-there,hutchins96a}, primatology \shortcite{tomasello_cultural_origins_1999} and the developmental robotics branch of AI \shortcite{BILLARD98,Brooks2002,cangelosi2010roadmap,MIROLLI2011298}.

Another pillar of modern developmental psychology is Jerome Bruner. 
He, too, emphasized the importance of culture in human development. 
Bruner writes: “\textit{it is culture, not biology, that shapes human life and the human mind, that gives meaning to action by situating its underlying intentional states in an interpretative system}” \shortcite{bruner1990acts}.
Most importantly for this paper, he presents a pragmatic view studying how referencing, requesting and finally language develop through routinized social interactions (formats) in which those abilities are \textit{necessary} to achieve various ends.
He describes these interactions as scaffolded -  the caretaker gradually helps less and demands more of the child to achieve those goals, and this bootstraps the child's development \shortciteA{bruner85childstalk}.

Finally, Michael Tomasello's work \shortcite{tomasello_cultural_origins_1999,tomasello2019becoming,Tomasello-role-of-roles} constitutes a representative and contemporary assessment of the nature and central importance of sociality in human cognition.
Through decades of theoretical and experimental studies with both humans and primates, Tomasello outlined core social abilities and motivations.
When combined with the relevant experience, they enable us to enter, benefit from, and contribute to the human culture.
This cumulative cultural evolution is a powerful form of cultural transmission enabling the development and perpetuation of our complex culture and knowledge, and it is made possible by those socio-cognitive abilities \shortciteA{tomasello_cultural_origins_1999}.

Given the key role social cognition plays in human cognition and cultural evolution, it is natural that the field of AI aims to model our social intelligence.
A socially competent AI could learn our culture and participate in its cultural evolution, i.e. improve our concepts, theories, inventions, and create new ones.
A system capable of out-of-the-box thinking creative solutions and discovering new relevant problems must learn our values and how we see and understand the world (it must learn our culture).
We do not claim that The SocialAI is sufficient to reach that far and complex goal.
We only propose that being informed by the concepts discussed in this paper is useful, and we present SocialAI as a tool which could be used to start investigating such questions in more details.

\begin{wrapfigure}{R}{0.4\textwidth}
\centering
\includegraphics[width=0.4\textwidth]{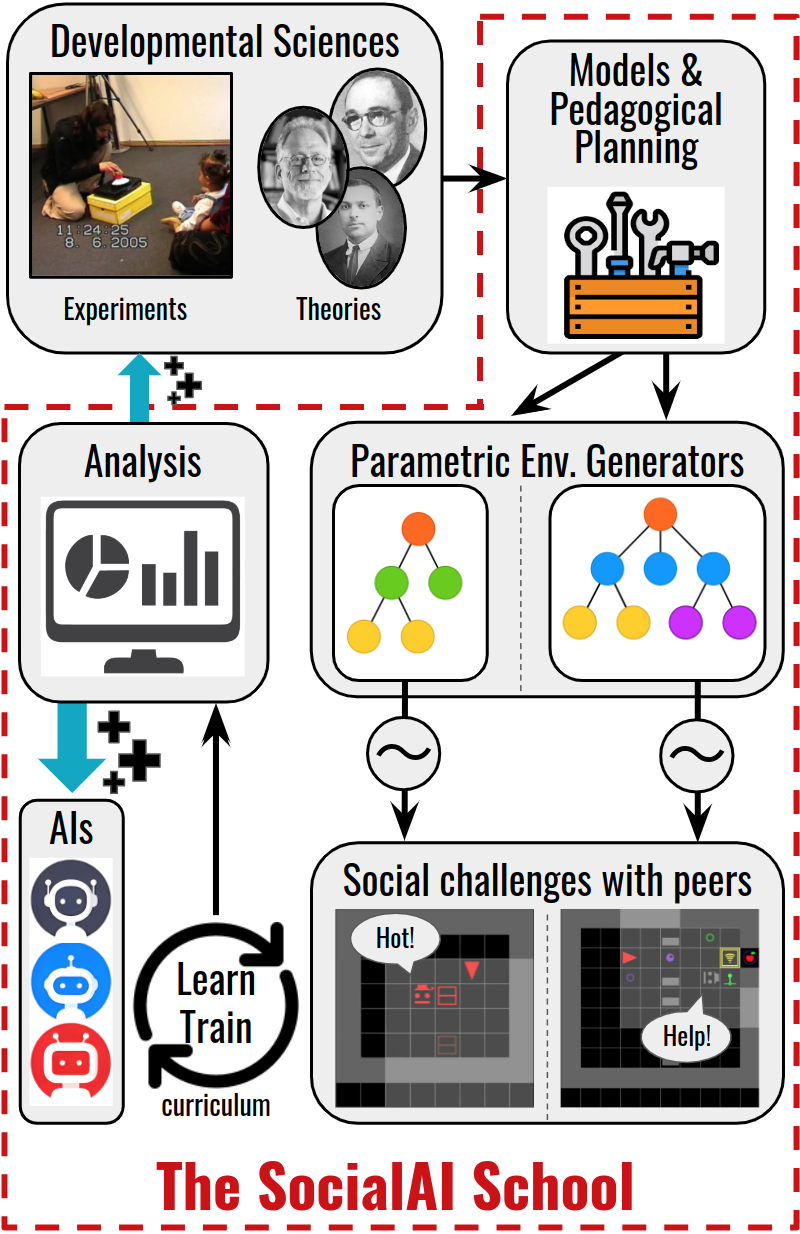}
\caption{\footnotesize The SocialAI School provides technical and conceptual tools aiming to simplify research seeking to design socially proficient artificial agents.}
\label{fig:main_vizu}
\end{wrapfigure}

Enriching AI with those skills also has numerous practical implications. 
Socially competent robots, capable of social learning, would be much easier to deploy and adapt to novel tasks and tools.
For example, performing collaborative tasks with a robotic learner able to detect, learn and reuse context-dependent sets of communicative gestures/utterances could be easily integrated into human teams, without requiring humans to adopt new conventions.
Furthermore, robots capable of learning human values and moral norms will be capable of performing tasks in the constraints defined by those values.

AI research on interactive agents is often focused on navigation and object manipulation problems, excised of any social dimension \shortcite{dqn,ddpg}.
The study of sociality is mostly studied in Multi-Agent settings, where the main focus is often on the \textit{emergence} of culture (often with only a weak grounding in developmental psychology) \shortcite{jacques2019-socialinfl,baker2019emergent}.
While we believe that those directions are both interesting and important, in this work we focus on \textit{entering} an already existing complex culture.
And we argue that it can be beneficial to be informed by developmental psychology theories.

In the rapidly emerging field of Large Language models, social cognition research consists of proof-of-concept simulations \shortcite{simulacra} and systematic benchmarks. 
Two most notable benchmarks are SiQA \shortcite{socialiqa}, which evaluates social common sense reasoning (without grounding in psychology), and ToMi \shortcite{tomi} which presents false-belief querries (false-belief representing only a small subset of social-intelligence in general).
We are encouraged by that relevant and fascinating work, and we believe it can be further enriched by a systematic overview of different aspects of social intelligence as presented here.

We do not claim that the SocialAI school is sufficient to construct a socially competent agent as this is a very far-reaching and complex goal.
However, we believe that in aiming for this goal, concepts from developmental psychology can serve as signposts for AI - give directions and enable us to define short term goals.
Given that the outlined skills are at the very core of human social and cognitive competences, artificial agents aimed at participating in and learning from social interactions with humans are likely to require the same core competences. 
We present the SocialAI school merely as a first step towards this goal.

Following the theories of Michael Tomasello and Jerome Bruner, this work identifies a richer set of socio-cognitive skills than those currently considered in most of the AI research.
More precisely, we focus on three key aspects of social cognition as identified by Tomasello: 1) social cognition: the ability to infer what others see and to engage in joint attention, 2) communication: the development of referential communication through pointing and the beginnings of conventionalized communication through language, and 3) cultural learning: the use of imitation and role reversal imitation in social learning.
We also outline two concepts from Jerome Bruner's work: formats and scaffolding. Formats refer to the way in which social interactions are structured and presented, while scaffolding refers to the temporary support provided by a caretaker to help a learner achieve a task that would be otherwise too difficult.

Based on this set of target abilities, we construct the SocialAI school, a tool (based on MiniGrid \cite{gym_minigrid}) which enables the construction of social environments whose diverse grid-world scenarios affords rich yet tractable research around social competence acquisition. Considered social scenarios are organized according to the key cognitive science experiments used to study the social cognition in children by highlighting core developmental steps.

In our experiments, we aim to show the versatility of the experiments which could be conducted with the SocialAI school. 
We present experiments regarding the following questions: generalization of social inferences (the pointing gesture) to new contexts, recreating an experiment from cognitive science (to study the knowledge transfer during role reversal), and the impact of a scaffolded environment on the agent's learning. 
To show the diversity of agents which can be used, we conduct those experiments with RL agents, and present an additional case study with LLMs as interactive agents.
In the appendix, we explore many more questions such as linguistic inferences, joint attention, and imitation.
We hope to encourage future work extending and building on these first experiments to study various questions regarding social competence. For example, new socio-cultural scenarios, architectures, training regimes, and so on.

We outline the following main contributions of this work:
\begin{itemize}
\item An introduction to Michael Tomasello's and Jerome Bruner's theories on child development and core socio-cognitive abilities
\item An outline of a set of core socio-cognitive abilities important for current AI research
\item The SocialAI school: a tool including a customizable procedural generation suite of environments aiming to simplify studies of socio-cognitive abilities of AI agents
\item Examples of case studies demonstrating how SocialAI can be used to study various questions regarding socio-cognitive abilities in AI
\end{itemize}

\vspace{0.2cm}\textbf{Social agents are not objects.~~} Although social peers could be seen as merely complex interactive objects, we argue they are in essence quite different.
Social agents (e.g. humans) can have very complex and changing internal states, including intents, moods, knowledge states, preferences, emotions, etc.
In cognitive science, an affordance refers to what things or events in the environment afford to an organism \shortcite{affordance}.
The resulting set of possible interactions with peers (social affordances \shortcite{social_affordance}) is essentially different from those with objects (classical affordances).
A flat surface can afford "walking-on" to an agent, while a peer can afford "getting help from".
The latter is a social affordance, which may require a social system and conventions (e.g. politeness), implying that social peers have complex internal states and the ability to reciprocate. Successful interaction might also be conditioned on the peer's mood, requiring communication adjustments.
Training an agent for such social interactions most likely requires drastically different methods -- e.g. different architectural biases -- than classical object-manipulation training.
In \socialai~we simulate such social peers using scripted peers. We argue that studying isolated social scenarios featuring scripted peers in tractable environments is a promising first step towards designing proficient social agents.

\section{Related work}
\label{related}

\subsection{Earlier calls for socially proficient agents}
This work aims to connect the recent social AI literature to the older developmental robotics field \shortcite{asada2009-dev-rob,cangelosi2014dev-rob}, which studies how to leverage knowledge from the cognitive development of human babies into embodied robots. Within this field, multiple calls for developing the social intelligence of autonomous agents have already been formulated \shortcite{billard99,lindblom-vygot-and-beyond,MIROLLI2011298}. The emphasis on the importance of social interactions for learning is probably what led Bruner to conceptualize the notion of formats (pragmatic frames) \shortcite{bruner85childstalk}, which has later been reused for example as a conceptual tool to theorize language development \shortcite{rohlfing_2016}. We intend to further motivate the relevance of this notion to enable further progress in DRL and AI. 

\subsection{Human-Robot Interaction}
Interactions with knowledgeable human teachers is a well-studied form of social interaction. Many works within the Human-Robot Interaction (HRI) and the Interactive Learning field studied how to provide interactive teaching signals to their agents, e.g. providing instructions \shortcite{grizou-2014-hri-instructions}, demonstrations \shortcite{argall-LfD-2009,grollman-fail-demo-2011}, corrective advice \shortcite{celemin-advice}, and even narratives \shortciteA{Mealier_2017_narrative}.
A review of this field \shortcite{vollmer_2016} argues that restricted predefined (not learned) interaction protocols (pragmatic frames) are usually used, and suggests the study of a broader set of social situations.
Catalyzing research on DRL and social skills seems even more relevant now that many application-oriented works are beginning to leverage RL and DRL into real-world humanoid social robots \shortcite{instrinsicHRI2018,socialRobots2021}.

\subsection{Disembodied Social Interaction Understanding}
Rather than directly learning behavior policies, multiple works focused on the study of disembodied machine learning models able to understand synthetic image or videos of social interactions. 
Two experimental setups are considered in this literature: classification tasks and prediction tasks.
In the former, the objective is to correctly label the nature of an observed social scenario, e.g. is the interaction surprising or expected \shortcite{shu2021agent}, are agents being cooperative, neutral or adversarial \shortcite{shu2020flatland}. Other works considered more precise scenario classifications \shortcite{netanyahu2021phase,tejwani2022incorporating}. For instance, in two-agents scenarios, \shortciteA{netanyahu2021phase} proposed a Bayesian approach to jointly detect each agent goals (protect object, move object) and their relative relationships (friends, opponents).
In prediction tasks, machine learning models are evaluated on their capacity to predict agent actions, which is especially useful to design theory of mind experiments \shortcite{DBLP:conf/icml/RabinowitzPSZEB18,baker2011bayestom}, along with more general social perception assessments \shortcite{netanyahu2021phase}.

Some recent works studied social reasoning abilities of large language models through textual problems. 
\shortciteA{sap2022neural} show that GPT models struggle on two question answering benchmarks: SocialIQA \shortcite{socialiqa} and TOMi \shortcite{tomi}.
\shortciteA{trott2022large} evaluate GPT models on variations of the Sally-Anne false-belief tasks and observe promising success rates (still below human performance).
Furthermore, \shortciteA{ullman_alterations_tom} show that GPT models fail on simple alterations of false belief tasks.
\shortciteA{ruis2022large} evaluate LLMs on problems (implicatures) which can only be resolved by understanding contextual information and show a significant gap with human performance. Furthermore, on more context-heavy examples, they show that increasing model size does not lead to performance improvements.


While our ambition is analogous to these aforementioned works– inviting ML scholars to focus on social interaction studies – the present work proposes to take an embodied and interactive perspective on sociality, whose experimental setups better aligns with real-world objectives: socially proficient interactive agents.

\subsection{Embodied Social DRL agents}

In the recent DRL literature, multiple agents able to showcase social skills have been presented.
\shortciteA{jacques2019-socialinfl} presented a multi-agent social dilemma environments requiring the emergence of cooperative behaviors through communication. Authors then showcased agents leveraging the maximization of causal influence as a way to foster cooperation.
In \shortciteA{ndousse2021emergent}, authors showed that, through the addition of an auxiliary next-state prediction task, DRL agents learning to perform navigation tasks among expert policies were able to learn to imitate social peers to overcome hard-exploration scenarios. \shortciteA{cultural2022learning} presents a similar social-imitation approach able to scale to complex 3D environments and to imitate experts online, i.e. within episodes (rather than through gradient-based updates).
In \shortcite{lee2021joint} authors showcase agents able to perform joint attention in cooperative tasks. They show that their intrinsic incentives towards joint attention helps to learn from experts (social learning).
In \shortcite{franzmeyer2021learning} authors present an intrinsic motivation mechanism able to foster altruistic helping in a multi-agent setting without requiring to know the true goal of other agents (the intrinsic signal is based on the maximization of other's choices).

One of the objectives of the \socialai project is to provide rich social scenarios in which to study and iterate on such learning systems within a broader range of social interactions.
For example, The SocialAI school simplifies the design of generalization tests, which are crucial to differentiate heuristic policies from robust social proficiency, as demonstrated in \shortciteA{aru2023mind} in theory of mind experiments.

\subsection{Multi-Agent Emergence of culture and communication}

Multi-agent systems are are an important subfield of interactive agents research.  It includes studying symbol negotiation among cooperative peers\shortcite{lazaridou2020emergent,moulin2020multi}, and scenarios where multi-step cooperation and communication is required for successful interaction: e.g. \shortciteA{DBLP:conf/aaai/MordatchA18} propose simple navigation environments with to study the emergence of grounded compositional language, \shortciteA{jacques2019-socialinfl} present multi-agent social dilemma environments requiring the emergence of cooperative behaviors through (non-verbal) communication. 
\shortciteA{nisioti2023dynamics} study how the interaction of agents and their changing environment leads to niche construction. 
\shortciteA{simulacra} use LLMs to simulate a complex social interactions such as organizing a party.

Closer to our work, an RL agent was shown to adapt (through social learning) to a new environment with an expert \cite{ndousse2021emergent}.
The independent RL agent was trained in a multi-agent environment with various environmental constraints and an auxiliary loss.
Similar experiments were also conducted at a larger scale \cite{cultural2022learning}.
The objective of the SocialAI school is to provide a tool simplifying similar studies, which could explore socio-cognitive abilities outlined by psychology.

While multi-agent emergence of culture is an interesting research direction to study, the present work propose to focus on a complementary setup, arguably closer to human infants' challenges: How to design agents able to enter an already existing social world? Rather than negotiating new modes of communication, how to learn existing social norms?

\subsection{Similar tools for fostering research}
Similar tools have been constructed, often in the form of benchmarks, to support various research including instruction-following \shortcite{chevalierboisvert2019babyai,MisraBBNSA18,RuisBench2020}, embodied question answering \shortcite{DBLP:conf/cvpr/GordonKRRFF18,eqa}, collaboration given human demonstrations \shortcite{puig2021watchandhelp,wan2022handmethat}, or text-based social environments requiring dialogue \shortcite{UrbanekLIGHT,how-to-dragon-2020,fantasy-text-world-2020}.
In contrast to those, we focus on fundamental socio-cognitive abilities and do not aim to create a benchmark.
By building on top of MiniGrid \cite{gym_minigrid}, we aim to provide a tool which can facilitate a diversity of research directions stemming from the outlined socio-cognitive abilities.

\section{Cognitive science background}
\label{sec:cog_sci_background}
The following section introduces core concepts and experiments from the two developmental psychologists that inspired the SocialAI School: Michael Tomasello and Jerome Bruner.

\subsection{Michael Tomasello - The Shared Intentionality Theory}
\label{sec:cog_sci_background_MT}

We are born into a culture filled with cultural artifacts, symbols and institutions like language, social norms, tool industries, or even governments \shortcite{boyd_richerson_2006,tomasello2019becoming}. 
These artifacts were not invented at once, rather they are a product of a series of improvements and modifications over many generations.
Tomasello calls this powerful form of cultural transmission \textit{cumulative cultural evolution}, and he argues that it is behind the most impressive human achievements \shortcite{tomasello_cultural_origins_1999}

Cumulative cultural evolution is grounded in our socio-cognitive abilities (e.g. social cognition, cultural learning, communication), which enable us to learn, improve, and teach our culture \shortcite{tomasello2019becoming}, i.e. \textit{enter} a culture.
Cultural artefacts inherited and learned in this process become the very core of our cognition.
An example of this is language, which influences our cognition in many ways. 
For example, it defines how we categorize and construe the world, and enables a powerful form of social learning : learning from instructions \shortcite{tomasello_cultural_origins_1999}.
This makes socio-cognitive abilities crucial, as their early development bootstraps both our social and asocial cognition \shortcite{herrman_2007_cultural_intelligence_hypothesis}.

Tomasello's \textit{Shared intentionality theory} argues that human socio-cognitive abilities, such as communication and social learning, are transformed by two big developmental steps
\footnote{These steps are referred to as \textit{maturational capacities} to highlight that both the maturation and the exposure to relevant experience is required for those developmental steps}
:
the emergence of \textit{Joint intentionality}  at around 9 months of age (the 9-month revolution), and the emergence of \textit{Collective intentionality} at around 3 years of age (the objective/normative turn) \shortcite{tomasello2019becoming}.

\paragraph{Joint intentionality} emerges at around 9 months of age \shortcite{tomasello2019becoming}.
It enables children to form a \textit{joint agent} (a dyadic “we”) - they understand that they work with a partner towards the same joint goal.
Children begin to view dyadic social interactions through a \textit{“dual-level structure”}: a joint agent "we" on one level, and a personal "I" on another, i.e. we both understand that we both have separate roles ("I"), and that we work together towards the same joint goal ("we").
This enables them to take the perspective of others, which can also be done recursively - 
they are not only both attending to the same goal, they are also both attending to the partner's attention to the goal, and they both know that they both are doing so.
This recursive thinking is also manifested in \textit{socially recursive inferences}: recursively embedding one intentional or mental state inside another.
When interpreting a pointing gesture, we make a recursive inference of what "\textit{you} intend for \textit{me} to think". 
For example, if we are looking for a ball together, and you point to a cupboard behind me.
I should infer that you are drawing my attention to the cupboard to communicate that I should look for the ball in the cupboard.

\paragraph{Collective intentionality} emerges at around 3 years of age \shortcite{tomasello2019becoming}.
It enables children to form a cultural \textit{group-minded “we”}, which in comparison with a dyadic "we" represents an identity for a group.
For example, a child might enforce a social norm because "this is how \textit{we}, in this culture, do things".
Consequently, children begin to participate in conventions and norms, and to view things from the “objective” perspective.

These two developmental steps transform countless abilities, motivations, and behaviors.
For the purpose of this paper, we focus on the following three developmental pathways:
social cognition (sec. \ref{sec:cog_sci_background_MT_soc_cog}), communication (sec. \ref{sec:cog_sci_background_MT_comm}), and social learning (sec. \ref{sec:cog_sci_background_MT_soc_learning}), as we consider them the most relevant for AI at the moment.

\shortciteA{tomasello2019becoming} argues that the 9-month-revolution and the objective/normative turn are uniquely \textit{human} developmental steps enabling uniquely \textit{human} socio-cognitive abilities. 
There has been a lot of debate regarding this hypothesis \shortcite{dewaal_2016}, and it still remains an open question.
However, for the purpose of this work, the social proficiency of other great apes (or our last common ancestor with them) is not of primary importance.
We find The Shared Intentionality Theory useful because it is systematic, extensive (covers a broad range of social abilities), and exact (is build upon a number of very clearly defined experiments).
Furthermore, it is concerned with the questions regarding the development of core socio-cognitive abilities.
We believe that this makes it a good basis to organize AI research on.

\subsubsection{Social cognition}
\label{sec:cog_sci_background_MT_soc_cog}

In this section, we discuss the development of the ability to coordinate perspectives and view things from the \textit{objective perspective} (a perspective independent from any individual) \shortcite{tomasello2019becoming}.
The starting point is the ability to \textbf{imagine what another sees or knows}.
The next step is the emergence of \textbf{joint attentions (JA)} at around 9 months of age.
Then, joint attention to mental content in the form of linguistic discourse results in \textbf{coordinating different perspectives with the \textit{objective} perspective}

\paragraph{Imagining what others perceive}

The earliest instance of this is when six-month-olds follow the gaze of others \shortcite{DENTREMONT1997569}. 
It is important to note that, as compared to the later emerging ability to coordinate perspectives, this ability requires that only \textit{one} perspective is processed at a time.
Numerous studies have shown that both apes and children are capable of making such inferences \shortcite{hare2001chimpanzees,moll2006level}. 
For example, in \shortciteA{hare2001chimpanzees}, a subordinate and a dominant chimpanzee were presented with a competitive scenario : competing for food.
Results showed that the subordinate chimpanzee attempted to eat the food only if it was hidden from the dominant one. 
This experiment was then extended to children who were presented with two toys: one observed by an adult and one occluded from him.
When asked to help the adult find a toy, 24-month-olds passed the occluded toy \shortcite{moll2006level}.
These experiments, demonstrate that both children and apes are capable of inferring what a conspecific observes - i.e. they are able to infer another's perspective.

\begin{wrapfigure}{R}{0.5\textwidth}
\centering
\includegraphics[width=0.5\textwidth]{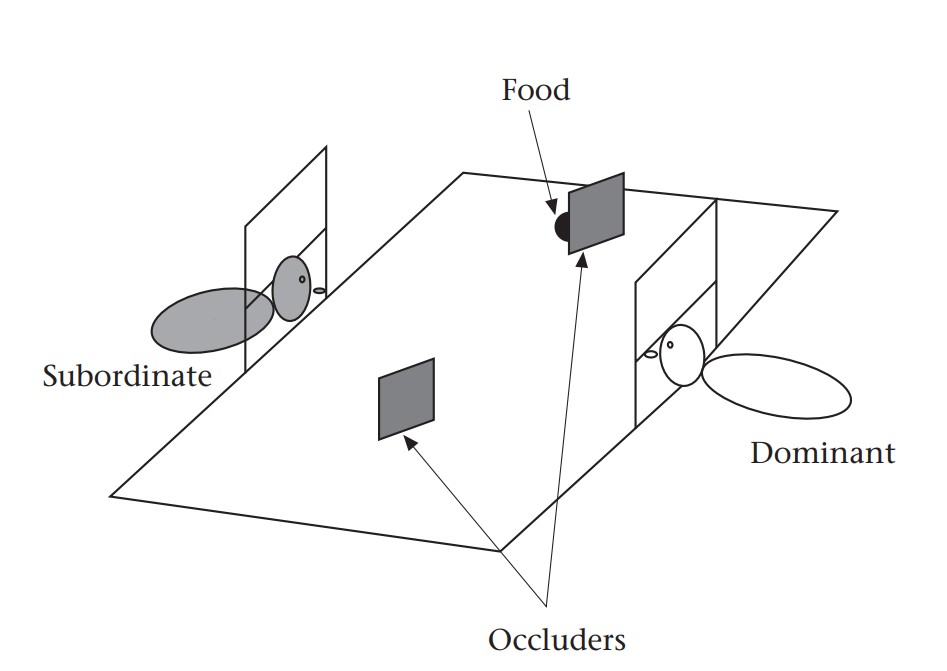}
\caption{\footnotesize Sketch of an experiment from \shortciteA{hare2001chimpanzees} showing that apes can infer the conspecific's field of view. As the subordinate ape does not want to get into trouble, it will not steal the food from the dominant ape. In the experiment, the food was either occluded from the dominant ape or placed in plain sight. The subordinate ape ate the food only when it was occluded from the dominant ape. This shows that it was able to infer the dominant's field of view.} 
\label{fig:ape_food_hiding}
\end{wrapfigure}

\paragraph{Joint Attention}

Joint attention has been defined in various ways \shortcite{sipoova2019ANL}. 
To avoid confusion, we take the definition of joint attention from \shortciteA{tomasello2019becoming}: joint attention consists of two aspects: \textit{triangulation} and \textit{recursiveness}.
Triangulation refers to the child and the adult attending to the same external referent, and recursiveness refers to them both recursively being aware that they are both sharing attention. 
Joint attentions is also characterized by the dual-level structure: shared attention on one level, and individual perspectives on another.
Joint attention enables children to process multiple perspectives at the same time, and they shortly start to align and exchange those perspectives. 

In cognitive science, the emergence of joint attention is studied by counting the child's alternating looks between the adult and referent object, or the child's attempts to initiate joint attention with the adult \shortcite{mundy1986}.
In \shortciteA{carpenter1998_monographs} the amount of joint attention (number of joint attention episodes and their length) was measured in free play interactions between infants and their mothers.
A steady rise in the amount of time spent in joint attention was observed in the period from 9 to 12 months. 
The exact nature of joint attention is not of primary importance for this paper.
It is not disputed that the ability to triangulate, and also be aware that this experience is shared, is of key importance.

\paragraph{Coordinating perspectives}

Once children reach sufficient linguistic competence, they start jointly attending to mental content in the form of linguistic discourse.
They begin to exchange and align perspectives of such content as well.
Through linguistic discourse, children often encounter conflicting perspectives, which they are then pushed to resolve (e.g. one parent says it's raining outside, but another says it's not). 
They resolve those conflicts by learning to form an "objective" perspective - a bird's-eye-view perspective distinct from anyone's personal perspective - and coordinating the conflicting perspectives with it.
For example, they are able to understand that the same object can, at the same time, "look like a sponge" (from their perspective) and "be a rock" (from the objective perspective) \shortcite{flavell81}.
Tomasello argues that this can only be achieved once a child has passed through the second developmental step, that of collective intentionality, which enables them to form such a "perspectiveless" bird's-eye view perspective \shortcite{tomasello2019becoming}.

\subsubsection{Communication}
\label{sec:cog_sci_background_MT_comm}


Communication starts with imperative gestures for self-serving purposes \shortcite{tomasello2019becoming}.
An example of such a gesture is the child pulling the adult's hand, requesting them to pick it up.
This gesture always has the same imperative meaning, and it never refers to an external object.
The 9-month revolution brings forth referential communication (pointing and pantomiming).
The next step is the appearance of conventionalized linguistic communication.
Linguistic communication gives rise to a myriad of different language uses, such as discourse or pedagogy.

\begin{figure*}[htb!]
\centering
\includegraphics[width=1.0\textwidth]{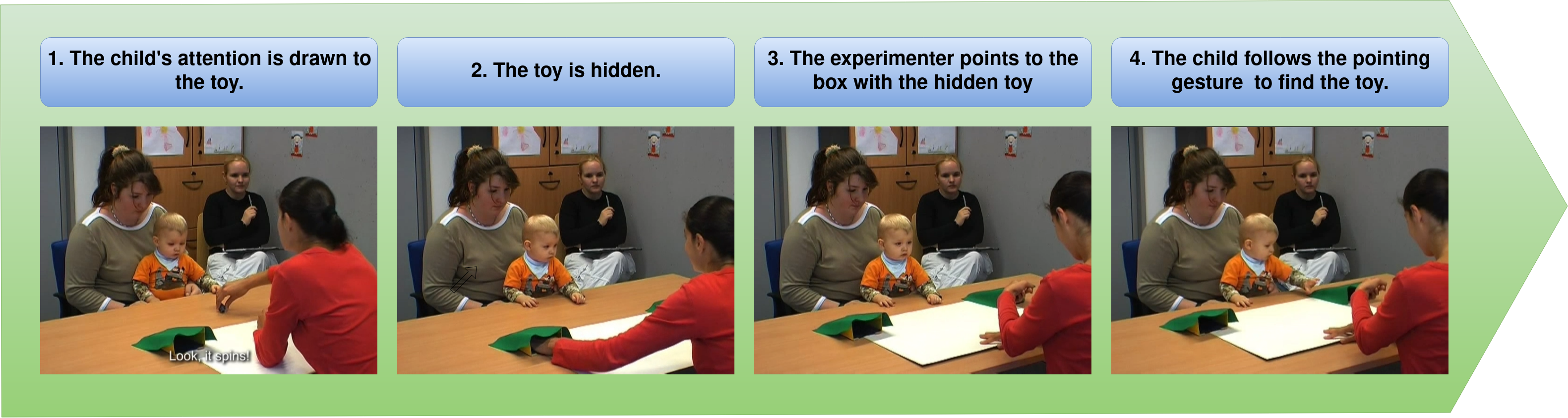}
\caption{\footnotesize An experiment with children from \shortciteA{tomasello_pointing} studying their ability to infer the meaning of a pointing gesture. The child's attention is drawn to a toy. This toy is then hidden in one of the two boxes (the child does not know which one). The experimenter then points to one of the two boxes, and the child is able to infer this to mean that the toy is in that box.}
\label{fig:pointing_comparison}
\end{figure*}

\paragraph{Referential communication - The Pointing gesture}

Following the 9-month revolution, children start to communicate \textit{referentially} - to an external referent \shortcite{tomasello2019becoming}.
This is primarily achieved through \textit{pointing} and pantomiming.
This is made possible by the emerging capacities for  joint attention and recursive inferences. 
The pointing gesture is a powerful way of communicating, as the \textit{same} gesture can be used to express many different meanings in many different scenarios, provided that the observer can correctly infer that meaning. 
This ability to infer the meaning is based on the newly emerging abilities of joint intentionality, most notably that of "socially recursive inferences" - to interpret a pointing gesture, we make a recursive inference of what "\textit{you} intend for \textit{me} to think". 
Hence, when someone directs our attention towards an object, we are able to infer the intended message.

Figure \ref{fig:pointing_comparison} depicts an experiment with children from \shortciteA{tomasello_pointing}. 
First, the child's attention is drawn to the toy, which is then hidden in one of the two boxes. 
The experimenter then points to a box, and the child infers this to mean that the toy is in that box.
14-month-old children were able to successfully follow a pointing gesture to find the toy.
In this scenario, the child makes the following recursive inference: the adult is helping by directing the child's attention to the box, and she wants the child to infer that the toy is in the box. 



\paragraph{Linguistic communication}

Linguistic communication is based on the same principle as gestural referential communication: sharing attention to a referent and recursively inferring the intended meaning.
However, linguistic communication in addition requires learning conventionalized means of reference, such as words or phrases.
Where once was a single pointing gesture, now there is a complex grammar of gestures, with specific conventions assigned to each gesture.
In \shortciteA{carpenter1998_monographs} children's understanding of words steadily increased in the period after 9 months.
This was measured by questioners given to their caretakers on regular intervals. 

Tomasello argues that when language use first appears, children do not yet understand it as conventional, rather they use it as any other artifact or tool.
It is only after the emergence of collective intentionality, when children start to understand and use conventions and norms, that they also begin to perceive language as such.
This is evidenced by specific new ways in which they come to use and understand language.
For example, when others break the rules of a game they protest by normative statements such as "No! It does not go like this!" \shortcite{wyman2009normativity}. 
It is needless to say that language plays many important roles in children's development.
Here we will outline just a few of countless possible examples.
Language provides children with abstract constructions which gives them a new organized format for cognitive representation.
Through discourse, children encounter many conflicting perspectives, which brings them to resolve those conflicts by forming the "objective" perspective. 
Finally, language opens up a new way of cultural learning - instructed learning - in which adults directly teach children  "objective" truths about the world.
Knowledge learned in that manner is easier to generalize \shortcite{butler_tomasello_2016}.

\subsubsection{Cultural Learning}
\label{sec:cog_sci_background_MT_soc_learning}


Human culture is characterized by a powerful form of cultural transmission called \textit{cumulative cultural evolution} - inventions quickly spread and are improved by following generations \shortcite{tomasello_cultural_origins_1999}.
These inventions spread at such a rapid pace that they are rarely forgotten.
This is referred to as the \textit{ratchet} effect \shortcite{tomasello_kruger_ratner_1993_cultural} - as inventions are iteratively improved without \textit{slippage}.
This is made possible by advanced social learning abilities, such as imitation and instructed learning, but also by motivation not only to learn instrumental actions, but also to affiliate and conform. 
Tomasello prefers the term "Cultural learning" for learning motivated by cultural, and not only instrumental, motives.

The earliest form of cultural learning is the mimicking of facial expressions, which is observed even in neonates \shortcite{meltzoff1997explaining_facial_imitation}.
Over the course of the first year, children begin to \textbf{imitate other's actions and goals}, and then, they begin doing so in ways which demonstrate their understanding of other's as intentional agents \shortcite{Meltzoff1995UnderstandingTI}.
Then, \textbf{role reversal imitation} appears as children begin to learn about the partner's role during a collaborative activity.
The next big step in the development of cultural learning is learning from instructions - \textbf{instructed learning} (following the emergence of collective intentionality).
It is based on the adults' motivation to teach children as well as on the children's ability to understand and learn from linguistic instructions. 
It has been shown that children understand knowledge acquired through instructions as objective truth, and generalize it much better than knowledge acquired by other means \shortcite{butler_tomasello_2016}.
It is needless to say that in this way we acquire the most complex knowledge and skills such as reading or algebra.
At around four years of age, children internalize this process, arguably, by reversing the roles (children take on the role of the adult giving instructions).
This leads to a new type of self-regulation, a normative self-regulation based on conventions and norms.

\begin{figure*}[htb]
\centering
\includegraphics[width=1.0\textwidth]{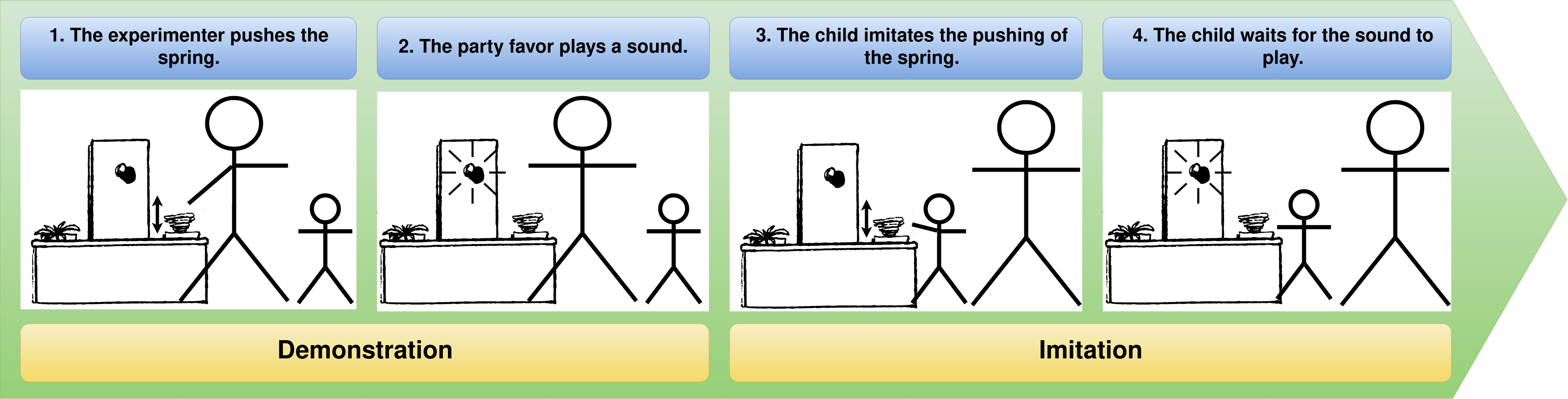}
\caption{\footnotesize Depiction of an experiment from \shortciteA{carpenter1998_monographs}. The experimenter activates the party favor (sound) by pushing the spring, and the child imitates and waits for the sound. The sketch was taken and modified from \shortciteA{carpenter1998_monographs}}
\label{fig:comp_imitation}
\end{figure*}

\paragraph{Imitation and Emulation}
Imitation and emulation learning both refer to observing a demonstration and learning from it.
Imitation learning refers to the learning of means (actions), while emulation to the learning of ends (goals) of a demonstration \shortcite{whiten2004apes,whiten2009emulation_imitation_overimitation,tennie2006push}. Refer to \shortciteA{whiten2009emulation_imitation_overimitation} for a discussion and taxonomy of imitative and emulative learning processes.

Figure \ref{fig:comp_imitation} shows an experiment from \shortciteA{carpenter1998_monographs} studying children's imitation abilities.
In this experiment, the experimenter demonstrates an instrumental action (e.g. pressing a spring attached to a box) which activates the light on top of the box.
The children repeated the instrumental action and looked expectedly at the light.
This kind of learning emerges over the course of the first year - children reconstruct the outcome of others' actions.
However, soon after this, children begin imitating in a way which demonstrates the understanding of other's goals.
Children perform an action that an adult attempted, but failed to perform \shortcite{Meltzoff1995UnderstandingTI}, and do not imitate accidental actions \shortcite{carpenter1998_imitation}. 
Similarly, rational imitation appears. If the action was forced upon the demonstrator, the children recreate the result through more rational means \shortcite{gergely_rational_imitation}. 
For example, in \shortciteA{gergely_rational_imitation} the demonstrator pressed a button with its head while having tied hands and 14-month-olds responded by pressing the button with their hands. 

\textit{Emulation} is a type of social learning where the focus is on the outcome, and not on the actions performed \shortcite{wood1989interaction}.
In other words, the learning is about some property of the environment. 
The learner tries to recreate some observed outcome, in doing so they can, but don't have to, recreate the actions.

On the other side of this spectrum is \textit{overimitation} - children repeat actions that are not relevant for the outcome.
Children often prefer to not only recreate the outcome (as in emulation), but also do it in the same way as the adults (even if this requires doing additional unnecessary actions). 
For example, in \shortciteA{tennie2014limitations} children were presented with a demonstration of a rice-pouring task. The experimenter performed a useless preliminary action before grabbing the rice. 4-year-old children responded by repeating both the useless and the necessary actions.
It has been proposed that children overimitate to affiliate and conform for the purpose of in group bonding \shortcite{over_carpenter_2011_putting_the_social_into_social_learning}, but this remains an open question \shortcite{keupp2013,Lyons2007TheHS}

\paragraph{Role reversal Imitation}

\begin{figure*}[htb]
\includegraphics[width=1.0\textwidth]{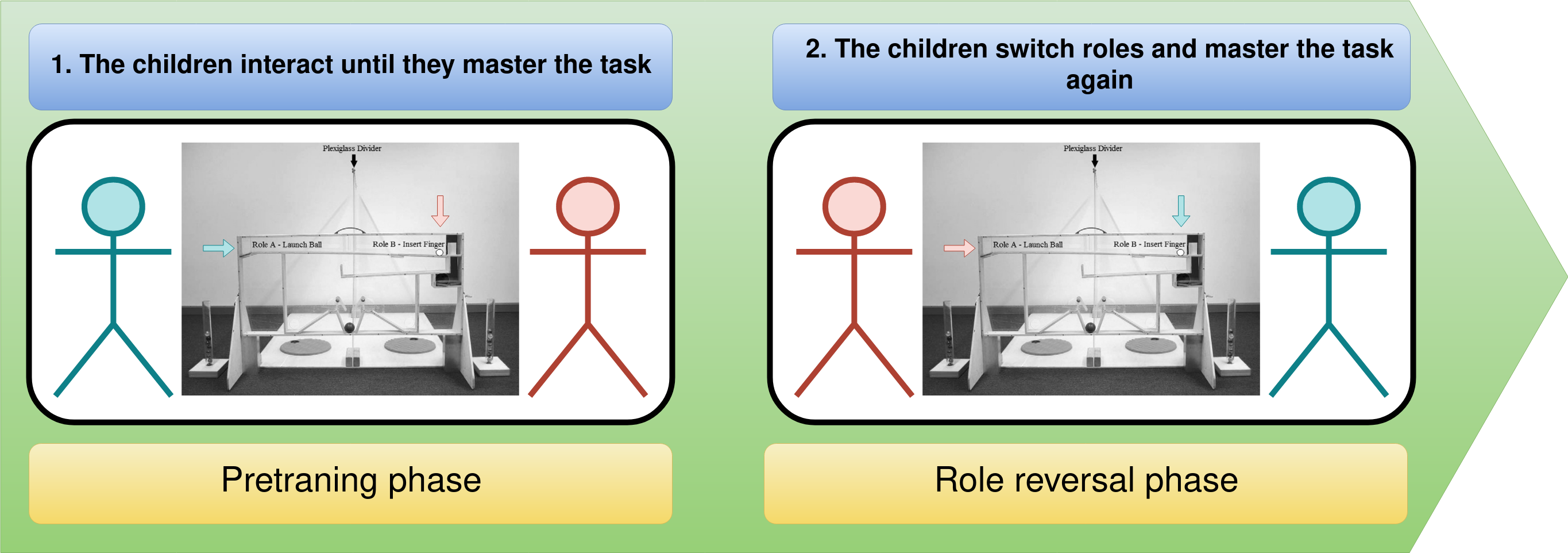}
\caption{\footnotesize Depiction of an experiment on role reversal from \shortciteA{fletcher_tomasello_role_reversal}. The task consists of two roles: one participant pushes a ball into the apparatus, and the other redirects it with their finger. The ball then pushes two marbles toward each of the participants. In the pretraining phase, children collaborate until they master the task (three consecutive successful trials). Then, in the role reversal phase, their roles are reversed and they master the task again. Total number of trials to master the task is compared between the two phases. Children, but not apes, needed less trials to master the task in the role reversal phase than in the pretraining phase.
}
\label{fig:comp_role_reversal}
\end{figure*}

Following the 9-month revolution, a new form of imitation appears - role reversal imitation.
An example of this is when children respond to an adult tickling their arm, by tickling the adult's arm instead of its own \shortcite{carpenter_2005_role_reversal}.
The emerging dual-level structure of joint intentionality enables children to understand, at the same time, the joint goal of a dyadic interaction, and the individuals' separate roles.
This enables the child to reverse the roles of a collaborative activity, and learn about the partner's role from only experiencing its own, which enables much faster transmission and acquisition of cultural practices and knowledge.

Figure \ref{fig:comp_role_reversal} depicts an experiment with children and apes from \shortciteA{fletcher_tomasello_role_reversal}. 
An apparatus is used where one participant pushes the marble, and the other inserts a finger to redirect the ball so that it falls to the correct location.
Then, both participants get a reward.
Children who previously played role A mastered role B in less trials than children who never played role A.
In another experiment \shortcite{carpenter_2005_role_reversal}, children were asked to immediately reverse the role.
An experimenter did some action on the child (e.g. poke the child and say "your turn") and the child responded with the same action on the experimenter (poked the experimenter back).
These experiments show that children understand the separate roles and how each is relevant for the activity.

\subsection{Jerome Bruner}
\label{sec:cog_sci_background_JB}

This work is also influenced by the work of Jerome Bruner, most notably by his concepts of scaffolding \shortcite{wood_bruner_ross_1976} and formats \shortcite{bruner85childstalk}, which were recently reintroduced to AI as pragmatic frames \shortcite{vollmer_2016,rohlfing_2016}.

\begin{wrapfigure}{htb}{0.5\textwidth}
\includegraphics[width=0.5\textwidth]{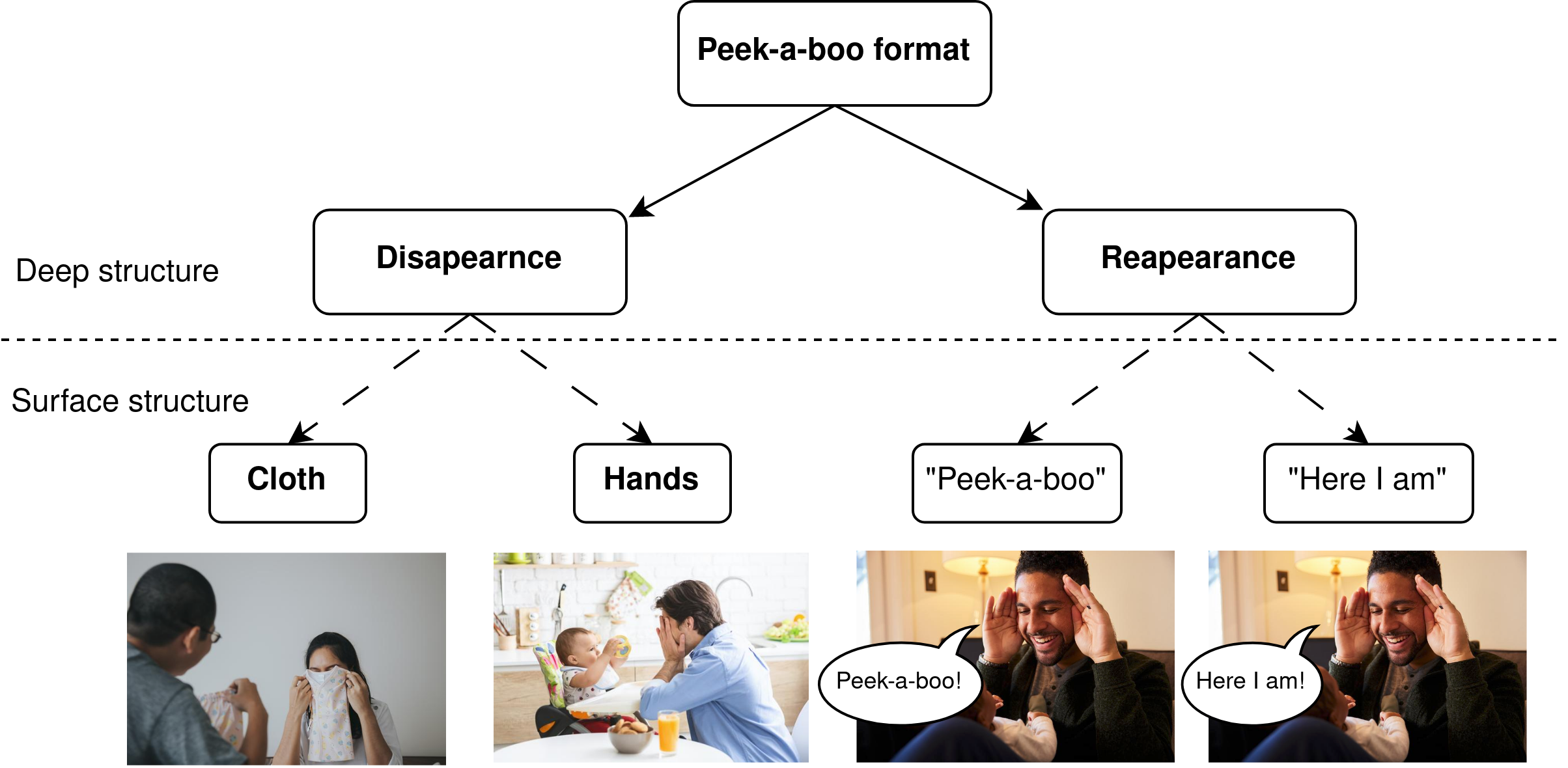}
\caption{
A simplified depiction of a format of the common children's game "peek-a-boo" . Formats consist of the deep structure (the static part), and the surface structure (varying realization managed by some rules). In this example, the deep structure is the disappearance and the reappearance of the adult's face, and the surface structure refers to different ways of hiding the face and signalizing its reappearance.
}
\label{fig:peekaboo}
\end{wrapfigure}

Formats (Pragmatic frames) \shortcite{bruner85childstalk} simplify learning by providing a stable structure to social interactions.
They are regular patterns characterizing the unfolding of possible social interactions (equivalent to an interaction protocol or a grammar of social interactions).
Formats consist of a deep structure (the static part) and a surface structure (the varying realizations managed by some rules). 
An example of a format is the common peek-a-boo game (depicted in figure \ref{fig:peekaboo}).
The deep structure refers to the appearance and the reappearance of an object. 
The surface structure can be realized in different ways. 
For example, one might hide an object using a cloth, or hands; one might hide their face or a toy; one might do shorter or longer pauses before making the object reappear.
We understand social interactions through such formats, and our social interactions are based on our ability to learn, negotiate, and use them. 

Another relevant concept is scaffolding \shortcite{wood_bruner_ross_1976}, which is very similar to Vygotsky's zone of proximal development \shortcite{vygotsky_1978}. 
This concept is also related to Csikszentmihalyi's flow theory \shortcite{csikszentmihalyi_flow}, with the distinction that in flow the learning is not necessarily mediated by a caretaker.
Scaffolding is a process through which the adult bootstraps the child's learning.
The adult controls aspects of a task which are currently too hard for the child, i.e. reduces the degrees of freedom in the task.
Then the scaffold is gradually removed as the child is ready to take on more aspects of the task, until they can solve the task alone (without scaffolding).
An example is a child constructing a pyramid with the help of an adult \shortcite{wood_bruner_ross_1976}.
At first, the child is not even focusing on the task, and the adult tries to get its attention to the task by connecting blocks and building the pyramid in front of them.
Once the child is able to focus on the task, the adult starts passing the blocks to the child to connect. 
In the next phase, the child is grabbing  blocks by itself, and the adult is helping through verbal suggestions.
Then, only verbal confirmations are needed to guide the child.
Finally, the child can construct the pyramid by itself.
We can see how the adult observes the child and gradually transfers parts of the task (removes the scaffold) to the child. Through this process the caretaker enables the child to master a task they would not be able to master alone.

\section{The SocialAI school}
\label{sec:socialai}
The SocialAI school is a tool for building interactive environments to study various questions regarding social competence, such as "What do concepts, such as social abilities and motivations, outlined by developmental psychology mean in the scope of AI?", "How can we evaluate their presence in different agents?", "What are their simplest forms and how can agents acquire them?"

To construct SocialAI, we rely on a set of key experiments and studies from developmental psychology, which were used to outline the most important abilities, motivations and developmental steps in humans.
From the work of Tomasello, we focus on developments before and around the age of 9 months (we believe it is important to address those before more complex ones relating to development of 3-year-olds, see section \ref{sec:cog_sci_background_MT}).
We study the following developmental pathways: Social cognition (inferring other's perception and joint attention), Communication (referential communication through the pointing gesture and the beginning of conventionalized communication through simple language), and Cultural Learning (imitation and role reversal imitation).
From the work of Bruner, we study the concepts of Formats and  Scaffolding (see section \ref{sec:cog_sci_background_JB}). 
Using The SocialAI school, we construct environments and conduct experiments regarding all of those concepts.

SocialAI, which is built on top of Minigrid \cite{gym_minigrid}, includes a \textit{customizable} \textit{parameterized} suite of \textit{procedurally} generated environments.
We implement this procedural generation with a tree-based structure (the parametric tree).
This makes it simple to add and modify new environments, and control their sampling.
All the current environments are single-agent and contain a scripted peer.
The agent has to interact with the peer to reach an apple.
This setup enables a controlled and minimal representation of social interactions. 
To facilitate future research, SocialAI was made to be very easy to modify and extend.
It is completely open sourced, and we hope that it will be useful to the community to study the questions regarding social intelligence in AI.

The remainder of this section is organized as follows. 
First, section \ref{sec:technical_details} describes technical details such as the observation and the action space.
Then, section \ref{sec:param_tree} introduces the parameter tree and explains how it can be used to sample environments.
Finally, section \ref{sec:env_types} describes two environment types, which were used in case studies in section \ref{sec:experiments}.
In the appendix, we discuss one additional environment type (appendix \ref{app:adversarial_type}) and additional case studies (appendix \ref{app:additional-cstudies}).

\subsection{Parameterized Social Environments}
\label{sec:technical_details}

The SocialAI school is built on top of the MiniGrid codebase \shortcite{gym_minigrid}, which provides an efficient and easily extensible implementation of grid world environments.
SocialAI environments are grid worlds consisting of a room. 
In all of our environments, the task of the agent is to eat the apple, at which point it is rewarded.
The reward is diminished according to the number of steps it took the agent to complete the episode.
The episode ends when the agent eats the apple, uses the \textit{done} action, or after a timeout of 80 steps.

\begin{figure*}[htb]
\centering
\includegraphics[width=\textwidth]{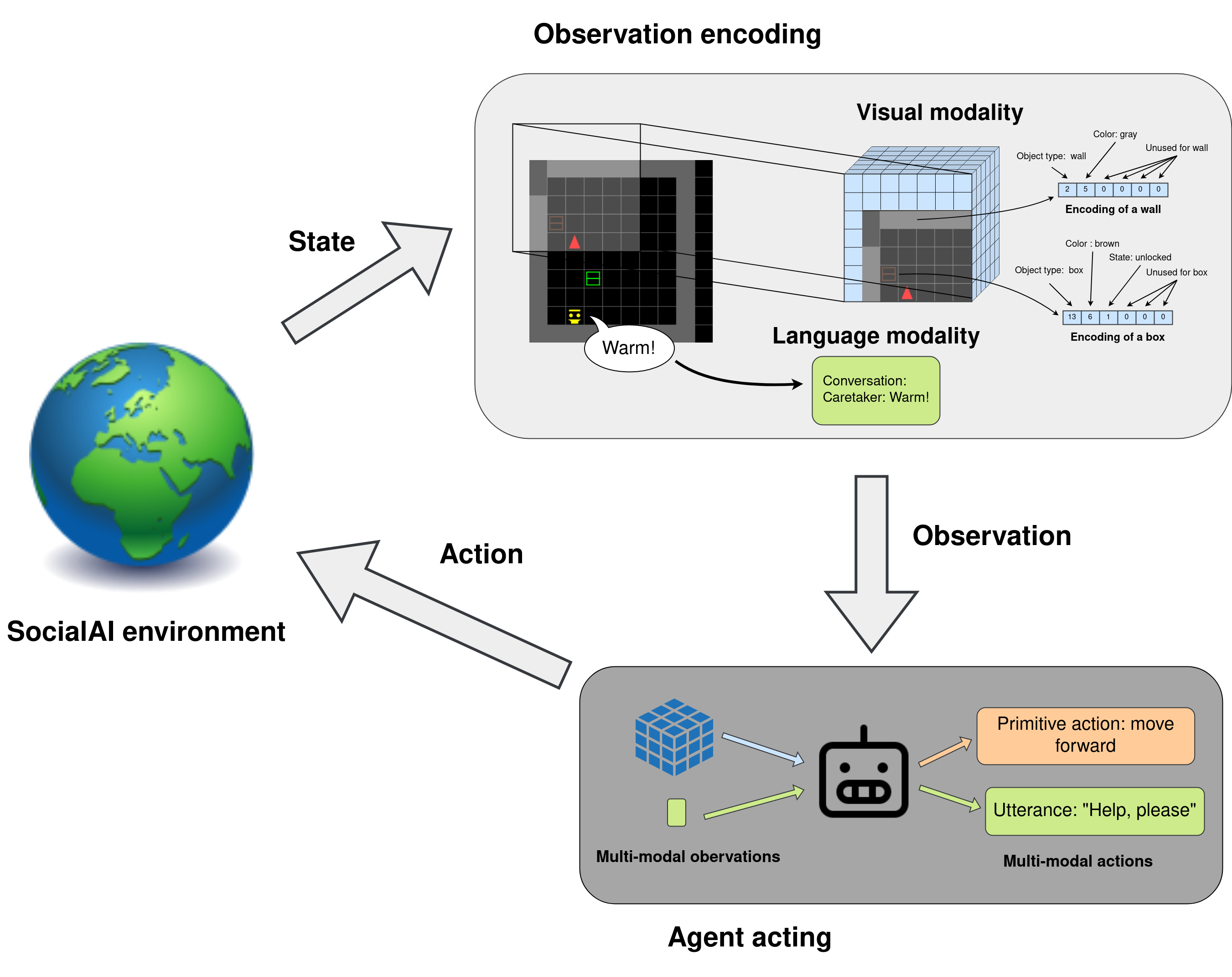}
\caption{\footnotesize Workflow of an agent acting in the SocialAI school. The environment generates a state, which is represented as multimodal observations: a 7x7x6 tensor and the full dialogue history. The agent acts through a multi-modal action space consisting of primitive actions and utterances.}
\label{fig:agent}
\end{figure*}

The agent's observation space is shown in figure \ref{fig:agent}.
This multimodal observation space consists of the full dialogue history, and a 7x7x8 tensor corresponding to the 7x7 grid in front of the agent.
Each cell is encoded by six integers representing the object type, color, and some additional object-dependent information (e.g. is the door open, point direction, gaze direction, etc).
Refer to figure \ref{fig:objects} in the appendix for a list of all objects.

The agent acts in the environment through a multimodal action space, which consists of 6 primitive actions (\textit{no}, movement actions, \textit{toggle}, and \textit{done}) and a 4x16 templated language.
The agent also has the option not to speak, which is implemented with an additional binary output from the agent.
Refer to appendix \ref{app:architecture} for details about the architecture of the agent.

\begin{wrapfigure}{R}{0.6\textwidth}
\centering
\includegraphics[width=0.4\textheight]{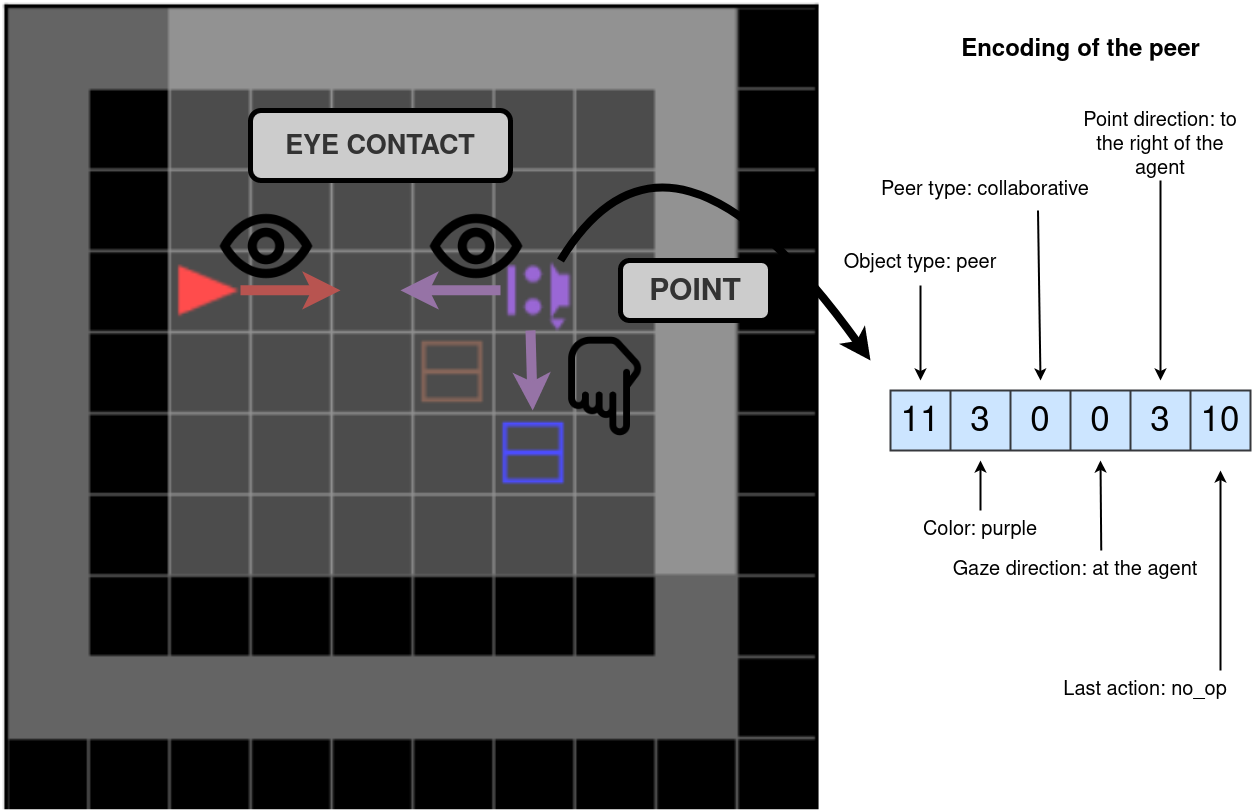}
\caption{\footnotesize A depiction of a peer and its encoding. The agent and a peer are in eye contact, and the peer is pointing to the blue box. To the right is an encoding of the peer. The encoding contains information about the peer, e.g. the gaze and point direction. Refer to figure \ref{fig:objects} in the appendix for a list of all objects.}
\label{fig:npc}
\end{wrapfigure}

All environments, unless otherwise stated, contain a scripted social peer, and the task can only be solved by interacting with this peer (for which socio-cognitive abilities are needed).
A social peer observes the world in the same way as the agent does (as a grid in front of it), and it also observes the agent's utterances.
Their action space consists of primitive actions for movement, pointing, and the \textit{toggle} action.
The peer can also communicate with words and sentences.
As the peer is scripted, there are no constraints on the language it can utter (it is not constrained to a templated language).
The language it uses depends on the environment, which defines which sentence the peer will utter at which point.
The peer is represented in the agent's observation by 7 integers depicting their: object type, position, color, type (cooperative or competitive), gaze direction, point direction, and the last executed primitive action.
The peer's gaze and point directions are represented relative to the agent (e.g. 1 - to the left of the agent).
The pointing direction can also be set to 0, which signifies that the peer is not pointing.
Figure \ref{fig:npc} shows an example of an environment with the corresponding encoding of the peer.
The agent (red) and the scripted peer (purple) are making eye contact - the peer and the agent are in the same row or column and their gazes meet frontally.
In this example, the scripted peer is also pointing to the blue box.

The SocialAI environments are parameterized, and those parameters define the social dimensions of the task.
In other words, parameters define which socio-cognitive abilities are needed to solve the task.
For example, depending on the \mytextsc{Environment type} parameter, the peer can give information, collaborate with the agent, or be adversarial.
In the case of the peer giving information, additional parameters define what is the form of this information (linguistic or pointing).

\begin{figure*}[htb]
\centering
\includegraphics[width=\textwidth]{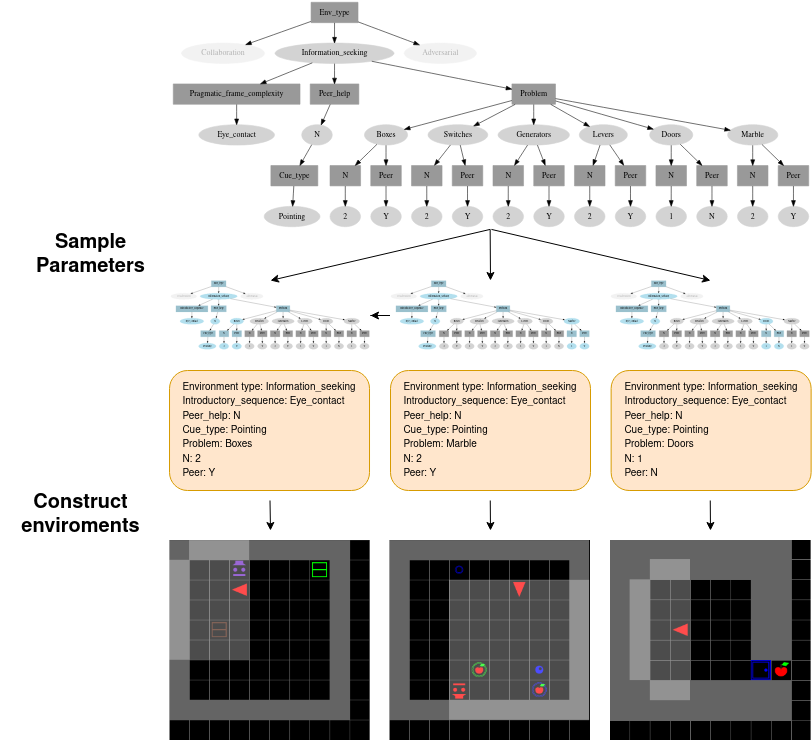}
\caption{\footnotesize An example of procedural environment generation using tree-based parametric sampling.
There are two kinds of nodes: parameter nodes (rectangles) and value nodes (ovals).
Parameter nodes require that one of its children (a value node) is selected.
Value nodes require that sampling progresses through all of its children (parameter nodes).
In this tree, all parameter nodes except "Problem" have only one child.
This means that only the Problem parameter can be set in different ways.
We show three examples of parameter sampling, and the three environments constructed from those parameters.
}
\label{fig:param_tree_example}
\end{figure*}

\subsection{Parameter tree}
\label{sec:param_tree}

SocialAI enables the creation of many parameterized environments, and those parameters are implemented as nodes in a parameter tree.
A parameter tree is a structure through which the experimenter can easily define which parameters (and their values) can be sampled. 
An example of such a tree can be seen in figure \ref{fig:param_tree_example}. 
The standard procedure is that an experimenter defines a parameter tree.
Then each episode begins with the sampling of a new parameter set from this tree.
Once a parameter set has been sampled, an environment is created, and the agent placed inside.

A parameter tree is used to sample parameter sets from it, an example of such sampling is shown in figure \ref{fig:param_tree_example}.
There are two kinds of nodes: parameter nodes (rectangles) and value nodes (ovals).
Parameter nodes correspond to parameters, and value nodes corresponds to possible values for those parameters.
Sampling proceeds in a top-down fashion, starting from the root node.
In all our experiments, \mytextsc{Env\_type} parameter node is the root.
Sampling from a parameter node selects one of its children (a value node), i.e. sets a value for this parameter. 
This can be done by uniform sampling over the node's children, or by prioritized sampling with a curriculum.
Once a value node has been chosen, the sampling continues through it to all of its children (parameter nodes).
In other words, setting a value for one parameter, defines which other parameters (the value node's children) need to be set.
In our codebase, it is simple to create such trees, and add additional parameters and environments.
In the following sections, we explain the most relevant parameters.
Refer to figures \ref{fig:pointing_tree}, \ref{fig:rr_tree} and \ref{fig:scaf_tree} in the appendix for examples of parametric trees. 

\subsection{Environment types}
\label{sec:env_types}
The most important parameter is the environment type - \mytextsc{Env\_type}.
This parameter node is always the root node.
We implemented three different environment types: \mytextsc{InformationSeeking}, \mytextsc{Collaboration}, and \mytextsc{AdversarialPeer}.
A parameter tree doesn't have to contain all of them.
This choice entirely depends on the type of experiment one wants to conduct, most often only one of type will be present in a tree.
For example, figure \ref{fig:param_tree_example} shows the tree with only the \mytextsc{InformationSeeking} environment type.
This tree was used to study understanding of the pointing gesture in section \ref{sec:exp_pointing}. 
In the rest of this section, we describe the \mytextsc{InformationSeeking} and the \mytextsc{Collaboration} environment types.
We describe the \mytextsc{AdversarialPeer} type in the appendix \ref{app:adversarial_type}.

\begin{figure*}[htb]
\subfloat[\footnotesize A scripted peer pointing to a box. The agent needs to open the red box.]{
\includegraphics[width=0.3\textwidth]{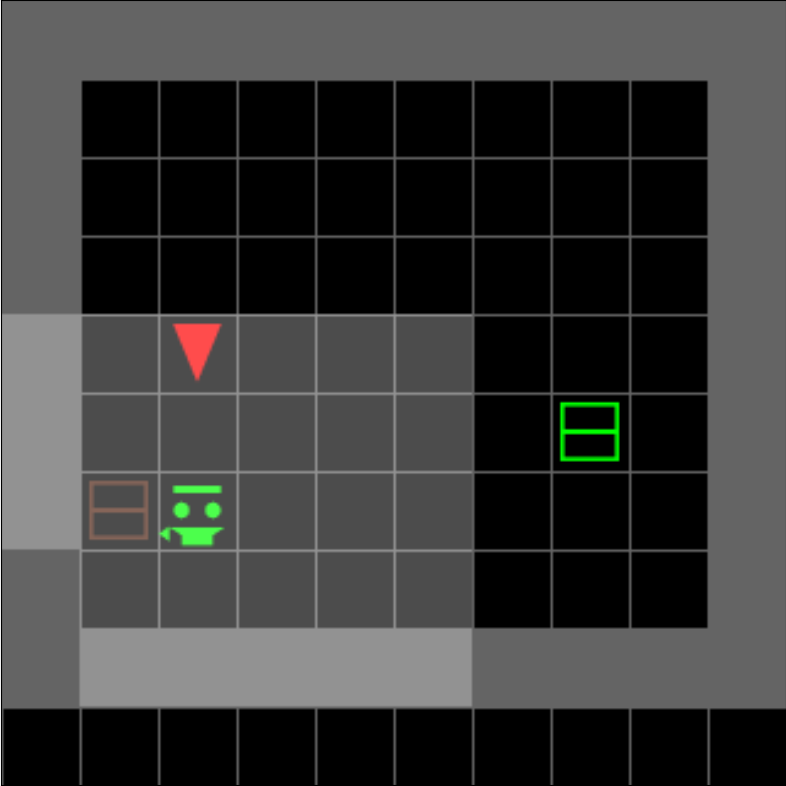}
\label{fig:pointing_boxes_env}
}
\hfill 
\subfloat[\footnotesize A scripted peer uttering the color of the correct generator. The agent needs to push the marble onto the blue generator.]{\includegraphics[width=0.3\textwidth]{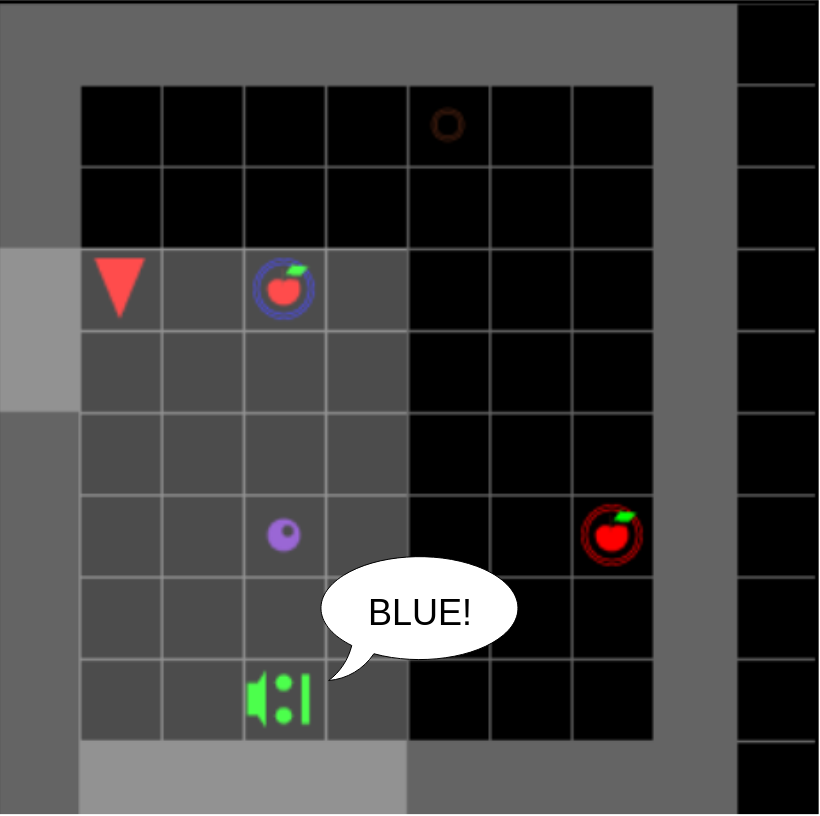}}
\hfill 
\subfloat[\footnotesize A scripted peer hinting the distance to the correct lever ("Hot" means very close). The agent needs to pull the purple lever to open the door.]{
\includegraphics[width=0.3\textwidth]{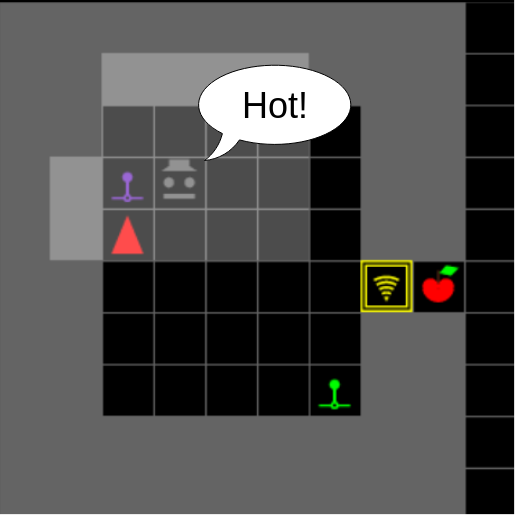}
\label{fig:language_feedback_env}
}
\hfill
\caption{\footnotesize Examples of \mytextsc{InformationSeeking} type environments, in which agents learns to find hidden apples using textual or non-verbal communication with social peers. }
\label{fig:inf_seek_examples}
\end{figure*}

\textbf{Information Seeking type environments}
This environment type will be used in case studies regarding communication, joint attention, and imitation learning. 
In figure \ref{fig:inf_seek_examples} we can see examples of \mytextsc{InformationSeeking} type environments.

The general principle of this environment type is as follows.
The agent is rewarded upon eating the apple, which is hidden.
The apple can be accessed by interacting with an object.
The \mytextsc{Problem} parameter defines which objects will in the environment.
There are six different problems: \mytextsc{boxes}, \mytextsc{switches}, \mytextsc{marble}, \mytextsc{generators}, \mytextsc{doors}, or \mytextsc{levers}.
Different objects make the apple accessible in different ways. 
For example, opening the box will make the apple appear at the location of the box, while pulling the lever will open the door in front of the apple.
A distractor can also be present (if \mytextsc{N} is set to 2).
A distractor is an object of the same type as the correct object.
If the distractor is used, both objects are blocked and the apple cannot be obtained in this episode.

To find out which object is the correct one, the agent must interact with the scripted peer.
This interaction starts with the agent introducing itself. 
The way in which the agent should introduce itself is defined by the \mytextsc{Introductory sequence} parameter.
We define the following four values: \mytextsc{No}, \mytextsc{Eye\_contact}, \mytextsc{Ask}, \mytextsc{Ask-Eye\_contact}.
For the value \mytextsc{No}, no introduction is needed and the peer will give information at the beginning of the episode.
In most of our experiments, we will use the value \mytextsc{Eye\_contact}. For this value, the scripted peer will turn to look at the agent and wait for the agent to look at it.
The agent must direct its gaze directly towards the scripted peer.
An example of an established eye contact can be seen in figure \ref{fig:npc}.
For the value \mytextsc{Ask}, the agent needs to utter "Help, please".
The agent does so using templated language, by selecting the "Help, \textit{X}" template and the word "please".
A full grammar of the language is given in table \ref{tab:grammar} in the appendix.
Finally, the \mytextsc{Ask-Eye\_contact} value is a combination of the previous two.
It requires that the agent utters "Help, please" during eye contact.

Once the agent introduces itself, the \mytextsc{Help} parameter defines the peer's behaviour.
If it is set to \mytextsc{Y} the peer with obtain the apple, and leave it for the agent to eat. Alternatively, it will give cues to the agent about which object to use.
The nature of this cue is defined by the \mytextsc{Cue type} parameter.
We define four different values: \mytextsc{Pointing}, \mytextsc{Language Color}, \mytextsc{Language Feedback}, and \mytextsc{Imitation}.
For the \mytextsc{Pointing} type, the peer will point to the correct object.
It will move to a location from which it can unambiguously point (e.g. the same row) and point to the object.
For the \mytextsc{Language Color} type, the peer will say the color of the correct object.
For the \mytextsc{Language Feedback} type, the peer will hint how close the agent is to the correct object. Every step, the peer will say "Cold", "Medium", "Warm" or "Hot", depending on how close the agent is to the correct object. For example, "Cold" means that the agent is far from the object, and "Hot" that it is right next to it.
For the \mytextsc{Imitation} type, the peer will demonstrate the use of the correct object.
The peer will use the correct object, obtain the apple, and eat it.
Then it will reset the environment to its initial state.

For the purpose of analyzing the agent's behavior more thoroughly, Information seeking environments can also be created without the distracting object, i.e. in their asocial versions. 
This can be achieved by setting parameter \mytextsc{Peer} to \mytextsc{N} and parameter \mytextsc{N} to \mytextsc{1}.
The asocial version of an information seeking environment contains no distractor, and no peer, i.e. the agent just needs to use the only object in the environment.

\begin{figure*}[!ht]
\subfloat[\footnotesize The \mytextsc{MarblePass} problem with the agent in role \mytextsc{B}. The peer pushes the marble to the right and then the agent pushes it further to the purple \textit{marble generator}. This makes two apples appear on the blue and red platforms.
]{
\includegraphics[width=0.3\textwidth]{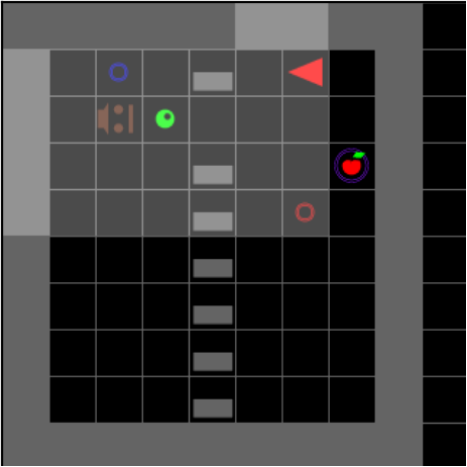}
\label{fig:marble_pass_env}
}
\hfill
\subfloat[\footnotesize The \mytextsc{LeverDoor} problem with the agent in role \mytextsc{B}. The peer opens the red door by pulling on the green lever. This enables the agent to go through the door and activate the purple generator This makes two apples appear on the gray and yellow platforms.]{\includegraphics[width=0.3\textwidth]{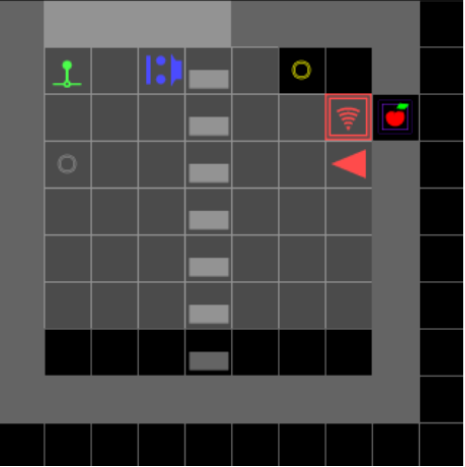}}
\hfill
\subfloat[\footnotesize The \mytextsc{MarblePush} problem with the agent in role \mytextsc{A}. The peer opens the yellow door using the green lever. Then the agent pushes the marble through the door to the purple \textit{marble generator}. This makes two apples appear on the purple and green platforms.]{\includegraphics[width=0.3\textwidth]{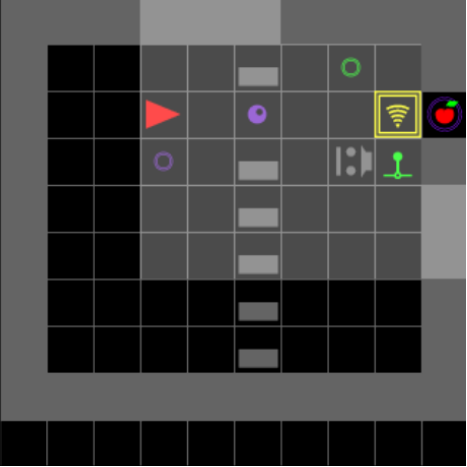}}
\hfill
\caption{\footnotesize Examples of \mytextsc{Collaboration} type environments, in which agents must learn cooperative strategies with a (scripted) peer to achieve two-player puzzles.}
\label{fig:colab_examples}
\end{figure*}

\textbf{Collaboration type environments}
This environment type will be used to study the ability of the agent to reverse roles.
It consists of collaborative activities with two clearly defined roles.
Environments are separated into two halves (corresponding to different roles) by a fence over which the agent can see, but which it cannot cross.
If both roles are fulfilled correctly, two apples will become accessible (one on each side of the fence).

The most important parameters are \mytextsc{Role} and \mytextsc{Problem}.
The \mytextsc{Role} parameter defines in which role to put the agent.
The \mytextsc{Problem} parameter defines the collaborative activity, of which we implemented seven: \mytextsc{DoorLever}, \mytextsc{MarblePush}, \mytextsc{MarblePass}, \mytextsc{Boxes}, \mytextsc{Switches}, \mytextsc{Generators}, \mytextsc{Marble}. 
In \mytextsc{DoorLever} one participant opens the door by pulling the lever and the other passes through them, and activates the generator (generating two apples).
In \mytextsc{MarblePush} one participant opens the door by pulling the lever, and the other pushes a marble through them.
This marble activates the \textit{marble generator} upon contact with it.
In \mytextsc{MarblePass} one participant pushed the marble to the right side of the room, and then the other pushes it towards the \textit{marble generator}.
In the remaining four problems, one participant is presented with two boxes of different colors. The other participant is presented with two objects of the same colors as the two boxes and of the type defined by the \mytextsc{problem} parameter (e.g. two generators).
First, the participant that was presented with boxes opens one box (an apple will be in both).
After this, to obtain its apple, the other participant must use the object of the same color as the opened box.
In figure \ref{fig:colab_examples} we can see examples of \mytextsc{Collaboration} type environments.

Like the information seeking environments, collaboration environments can also be instantiated in their asocial versions.
This can be achieved by setting the \textsc{Version} parameter to \textsc{Asocial}.
The peer is not present in the environment, and the environment is initialized so that the task can be solved alone.
For example, in \mytextsc{MarblePass} the marble is already on the right side of the room, so the agent just has to push it towards the \textit{marble generator}.

\section{Experiments}
\label{sec:experiments}

In this section we demonstrate how the SocialAI school can be used to conduct diverse experiments motivated by cognitive science.
We present a set of case-studies inspired by theories and studies described in section \ref{sec:cog_sci_background}.
To facilitate future research, SocialAI was made to be very easy to modify and extend. 
It is completely open sourced, and we hope that it will be useful to the community to study the questions regarding social intelligence in AI.

The remainder of this section is organized as follows.
In section \ref{sec:baselines} we describe the agents used in case studies with reinforcement learning.
In section \ref{sec:exp_pointing} we evaluate the generalization of socially recursive inferences by RL agents to new contexts - pointing in a new context.
In section \ref{sec:exp_role_reversal_imitation} we show how an experiment from cognitive science can be recreated in the context of AI - we study the transfer of knowledge from one role to another, i.e. role reversal.
In section \ref{sec:exp_scaffolding} we study how an RL agent can be made to learn a complex task by changing the environment (scaffolding) rather than the agent.
Finally, in section \ref{sec:exp_llm} we show how SocialAI environments can be easily transformed to pure text, and how large language models can be used as interactive agents.
Additional case studies are briefly mentioned in section \ref{sec:exp-additional}, and discussed in detail in appendix \ref{app:additional-cstudies}.
These additional case-studies regard linguistic communication, joint attention, meta imitation learning, inferring the other's field of view, and formats (pragmatic frames).  


\subsection{Baselines}
\label{sec:baselines}

In all of our case studies, except the study with language models (sec. \ref{sec:exp_llm}), we use a PPO \shortcite{ppo} reinforcement learning agent as depicted in figure \ref{fig:agent}.
The multimodal observation space consists of a 7x7x6 tensor (vision) and the full dialogue history (language).
The multimodal action space consists of 6 primitive actions (no\_op, turn left, turn right, go forward, toggle, and done), and a 4x16 templated language.
The architecture of the agent is taken from \shortciteA{hui2020babyai} and adapted for the multimodal action space with an additional output head (see appendix \ref{app:architecture}).
This additional head consists of three outputs: a binary output indicating if the agent will speak, and outputs for the template and the word to use.

In a set of pilot experiments (see appendix \ref{app:pilot}), we proposed two exploration bonuses (see appendix \ref{app:exploration_bonuses}).
We compared them to other exploration bonuses including RND \shortcite{rnd} and RIDE \shortcite{ride} on top or PPO \shortcite{ppo} agents.
Visual count-based exploration bonus ("PPO-CB") performed best on the tasks in which language is not used, and its linguistic variant "PPO-CBL" performed best in environments with the peer giving linguistic cues.
For this reason, we sued them in the remainder of our experiments.
Both of those two exploration bonuses are episodic.
They estimate the diversity of observations in an episode and give reward proportional to that diversity.
The linguistic exploration bonus uses the number of different words, and the vision-based exploration bonus the number of different encodings observed.

In the case studies in sections \ref{sec:exp_pointing} and \ref{sec:exp_role_reversal_imitation} we use the "PPO-CB" exploration bonus.
The case study in section \ref{sec:exp_scaffolding} requires raw PPO for the purposes of the study, and one in section \ref{sec:exp_llm} uses LLMs as agents.
In appnedix \ref{app:additional-cstudies}, we use PPO-CB and PPO-CBL (in those case-studies in which the peer provides linguistic feedback).

\subsection{Understanding the pointing gesture}
\label{sec:exp_pointing}

\begin{wrapfigure}{R}{0.5\textwidth}
\includegraphics[width=0.45\textwidth]{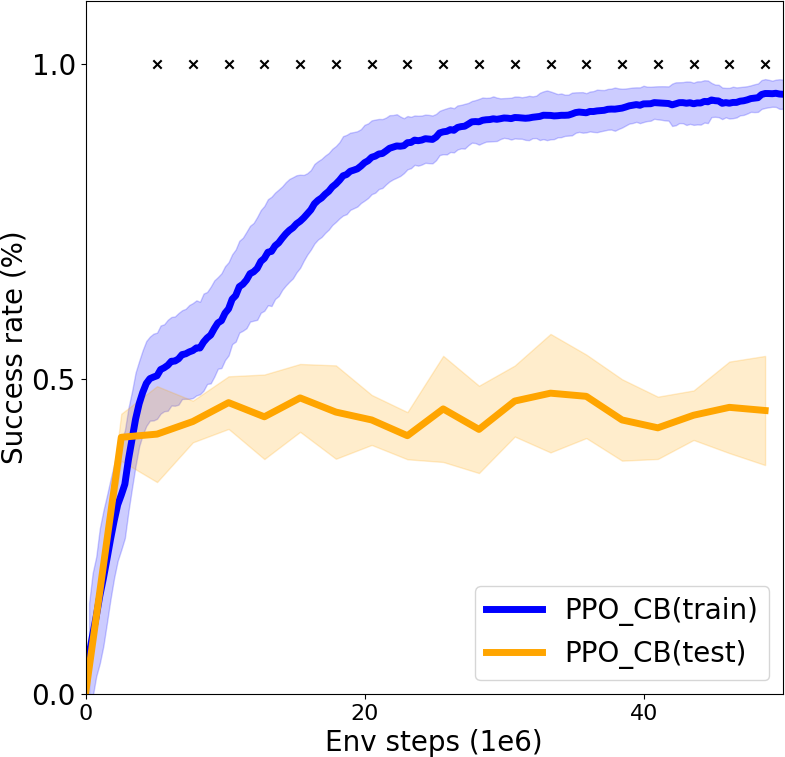}
\caption{
\textbf{The Pointing experiments.} We study if an RL agent is able to infer the meaning of a pointing gesture. 
The agent was trained on five different problems, and on the asocial version of the Doors problem (only one door and no peer in the environment).
The figure compares the success rate (mean +/- std over 8 seeds) on the training environments with the evaluation on the testing environment - the social Doors problem (two door and the peer pointing to the correct door).
The cross marks depict statistical significance ($p=0.05$). We can see that the agent achieves high performance on the training environments, but it is not able to infer the meaning of a pointing gesture in a new context (the social Doors task)). 
Figure \ref{fig:pointing_boxes_env} showns an example of a SocialAI environment with pointing.
}
\label{fig:pointing_exp}
\end{wrapfigure}

This experiment is motivated by a study of childrens' ability to understand pointing gestures \cite{tomasello_pointing}, discussed in section \ref{sec:cog_sci_background_MT_comm}.
We study if an RL agent (with a visual count-based exploration bonus) can infer the meaning of a pointing gesture, and generalize this ability to new situations (infer the new meaning of a pointing gesture in a new context).
This kind of generalization is relevant because the power of inferring pointing gestures is based on being able to infer it's meaning to \textit{new} referents based on \textit{new} social contexts.

The environment consists of two objects (ex. boxes) and the peer that points to the correct object.
The agent then has to interact with that object (ex. open the box) to get access to an apple. 
The agent is trained on five problems each with different objects (Boxes, Switches, Levers, Marble, Generators), and on the \textit{asocial} version of the Doors problem (only one door and no peer).
Training on the asocial version enables the agent to learn how to use a door, which is a prerequisite for generalization of the pointing gesture to an environment with two doors.
The agent is evaluated on the Doors problem in the social setting (two doors and a peer pointing to the correct one).
The agent needs to combine the knowledge of how to use a door (learned on the asocial version of that problem), with inferring the meaning of the pointing gesture (learned on the other five problems), and generalize that to a new scenario where the peer points to a door.
To succeed, it needs to do pragmatically infer the intended meaning of the point (a socially recursive inference).
Refer to section \ref{app:point_exp_params} in the appendix for details.

Figure \ref{fig:pointing_exp} shows the success rate of the agent on the training environments ("PPO\_CB(train)") and its evaluation on the evaluation environment (PPO\_CB(test)).
We can see that while the agent easily solves the training environments (with the success rate of $95.2\%$), it fails to generalize.
It reaches the success rate of $45.2\%$, which corresponds to randomly guessing the correct object.
These results demonstrate that the agent can learn to infer the meaning of a pointing gesture in a familiar context, but cannot generalize to new social contexts.
These results motivate future research on how an agent can be endowed with abilities for such combinatorial generalization, a potential solution could leverage LLMs.

Appendix \ref{sec:exp_language} presents two experiments in which the peer, instead of pointing, provides linguistic cues for the color and for the proximity of the correct object.
As in the pointing experiments, we observe that while PPO agents master the training environments, they fail to generalize to a new context.

\subsection{Role reversal imitation}
\label{sec:exp_role_reversal_imitation}


In this experiment, we study the role-reversal capabilities of an RL agent (with the visual count-based exploration bonus): to what extent can it learn about the partner's role from playing its own. 
In doing so, we also show how a cognitive science experiment can be recreated in the scope of AI.
In \shortciteA{fletcher_tomasello_role_reversal} apes and children were trained on one role (role B), and then tested on how long it took them to master the opposite role (role A).
Results showed that children, but not apes, master role A faster than the control group (not pretrained).
These results imply that children learn about the opposite role just from playing their own, i.e. they see the interaction from a bird's eye perspective.
We study the following two questions:
1) How much do RL agents learn about the partner's role during a collaborative activity?
2) Does increasing diversity in the training (training on more tasks in both roles) enable the agent to learn more about the partner's role?

We conduct this study on the MarblePass task.
This task consists of two roles: one participant pushes the marble to the right side of the environment (role A), from where the other can push it to the a generator, which generates apples (role B).
We aim to assess how much the agent learns about the opposite role (role A), from training in its own (role B).
Following \shortciteA{fletcher_tomasello_role_reversal} we measure the sample efficiency of fine-tuning agents to the test role.
Unlike in \shortciteA{fletcher_tomasello_role_reversal} it is not sufficient to compare an agent pretrained on the training role with an unpretrained agent.
Even if the agent pretrained on the training role learns nothing about the testing role, it would still learn about environment dynamics and one would expect it to learn faster than the unpretrained agent. 
For this reason, we compare with an agent pretrained on the asocial version of the training role.
In this version, the agent obtains reward in the same way as in the social version, but no peer is needed - the agent and the marble are placed on the right side of the environment and the agent has to push the marble towards the generator.
Therefore, this agent learns all about the relevant environment dynamics, but not about the specific collaborative activity.
This agent represents the control group in \shortciteA{fletcher_tomasello_role_reversal}.

We conduct two experiments: \textit{single} and \textit{group}.
In \textit{single} experiments, the agents are trained only on one task : role B and the asocial version of the MarblePass problem.
In \textit{group} experiments, both agents are also trained both roles of all additional six collaborative problems (a total of 13 environments).
In other words, we compare the agents pretrained in the four following ways:
1) experimental (\textit{single}): pretrained only on role B of the MarblePass problem,
2) control (\textit{single}): pretrained only on the asocial version of the MarblePass problem,
3) experimental (\textit{group}): pretrained on role B of the MarblePass problem, and on both roles of all other problems,
4) control (\textit{group}): pretrained on the asocial version of the MarblePass problem, and on both roles of all other problems.
Refer to appendix \ref{app:role_reversal_exp_params} for additional details.

\begin{figure}[htb]
\centering
\subfloat[
\textit{Single} experiment: learning role A given pretraining on role B (1 environment).]{
\includegraphics[width=0.40\textwidth]{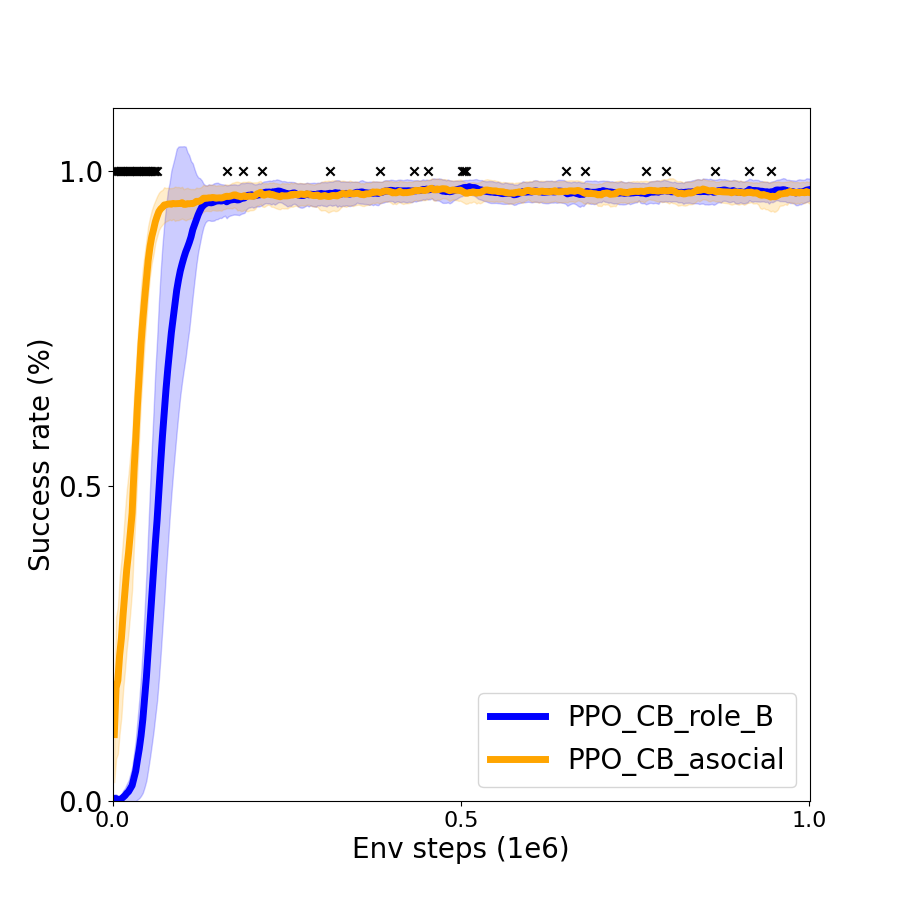}
\label{fig:rr_exp_single}
}
\hfill
\subfloat[
\textit{Group} experiment: learning role A given pretraining on role B and 6 other two-roles tasks (13 environments).]{
\includegraphics[width=0.40\textwidth]{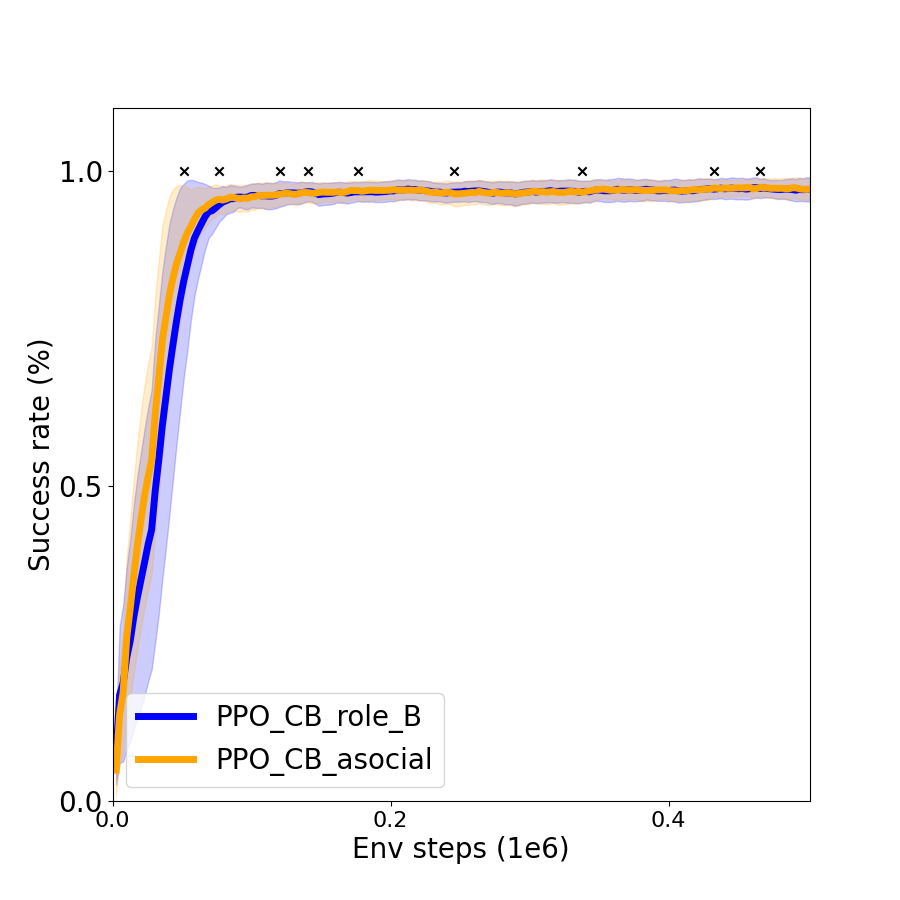}
\label{fig:rr_exp_group}
}
\caption{
\textbf{Role reversal imitation experiments}. We study to what extent is an RL agent able to transfer knowledge from one role of a collaborative activity to another.
Figure shows the success rate of fine-tuning to role A (mean $\pm$ std over 8 seeds), the cross marks depict statistical significance ($p=0.05$).
We compare a PPO agent pretrained on role B ("role\_B") to that pretrained on the asocial version of the environment ("asocial"), which learns only about the environment dynamics. Agents pretrained on role B do not master role A faster than asocially pretrained agents, implying that the RL agents do exhibit role reversal capabilities.
}
\label{fig:rr_exp}
\end{figure}

\textbf{How much do RL agents learn about the partner's role during a collaborative activity?}
Figure \ref{fig:rr_exp_single} shows the success rate of fine-tuning to role A of the MarblePass task.
It compares the experimental and the control conditions of the \textit{single} experiments.
It is interesting to note that the agent pretrained on the asocial version ("asocial") masters role A of the task slightly faster than the agent pretrained on role B of the task ("role\_B").
This implies that, not only, the agent does not learn anything useful about the peer's role, but pretraining on role B actually makes it harder for the agent to learn about role A.
We believe that this is because, during training in role B, the agent learns to first wait for the peer, while in the asocial version it pushes the marble right away.
As, in role A, the agent pushes the marble right away too, we believe this makes it slightly easier for the asocially pretrained agent to adapt to the new role. 
In other words, from an egocentric view the asocial version is closer (than role B) to role A.
This shows that the RL agent, rather than understanding the interaction from a bird's-eye perspective, finds the simplest way to solve the task.

\textbf{Does training on additional problems enable the agent to learn more about the partner's role?}
Figure \ref{fig:rr_exp_group} shows the success rate of fine-tuning to role A of the MarblePass task.
It compares the experimental and the control conditions of the \textit{group} experiments.
Here we can see that there is no significant difference in sample efficiency.
We can make two observations from this.
First, as the socially pretrained agent was less sample efficient in the \textit{single} experiments, we can conclude that pretraining on many tasks reduces overfitting on role B.
And second, as this agent is not more sample efficient than the asocially pretrained baseline, we can conclude that this agent does not learn anything usefull about the peer's role too.

These results imply an interesting avenue of research into how agent's attention can be directed to the partner's role and the birds-eye-view of the activity.

\subsection{Scaffolding}
\label{sec:exp_scaffolding}

\begin{wrapfigure}{R}{0.5\textwidth}
\centering
\includegraphics[width=0.50\textwidth]{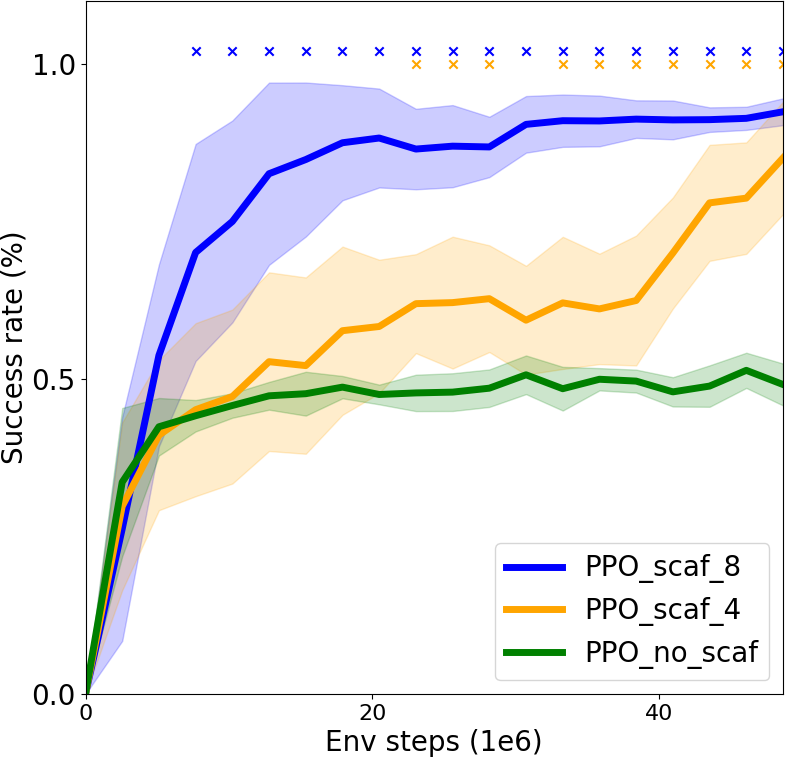}
\caption{
\textbf{The scaffolding experiment.} 
The comparison of agents trained on multiple environments of varying difficulty to that trained on an unscaffolded environment.
The figure show success rates on the testing environments (mean $\pm$ std over 8 seeds) and the cross marks depict statistical significance ($p=0.05$) with respect to the "no\_scaf" baseline.
Only the scaffolded agents ("scaf\_4" and "scaf\_8") solve the environment, and the scaffolding with eight difficulty levels is more sample efficient.
}
\label{fig:scaf_exp}
\end{wrapfigure}
In this section, we study the concept of scaffolding (see sec. \ref{sec:cog_sci_background_JB} for details).
We show how modifying the environment can make it easier for the agent to learn a complex task, i.e. we explore if a scaffolded environment can help an agent learn more complex interaction sequences (formats).
This can be seen in contrast to the standard approach, where the environment is kept fixed and the agent improved (e.g. with an exploration bonus).

For this reason, here we use a PPO agent without an exploration bonus.
From the AI perspective, scaffolding can be seen as analogous to curriculum learning \shortcite{bengiocl}.
In curriculum learning, the task is made gradually more complex, enabling the learner to gradually acquire it part by part.
Scaffolding refers to the caretaker taking a large part of the task on itself, and then gradually, as the learner becomes more proficient, transferring parts of the task to the learner until the learner can do the whole task by themselves.

The environment is similar to the one in section \ref{sec:exp_pointing} with small changes. 
We evaluate on all six problems (instead of one) in the social version.
Instead of pointing, the peer gives linguistic cues for how close the agent is to the target object (e.g. "Hot" for very close), and these cues are given after a more complex introductory sequence (established eye contact and the utterance of "Help, please").
The agent is trained in two phases.
In the first phase, the agent is trained on environments with different complexity. 
After reaching a set success rate, the training goes to the second phase in which the agent is trained only on the six testing environments.
We compare two types of scaffolding: "scaf\_4" and "scaf\_8", which define the enviroments in the first phase.
The agent denoted by "scaf\_4" is trained on four different introductory sequences (requiring or not requiring eye contact and the utterance).
This agent is trained on 18 different environments (six problems, four sequences).
The "scaf\_8" agent is also trained with those four different options.
In addition, the peer can help in two different ways: linguistically hinting to the object or interacting with it and leaving the apple for the agent to eat (36 environments).
The easiest environments on which the "scaf\_8" agent is trained do not require an introduction and the peer leaves the apple for the agent (the agent just goes to the apple and eats it).
The hardest ones require the introduction with both the utterance and eye contact and include the peer linguistically hinting to the object.
Those hardest environments constitute the testing set.
See appendix \ref{app:scaffolding_exp_params} for more details.

Figure \ref{fig:scaf_exp} compares the success rate of the agents trained with the two scaffolding types ("scaf\_4" and "scaf\_8") to that of an agent trained only on the six testing environments ("no\_scaf").
We can see that only the scaffolded agents solve the testing environments, and that the agent with a more detailed scaffolding ("scaf\_8") solves the environment faster.
These results show that scaffolding enables the agents to learn more complex formats, and that a more thorough scaffolding further improves the efficiency.
In future work, more advanced scaffolding could be explored, ex. based on learning progress \cite{oudeyer2009intrinsic} or other surrogate objectives \cite{portelas2020-acl-drl}.

\subsection{Large language models as interactive agents}
\label{sec:exp_llm}

Large language models (LLMs) are staring to be used in various tasks \shortcite{gpt3,bert,opt,instruct_gpt}, including to control interactive agents \shortcite{yao2022react,Carta2023GroundingLL}.
In order to be able to study LLMs as interactive agents, SocialAI school enables the parsing of visual grid observations to pure text, i.e. to Textworlds \shortcite{textworld}.
This process can be easily modified, which simplifies prompt engineering \shortcite{liu2021pre} and similar experimentation.

We use two environments: AsocialBox and ColorBoxes.
In AsocialBox there is a box in the environment and the agent has to open it to get the apple.
In ColorBoxes there are two boxes and the peer.
At the beginning of the episode, the peer says the color of the correct box (the box with the apple).
When testing for generalization on the ColorBoxes environment, we create in-context examples in environments with other objects (e.g. doors, levers) and in the asocial version of the Boxes problem (analogous to the training environments in section \ref{sec:exp_pointing}).
To generalize, an agent must infer the meaning of the peer's utterance in a new context (to select the correct box) and combine this with the knowledge of how to open a box (from the asocial version).

\begin{figure*}[htb]
\centering
\includegraphics[width=1\textwidth]{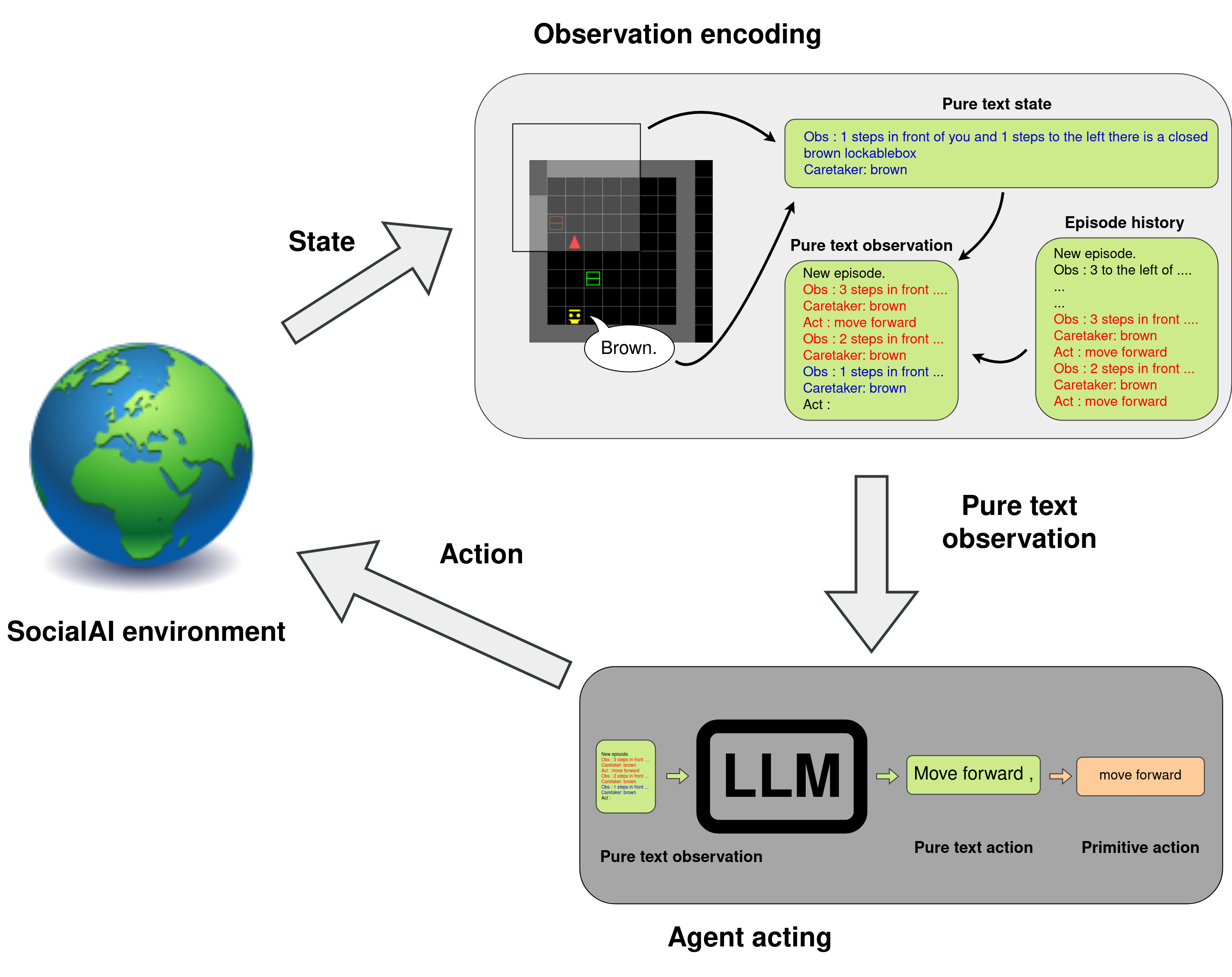}
\caption{
\footnotesize
An example of how a language model can be used as an interactive agent in SocialAI.
A state is parsed into a pure text observation and combined with previous two observations and actions. This is, appended to the in context examples, is used as prompt for the LLM. The agent generates the text which is then matched (as case insensitive substring) with the list of possible actions.
The matched action is executed in the environment.
}
\label{fig:text_envs}
\end{figure*}

A language model acts by generating text, given some textual prompt and the observations are parsed into pure text as shown in figure \ref{fig:text_envs}.
In our experiments, the prompt contains the following: the in context examples, the last three steps (observations and actions) of the current episode, and the action query ("Act :").
We manually create expert trajectories to be used as in context examples - 6 episodes for the AsocialBox environment, and 5 for ColorBoxes (the full in context examples are given in appendix \ref{app:additional_llm}).
The model then generates the textual continuation of this prompt.
\footnote{We generate 3 tokens for GPT models, and 3 words for bloom.}
If one of the available actions ("turn left", "turn right", "move forward", "toggle") is a substring of the generated text, the action is executed and the environment generates the next observation.
However, if no action was matched to the generated text, the "no\_op" action is executed (the agent does not act this step).
The executed action and the new observation are then added to the prompt.

We compare six different large language models: the open-source multilingual bloom-560m \shortcite{bloom} (560M), and five models from the GPT \shortcite{gpt3} family "text-ada-001" (estimated to be 350M \footnote{https://blog.eleuther.ai/gpt3-model-sizes/}), "text-davinci-003" (175B parameters), "gpt-3.5-turbo-instruct-0913", "gpt-3.5-turbo-0613", and "gpt-4-0613".
We also compare with a random baseline, which samples a random action each step.
We evaluate these models on a fixed test set of 10 environments for AsocialBox and 20 environments for ColorBoxes, with a time limit of 15 steps.

Table\ref{tab:LLM_exp} shows that, on the AsocialBox environment, the best GPT models (gpt-4 and davinci-003) achieve a high performance (100\% success rate), despite only observing six expert trajectories.
On ColorBoxes, gpt-4 is the only model to achieve high performance (75\%).
This model escapes the local optimum of 50\% (randomly choosing a box to open), these results imply that the model uses the given social cue (the peer's utterance of the color). 
As gpt-4 was the only model to do so, we test only this model on generalization. 
The model reaches a performance of 55\%, which implies that the model doesn't generalize to a new social context - it randomly chooses a box to open.

The motivation of this experiment was only to show how LLM-based agents can be studied in SocialAI. Therefore, more detailed experiments and analysis are needed to reach stronger conclusions.
Even though the environments used in this case study are simpler than those RL case studies (only the Boxes problem, and no introductory sequence), we find it impressive that such performance is achieved from observing only a few expert trajectories: six for AsocialBox and five for ColorBoxes. 
 
We are optimistic that in future work LLM-based agents could solve much more complex tasks with further prompt engineering and more advanced methods. 
Promising methods include planning \shortcite{huang_llms_as_zeto_shot_planners}, chain-of-thought reasoning \shortcite{cot,zhang2023multimodal}, fine-tuning \shortcite{instruct_gpt,Carta2023GroundingLL}, and many more.
As the main motivation of this case study was to show that it is easy to study large language models with the SocialAI school, we leave those experiments for future work.

\begin{table}[htb]
\caption{\footnotesize Comparison of LLM-based agents on two SocialAI environments parsed into pure text (see figure \ref{fig:llm_envs}). The best model (gpt-4) reached the success rates of 100\% on AsocialBox, and 75\% on ColorBoxes. The score of 75\% suggests that the model is levering the peer to choose the correct box.
When tested for generalization this model reached 55\% success rate implying it is not able to generalize to a novel object. These expriments demontrate how LLM-based agents can be used in the SocialAI School. While more detailed analysis is needed to reach stronger conclusions, the performance is impressive given that the models observed only six (for AsocialBox) and five (for ColorBoxes) expert trajectories. We are confident that with more advanced LLM-based methods better performance can be achieved.
}
\vspace{0.4cm}
\centering
\begin{tabular}{llllllll}
           & \rot{gpt-4} & \rot{gpt-3.5-turbo} & \rot{gpt-3.5-turbo-instruct} & \rot{ada-001}   & \rot{davinci-003}     & \rot{bloom-560m}   &  \rot{random} \\
\midrule
AsocialBox          & \textbf{100\%} & 90\% & 90\% &   90\%    & \textbf{100\%}  &   10\%       &  0\% \\ 
ColorBoxes          & \textbf{75\%} &  5\% & 25\% &      0\%  &    15\%     &    5\%      & 5\%  \\
ColorBoxes (generalization)    & \textbf{55\%} \\
\end{tabular}
\label{tab:LLM_exp}
\end{table}

\begin{figure*}[htb]
\centering
\subfloat[\footnotesize The AsocialBox environment]{
\includegraphics[width=0.48\textwidth]{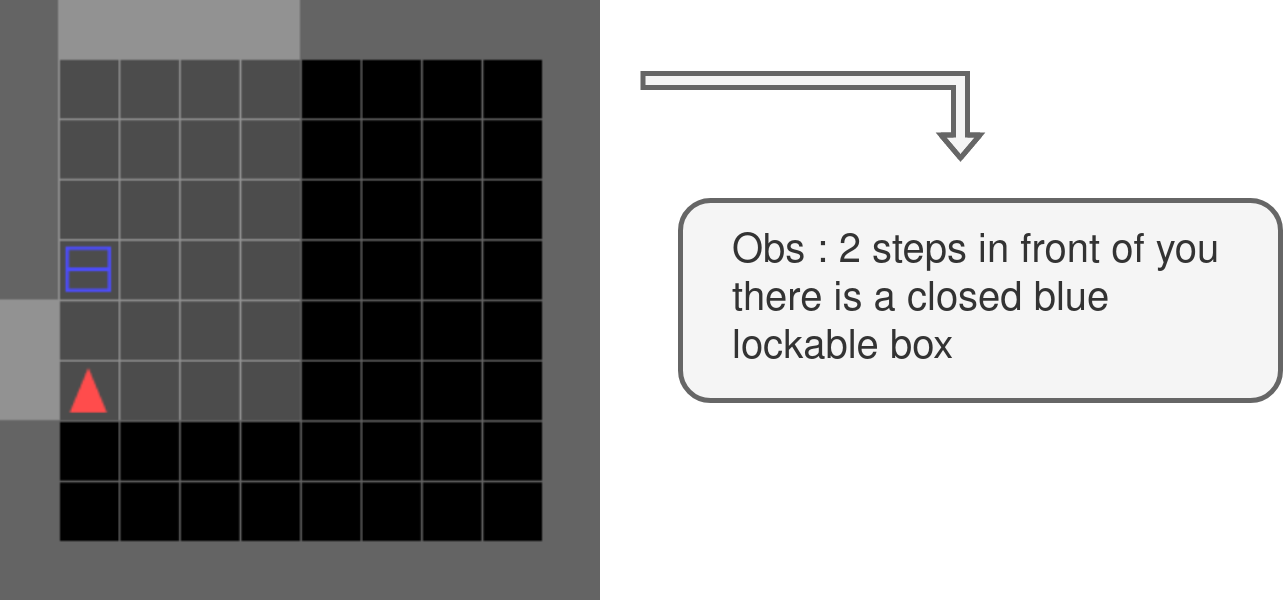}
\label{fig:asocial_box_llm}
}
\subfloat[\footnotesize The ColorBoxes environment]{
\includegraphics[width=0.48\textwidth]{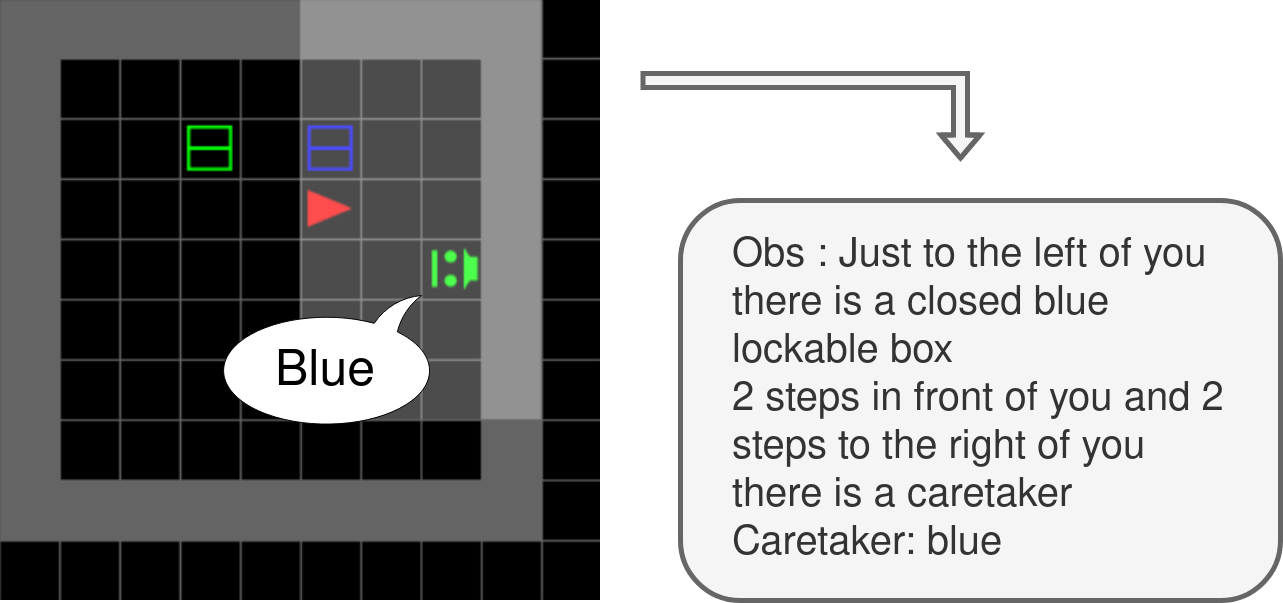}
\label{fig:color_boxes_llm}
}
\caption{\footnotesize Two environemnts used in the experiments with large language models. The observations are parsed into pure text.}

\label{fig:llm_envs}
\end{figure*}

\subsection{Additional experiments}
\label{sec:exp-additional}

We refer interested readers to appendix \ref{app:additional-cstudies} for details on a complementary set of case studies, which we briefly outline in this section.
As mentioned in the pointing case study (section \ref{sec:exp_pointing}), we performed analogous experiments to study whether the agent can leverage linguistic cues instead of the pointing gesture (appendix \ref{sec:exp_language}).
We obtained analogous results: while the agents master the training environments, they fail to generalize to new context.

In appendix \ref{sec:exp_ja}, we study joint attention as defined by Tomasello (see section \ref{sec:cog_sci_background}).
Environments feature a peer providing cues both inside and outside joint attention.
Informative cues are only given inside joint attention (after completing the introductory sequence), while misleading random cues are given outside joint attention.
In our experiments, the agent was unable to sufficiently discriminate between those cues to solve the task.

Appendix \ref{sec:exp_imitation} presents a case-study on the acquisition of an (in-episode) imitation learning mechanism.
From the AI perspective, this can be seen as social meta-learning: the agent acquires (through gradients) the imitation learning mechanism, which is used during the episode to learn a instrumental action on a new object.
This study is motivated by an experiment from cognitive science in which children showed such imitation abilities \shortcite{carpenter1998_monographs}.
Experiments showed that RL agents are not able to acquire a learning mechanism which would enable them to learn how to use a completely new object at test time.

In appendix \ref{sec:exp_adversarial} we test the agent on its ability to infer the peer's field of view. The agent is rewarded for eating the apple under the condition that it was not observed by the peer at that moment.
We show that the agent partially infers the peer's field of view, but is still not able to match the upper performance bound.

Finally, in appendix \ref{sec:exp_formats} we study the acquisition and use of formats as defined by Jerome Bruner (section \ref{sec:cog_sci_background_JB}), i.e. protocols of social interactions. Agents were trained on tasks in which cues can be obtained from a peer after a more complex introductory sequence (\mytextsc{Ask\_Eye\_Contact}).
The results show that, while an RL agent trained without the exploration bonus was unable to learn that introductory sequence, the agent with a linguistic count-based exploration bonus was.
This results can be interpreted in tandem with the scaffolding case study (section  \ref{sec:exp_scaffolding}) in which an RL agent without an exploration bonus is able to learn the most complex introductory sequence, given training in a scaffolded environment.
Therefore, the acquisition of complex formats can be achieved either through changing the learner or the environment.

These additional case studies show further examples of interesting research questions that can be explored with the SocialAI school. 

\section{Conclusion and Discussion}

Following contemporary research in developmental psychology, this work presents and studies a wider set of socio-cognitive abilities than those usually studied in the field of AI.
The motivation of this work is to introduce those concepts to AI and motivate related research.
We present an introduction to Michael Tomasello's and Jerome Bruner's theories of socio-cognitive development.
Following these theories, we outlined a set of key socio-cognitive abilities and concepts for AI: social cognition (inferring other's perception and joint attention), communication (referential and early conventionalized communication), cultural learning (imitation and role reversal imitation), scaffolding, and formats.

We introduce the SocialAI school - a tool simplifying the research of core socio-cognitive abilities.
We show how the SocialAI school can be used to easily create environments studying various questions inspired by developmental psychology.
With RL agents, we conduct experiments regarding the pointing gesture, scaffolding, and role reversal (by recreating an experiment from developmental psychology).
We demonstrate that, by using SocialAI to parse environments into text, Large Language Models be easily studied as well.
In the appendix, we present additional studies concerning linguistic communication, joint attention, imitation learning, inferring others' field of view, and formats.
Our experiments demonstrated the diversity of studies that can be conducted with the SocialAI school, highlighted the limitations of standard RL agents, and showed that while large language models learn with high sample efficiency, additional methods such as fine-tuning or chain-of-thought might be needed for generalization.

\paragraph{Limitations}
In this work, we outline and discuss several concepts from developmental psychology -- mostly regarding the development before and around 9 months of age -- which we found to be most relevant for AI at the moment.
Even among this restricted set it is not reasonable to aim for an exhaustive introduction.
As such, several socio-cognitive concepts are either discussed very briefly (e.g. conformity, social norms, instructed learning) and a lot of others are not mentioned (e.g. morality, fairness, sense of self).
We leave their analysis for future work.
Furthermore, while we argue that the work of Tomasello and Bruner provides an interesting framework to guide AI research in social skill acquisition, many other perspectives could have been considered, e.g. Erik Erikson \shortcite{erikson1993childhood}, Alison Gopnik \shortcite{gopnik1997words}, or Cecilia Heyes \shortcite{heyes_2019}.

Similarly, as the present work merely represents a first step towards socially proficient artificial learners, many technical dimensions were simplified.
In particular, we refrain from free form language dialogues and consider simple templated language.
Likewise, we do not use human or trained peers, but scripted peers (which enables to isolate social abilities).
Rather than implementing rich 3D visual worlds with continuous actions, we use grid-worlds with discrete primitive actions.
We argue that such simplifying assumptions affords tractable studies while maintaining enough social complexity to model and isolate various social challenges.
Assuming progress is made over these social scenarios, an interesting avenue for future work will be to extend the parametric generation towards environments with more complex sensorimotor challenges.

\paragraph{Future work}
Given recent works showcasing the importance of Automatic Curriculum Learning in "asocial" DRL \shortcite{parker2022evolving,portelas2020-acl-drl}, an interesting direction for future work would be to study whether this can also be observed in SocialAI.
Our short case study on the importance of scaffolding (sec. \ref{sec:exp_scaffolding}) suggests a positive impact, although we restricted our analysis to simple expert curricula.
An important challenge will be to design curriculum methods able to leverage the hierarchical structure of SocialAI's parametric tree, rather than the usual low-dimensional flat spaces of task-encoding parameters (predominant in the literature).

Large language models (LLMs) are present in many branches of artificial intelligence.
A promising avenue of future research is the application of language models to interactive agents \shortcite{andreas2022language}.
In this paper, we studied LLMs only on simple environments with a simple method - prompting the model with a few expert trajectories. 
While this approach showed impressive sample efficiency, it is very limited due to the constraints on the prompt size. 
These experiments should be revisited with more powerful methods such as fine-tuning or chain-of-thought prompting. 
Such methods could potentially make more complex social inferences, leading to better performance on many case studies in this paper, especially the ones related to generalization to new scenarios.

An important factor for the observed learning failures of our PPO agents in our case-studies might be linked to the simple forms of exploration bonuses that we used.
Finding efficient exploration bonuses for social settings is a challenging task.
In appendix \ref{app:pilot} we show that RIDE \shortcite{ride} and RND \shortcite{rnd}, two state-of-the-art exploration bonuses from classical DRL underperformed compared to our simple CountBased methods. An interesting avenue would be to study recent exploration bonus methods designed for social scenarios, e.g. \shortciteA{zhang2020bebold}.


\section*{Acknowledgments}
This work benefitted from the use of the Jean Zay supercomputer associated with the Genci grant A0091011996.

\appendix
\clearpage
\section{Architecture of the RL agent}
\label{app:architecture}

\begin{figure}
    \centering
    \includegraphics[width=0.4\columnwidth]{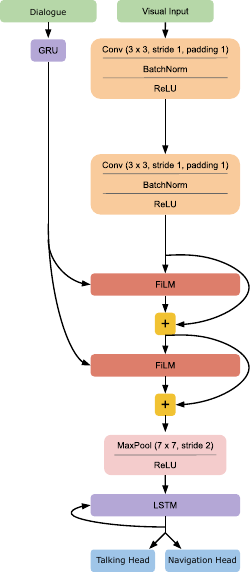}
    \caption{\footnotesize Our Multi-Headed PPO baseline DRL agent. Architecture visualization is a modified version of the one made by \shortciteA{hui2020babyai}. We perform two modifications: 1) Instead of fixed instruction inputs our model is fed with NPC's language outputs (if the agent is near an NPC), and 2) We add a language action head, as our agent can both navigate and talk.}
    \label{fig:baby-model}
\end{figure}

In this work, we use a PPO \shortcite{ppo} with an architecture initially designed for the BabyAI benchmark \shortcite{chevalierboisvert2019babyai}. The policy design was improved in a follow-up paper by \shortciteA{hui2020babyai} (more precisely, we extend their \textit{original\_endpool\_res} model). See figure \ref{fig:baby-model} for a visualization of the complete architecture. First, symbolic pixel grid observations are fed into two convolutional layers \shortcite{lecun1989backpropagation,krizhevsky2012imagenet} (3x3 filter, stride and padding set to 1), while dialogue inputs are processed using a Gated Recurrent Unit layer \shortcite{gru}. The resulting image and language embeddings are combined using two FiLM attention layers \shortcite{film}. Max pooling is performed on the resulting combined embedding before being fed into an LSTM \shortcite{lstm} with a $128D$ memory vector.
The LSTM embedding is then used as input for the navigation action head, which is a two-layered fully-connected network with tanh activations and has an 6D output (i.e. 5 navigation actions and no\_op action).

In order for our agent to be able to both move and talk, we add to this architecture a talking action head, which is composed of three subheads. All of them are consist of two fully-connected layers with tanh activations, and take the LSTM's embedding as input. The first one is used as a switch: it has a one-dimensional output to choose whether the agent talks (output > 0.5) or not (output < 0.5). If the agent talks, the two other networks are used to sample the template and the word. Grammar of the templated language is depicted in table \ref{tab:grammar} and examples of multi-modal actions in table \ref{tab:action_examples}.

Note that the textual input given to the agent consists of the full dialogue history as we found it works better compared to  giving only the current utterance.

\section{Exploration bonuses}
\label{app:exploration_bonuses}

The exploration bonuses we use are inspired by recent works in intrinsically motivated exploration \shortcite{pathakICMl17curiosity,curiositythroughreachability,tang2017exploration}.
These intrinsic rewards estimate the novelty of the currently observed state and add the novelty based bonus to the extrinsic reward.

In this work we present two techniques for computing the count-based exploration bonus.
Both of our count-based exploration bonuses are episodic - they estimate the diversity of states observed within an episode, and assume that beneficial episodes are those with more diverse observations.

\paragraph{Language-based exploration bonus (CBL)} For some utterance $s_{lang}$ observed at state $s$, we count how many times was this utterance observed during the episode. We compute the bonus for this step using the following equation:
\begin{equation}
r_{intr} = T*tanh \left( \frac{C}{ ( N(s_{lang})+1)^{M}} \right)
\label{eq:expl_bonus_lang}
\end{equation}
, where $M$, $C$, and $T$ are hyperparameters and $N(s_{lang})$ is the number of times the utterance $s_{lang}$ was observed during this episode so far.

\paragraph{Vision-based intrinsic reward (CB)} 
We reward the agent for observing diverse encodings. An encoding is the 6D representation of a cell (see figure \ref{fig:agent} for more details). A visual observation consists of 47 (7x7) encodings representing cells in front of the agent. For some visual observation $s_{viz}$ at step $s$, a set of encountered unique encodings is created (duplicates are removed) $U(s_{viz})$, and then the reward computed using the following equation:
\begin{equation}
r_{intr} = T * tanh \left( \sum_{e \in U(s_{viz})}{\frac{C}{(N(e)+1)^{M}}} \right)
\label{eq:expl_bonus_cell}
\end{equation}
, where $M$, $C$, and $T$ are hyperparameters, $U(s)$ is a set of unique encodings visible in state $s$, and $N(e)$ is the number of times an encoding $e$ was encountered in the current episode.

\section{Pilot experiments}
\label{app:pilot}

In this pilot experiment, we compare two exploration bonuses presented in section \ref{app:exploration_bonuses} to RIDE \shortcite{ride}, RND \shortcite{rnd}, and to the agent without any exploration bonus. \footnote{We verify our implementation of RIDE and RND by recreating the results of those baselines on environments from \shortciteA{ride}.}
We encoded the peer in a way which used the mix of egocentric and allocentric vision -  the peer's gaze and pointing direction were encoded in terms of absolute direction ("NSEW").
We decided to change this to fully egocentric as we found it more natural with regards to the question of socio-cognitive artificial intelligence.
We believe that the best performing baselines would also perform best with purely egocentric encodings (the one we use in the rest of the paper). 
For that reason, and to avoid unnecessary energy spending, we do not compare with other baselines on the purely egocentric encoding.

\begin{figure*}[htb]
\centering
\subfloat[\footnotesize Pilot experiments with the peer pointing to the correct object.]{
\includegraphics[width=0.40\textwidth]{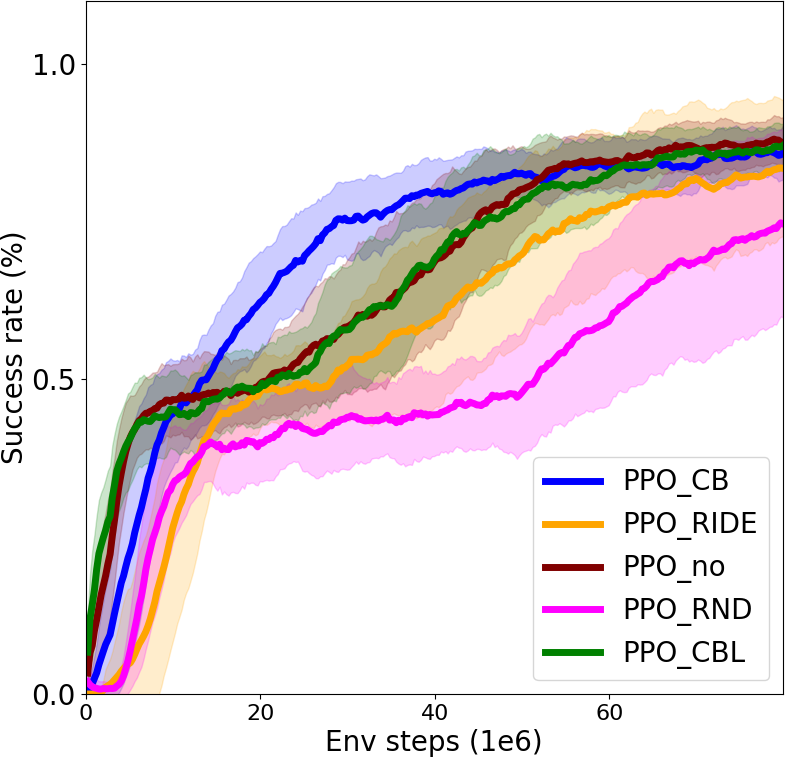}
\label{fig:pilot_pointing}
}
\hfill
\subfloat[\footnotesize Pilot experiments with the peer uttering the color of the correct object.]{
\includegraphics[width=0.40\textwidth]{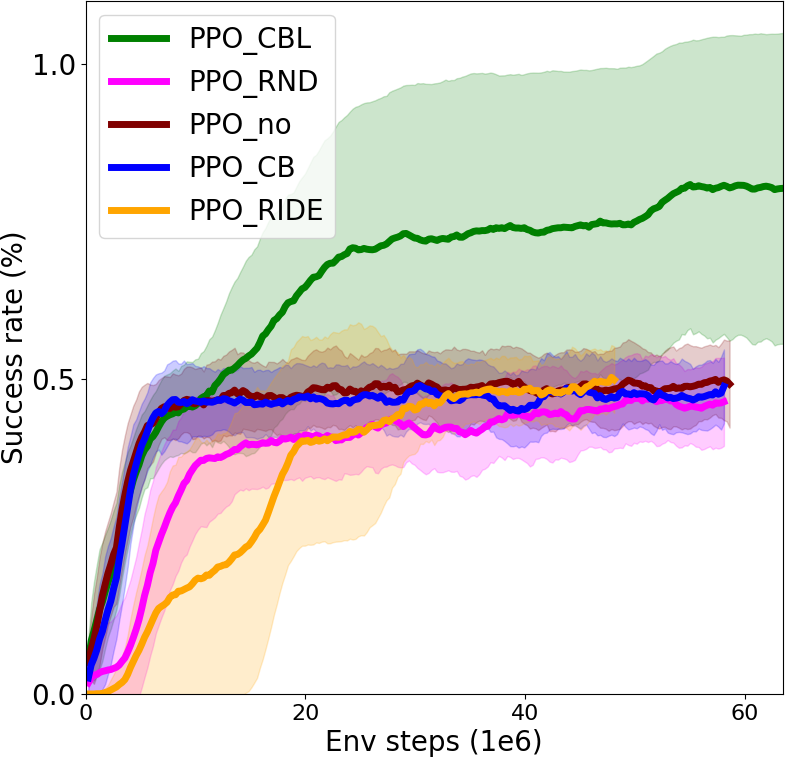}
\label{fig:pilot_color}
}
\caption{\footnotesize Pilot experiments showing that our count-based exploration bonuses outperform other baselines.
On the environments with the pointing gesture, visual count-based ("CB") exploration bonus is the best performing condition. On the environments with utterances, linguistic count-based ("CBL") exploration bonus is the best performing condition.}
\label{fig:exp_pilot}
\end{figure*}

Figure \ref{fig:exp_pilot} compares PPO agents trained with different exploration bonuses discussed in section \ref{app:exploration_bonuses} on two different \mytextsc{InformationSeeking} type environments.
The first environment involves the peer pointing to the correct object.
Figure \ref{fig:pilot_color} shows that the best performing agent is the one levering the visual count-based exploration bonus (PPO\_CB).
The second environment involves the peer uttering the color of the correct object.
Figure \ref{fig:pilot_pointing} shows that the best performing agent is the one levering the linguistic count-based exploration bonus (PPO\_CBL).
We conclude that PPO\_CBL is the most suitable baseline for environments involving linguistic cues, and PPO\_CB for the other environments. 

\section{Adversarial environment type}
\label{app:adversarial_type}
\begin{figure*}[htb!]
\centering
\subfloat[\footnotesize Adversarial peer environment without occlusions.]{
\includegraphics[width=0.40\textwidth]{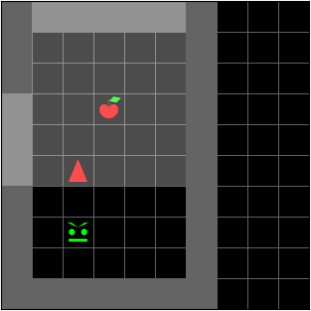}
\label{fig:env_adversarial}
}
\hfill
\subfloat[\footnotesize Adversarial peer environment with occlusions.]{
\includegraphics[width=0.40\textwidth]{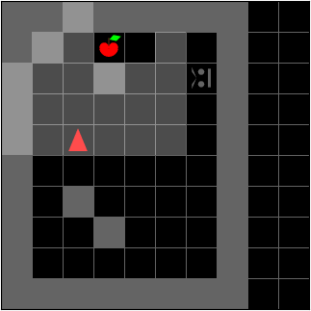}
\label{fig:env_adversarial_stumps}
}
\caption{\footnotesize Environments from the Adversarial peer experiments in which the agent has to infer the peer's field of view. The agent is rewarded upon eating the apple on the condition that it was not in the field of view of the peer while doing so. We run the experiments with two different settings: with and without occlusions (depicted in figures \ref{fig:env_adversarial_stumps} and \ref{fig:env_adversarial}). Occlusions make it harder to infer the peer's field of view as it is no longer rectangular.}
\label{fig:adversarial_env}
\end{figure*}

In the main text we discussed two environment types: \mytextsc{InformationSeeking} and \mytextsc{Collaboration}. 
In this section we explain an additional environment type - \mytextsc{Adversarial} type.
This environment type is used to study the ability of the agent to infer the peer's field of view.
An apple will be present in the environment right away. However, the agent will get rewarded only if it eats it while not being observed by the peer (the peer is adversarial).
Therefore, the agent needs to infer the right moment to eat the apple. 
There is one important parameter in this environment type. It refers to the amount of obstacles present in the environment.
Figure \ref{fig:adversarial_env} shows this environment type without any obstacles (figure \ref{fig:env_adversarial}) and with obstacles present (figure \ref{fig:env_adversarial_stumps}).

\section{Details on the parameters used}
\subsection{The Pointing experiment parameters}
\label{app:point_exp_params}

\begin{figure*}[htb!]
\centering
\subfloat[\footnotesize Training sampling tree]{
\includegraphics[height=0.15\textheight]{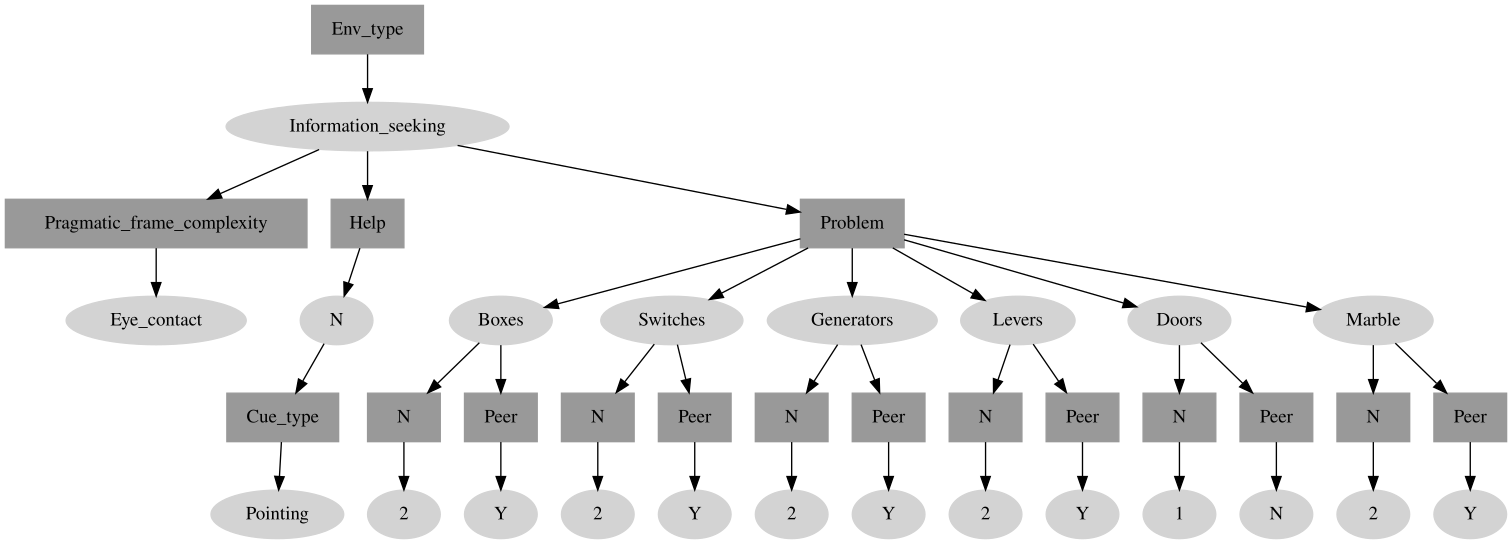}
\label{fig:pointing_tree_train}
}
\hfill
\subfloat[\footnotesize Testing sampling tree - Social Doors]{
\includegraphics[height=0.15\textheight]{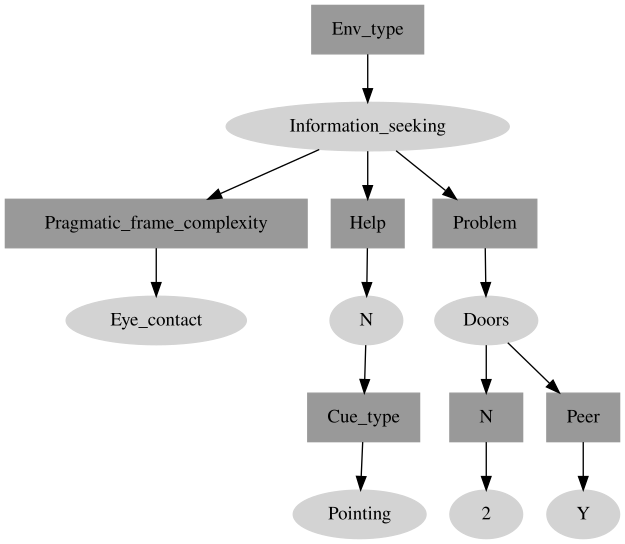}
\label{fig:pointing_tree_test}
}
\caption{\footnotesize Sampling trees used in the pointing case study in section \ref{sec:exp_pointing}}
\label{fig:pointing_tree}
\end{figure*}

The parameter trees used in this experiment are depicted in figure \ref{fig:pointing_tree}.
We used the \mytextsc{InformationSeeking} environment type described in section \ref{sec:env_types}.
The \mytextsc{Introductory\_sequence} is set to \mytextsc{Eye\_Contact}, and the \mytextsc{Cue\_Type} to \mytextsc{Pointing} - the peer will point to the correct object after eye contact.
The agent is trained on the following five problems: \mytextsc{Boxes}, \mytextsc{Switches}, \mytextsc{Levers}, \mytextsc{Marble}, \mytextsc{Generators}, and on the asocial version of the \mytextsc{Doors} problem (a version without the distractor or peer).
Training on this asocial version is important as it enables the agent to learn how to use a door, which is needed to evaluate generalization.

\subsection{Role reversal imitation parameters}
\label{app:role_reversal_exp_params}
\begin{figure*}[htb!]
\hfill
\subfloat[\footnotesize Role A]{
    \includegraphics[height=0.18\textheight]{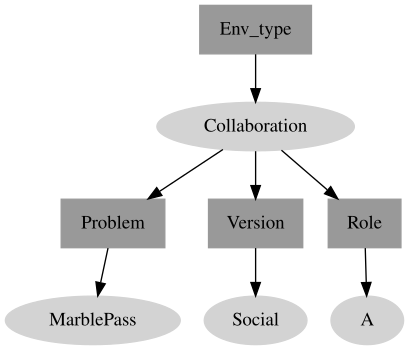}
    \label{fig:rr_tree_A}
} 
\hfill
\subfloat[\footnotesize Asocial single setting]{
    \includegraphics[height=0.18\textheight]{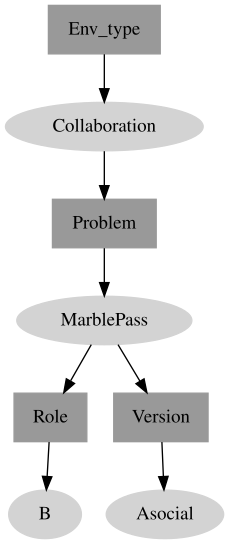}
    \label{fig:rr_tree_asoc_single}
}
\hfill
\subfloat[\footnotesize Role B single setting]{
    \includegraphics[height=0.18\textheight]{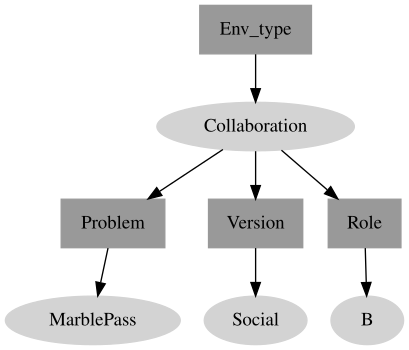}
    \label{fig:rr_tree_B_single}
}
\\
\centering
\subfloat[\footnotesize Asocial group setting]{
    \includegraphics[height=0.18\textheight]{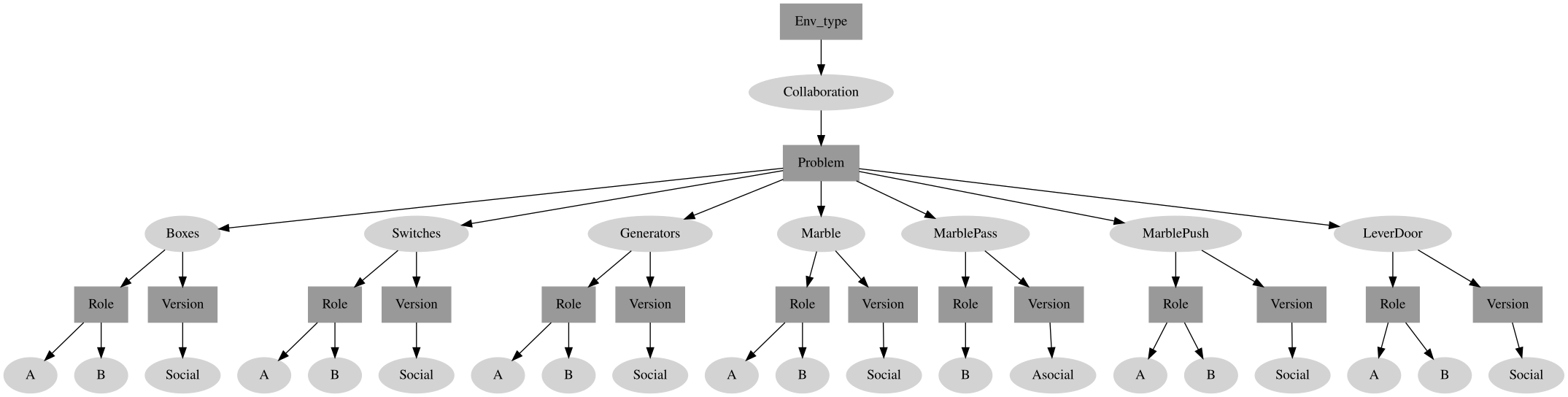}
    \label{fig:rr_tree_asoc_group}
}
\\
\centering
\subfloat[\footnotesize Role B group setting]{
    \includegraphics[height=0.18\textheight]{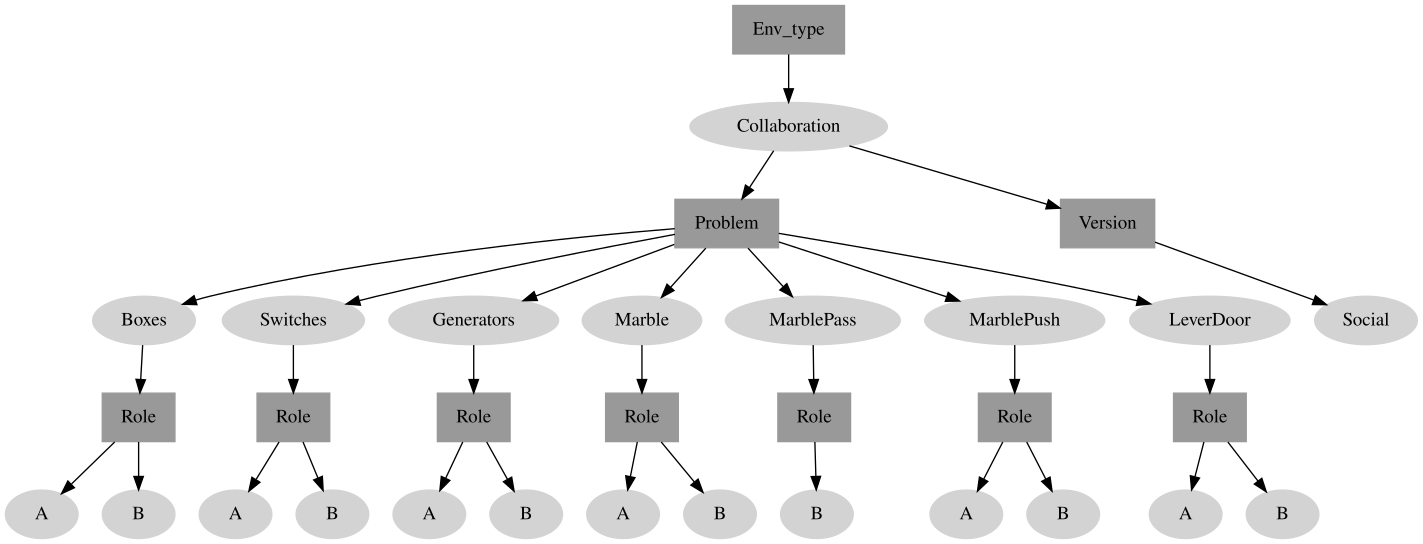}
    \label{fig:rr_tree_B_group}
}
\caption{\footnotesize Role reversal sampling trees from the case study in section \ref{sec:exp_role_reversal_imitation}
}
\label{fig:rr_tree}
\end{figure*}

The parameter trees used in this experiment are depicted in figure \ref{fig:rr_tree}.
We used the \mytextsc{Collaboration} type environments described in section \ref{sec:env_types}.
We evaluate agents on role A of the \mytextsc{MarblePass} task - the agent has to push the marble to the right side of the environment, from where the peer can push it to the \textit{marble generator}.

\subsection{Scaffolding parameters}
\label{app:scaffolding_exp_params}
\begin{figure*}[htb!]
\hfill
\subfloat[\footnotesize Testing tree.]{
\includegraphics[width=\textwidth]{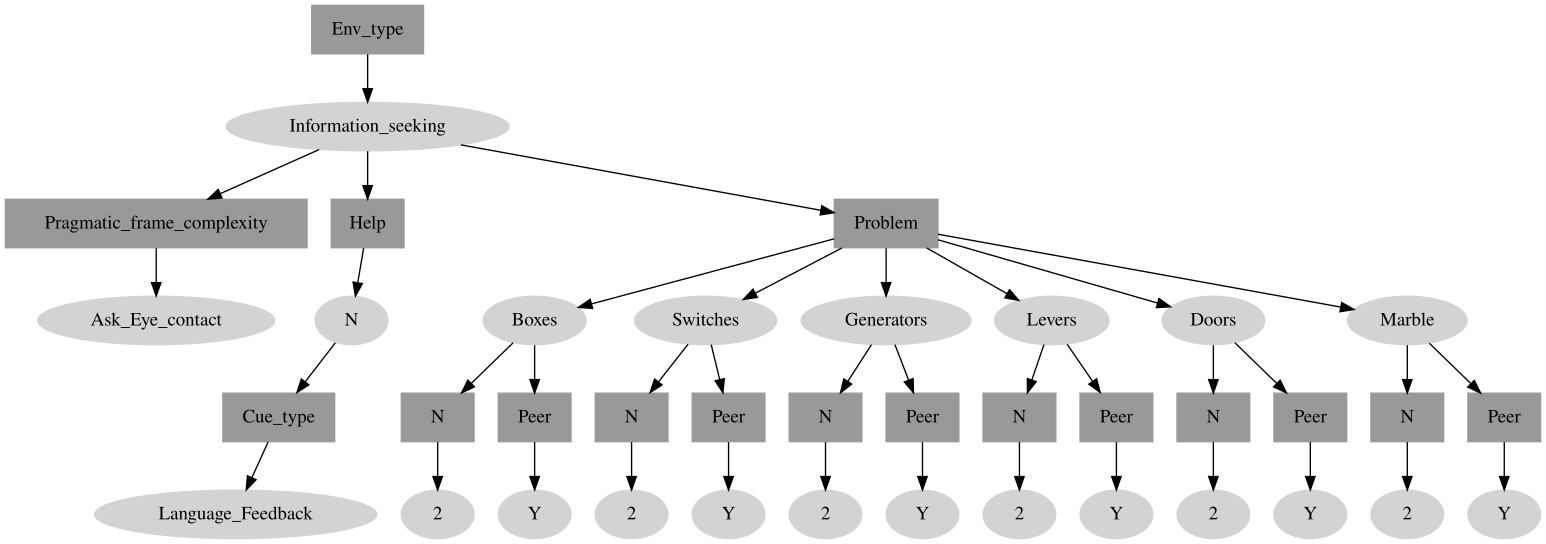}
\label{fig:scaf_tree_test}
}
\hfill
\subfloat[\footnotesize Scaf\_4 tree.]{
\includegraphics[width=\textwidth]{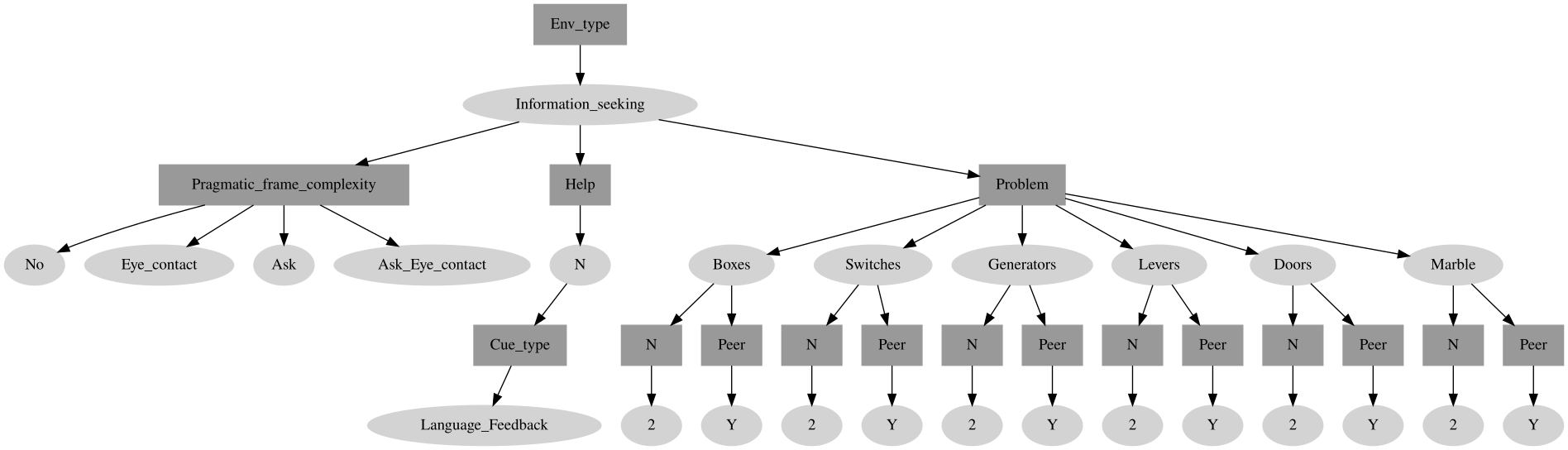}
\label{fig:scaf_tree_4}
}
\hfill
\subfloat[\footnotesize Scaf\_8 tree.]{
\includegraphics[width=\textwidth]{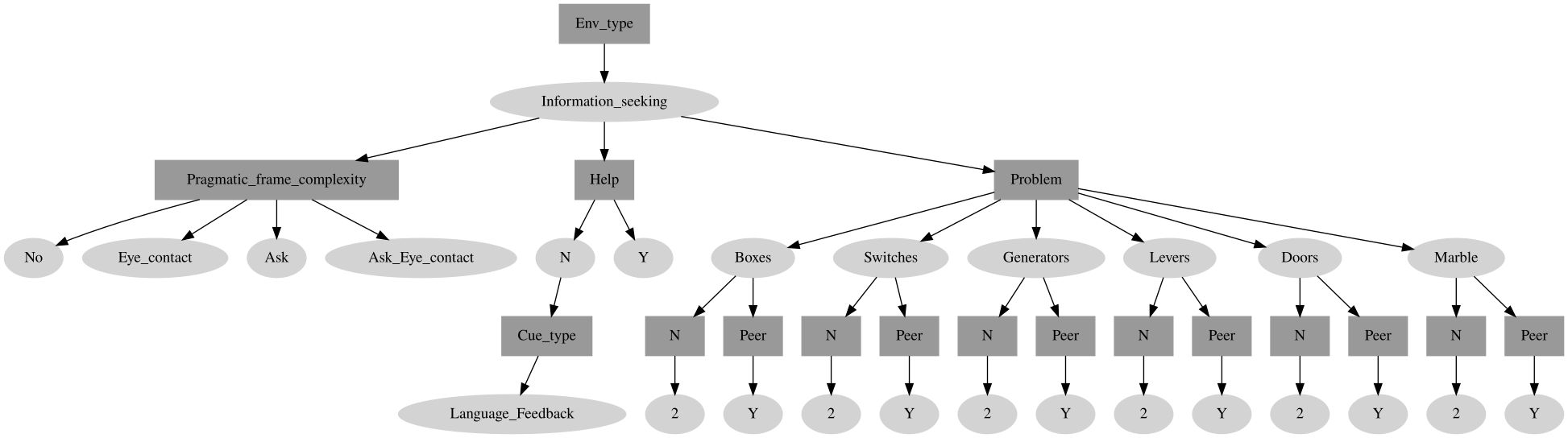}
\label{fig:scaf_tree_8}
}
\hfill
\caption{\footnotesize Sampling trees used in the first phase of the scaffolding case study in section \ref{sec:exp_scaffolding}}
\label{fig:scaf_tree}
\end{figure*}

The parameter trees used in this experiment are depicted in figure \ref{fig:scaf_tree}.
In this experiment, we use the \mytextsc{Information seeking} environment type with the \mytextsc{language feedback} cue type.
We train agents on all six problems, using different values of the \mytextsc{Introductory\_sequence} and \mytextsc{Help} parameters.
We evaluate the agents on all six problems, with the most complex introductory sequence - \mytextsc{Ask\_Eye\_contact}.

The agent denoted by "scaf\_4" is trained on four different values of the \mytextsc{introductory sequence} parameter, and with the \mytextsc{Help} parameter set to \mytextsc{N} (the peer will provide cues).
This agent will be trained on a total of 18 different environments: six problems, and four introductory sequences.
The second agent (denoted by "scaf\_8") is also trained on all values of the \mytextsc{Introductory\_sequence} parameter, but it is in addition trained on both values of the \mytextsc{Help} parameter (\mytextsc{N} and \mytextsc{Y}) - a total of 36 environments.
In half of those environments (with \mytextsc{Help} set to \mytextsc{Y}) the peer will provide the apple to the agent after the introduction (e.g. it will go to the correct box, and open it). 
In the other half (with \mytextsc{Help} set to \mytextsc{N}), the peer will only provide linguistic feedback cues.

In this experiment, we use the PPO agent without an exploration bonus.

\section{Additional case studies}
\label{app:additional-cstudies}

\subsection{Inferring the meaning of linguistic cues}
\label{sec:exp_language}

In this section, we study the ability of the agent to infer the meaning of simple words.
We follow the same procedure as in section \ref{sec:exp_pointing}.
This case study is motivated by the experiments from cognitive science discussed in section \ref{sec:cog_sci_background_MT_comm}.
In \shortcite{carpenter1998_monographs} infants' word understanding steadily increased in the period between 9 and 15 months after birth.
We study the following questions:
\begin{itemize}
\item Can an RL agent learn to interpret simple utterances?
\item Can the agent generalize to new situations, and infer the meaning of those utterances for objects in a new context?
\end{itemize}
The best performing agent on the linguistic environments in the pilot study was the one using the linguistic count-based exploration bonus (PPO-CBL) (see appendix \ref{app:pilot}).
We use this agent to address both questions.

\paragraph{Environments}
The environments are the same as those in section \ref{sec:exp_pointing}: the \mytextsc{Information\_seeking} environment type, with the \mytextsc{introductory\_sequence} set to \mytextsc{Eye\_contact}.
The only difference is that the peer will give linguistic cues instead of pointing.
We run two experiments with two different types of linguistic cues: \textit{Color} and \textit{Feedback}.
In \textit{Color} the peer will utter name color of the correct object.
In \textit{Feedback} the peer will utter a description of how close the agent is to the correct object: "Cold", "Medium", "Warm", and "Hot" meaning, respectively, "far", "medium", "close" and "right next to".
The experimental procedure is the same as the one in section \ref{sec:exp_pointing}.
The agent is trained on the same five problems and the asocial version of the \mytextsc{Doors} problem.

\textbf{Can RL agents learn to interpret simple utterances?}

\begin{figure*}[htb]
\subfloat[\footnotesize Language Feedback cue type experiments: the peer gives cues regarding the proximity of the agent is to the correct object (e.g. Hot, Warm, Cold).]{
\includegraphics[width=0.40\textwidth]{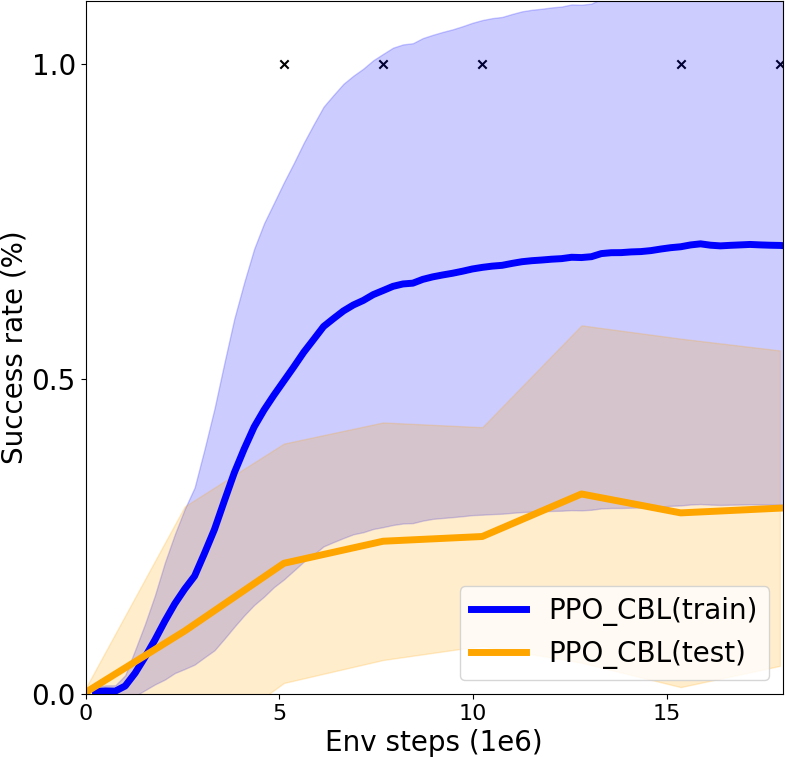}
\label{fig:feedback_exp}
}
\hfill
\subfloat[\footnotesize Language Color cue type experiments: the peer utters the color of the correct object.]{
\includegraphics[width=0.40\textwidth]{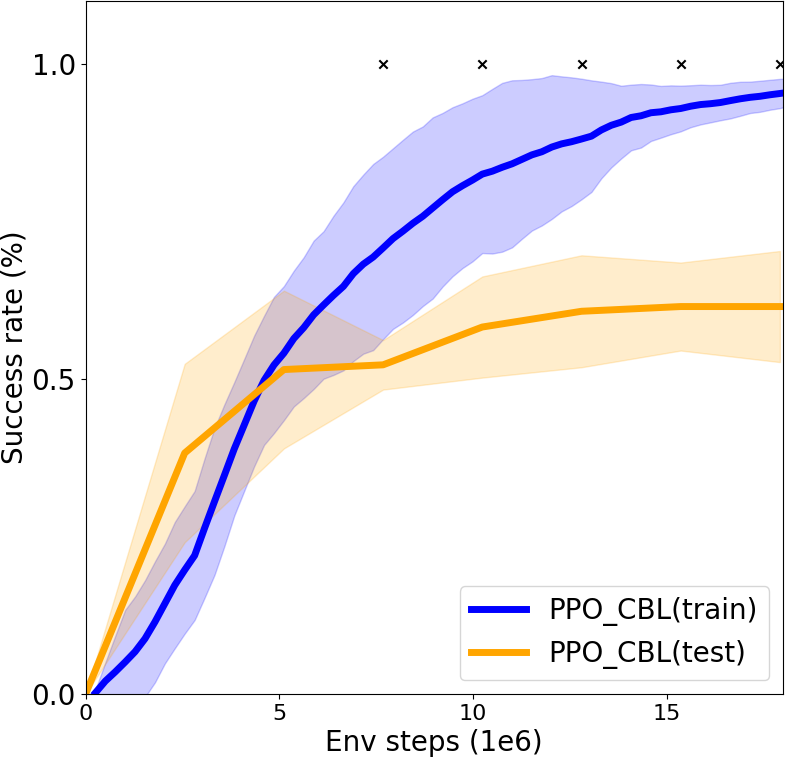}
\label{fig:color_exp}
}
\caption{\footnotesize \textbf{The linguistic cues experiments.} We study if an RL agent is able to infer the meaning of linguistic cues in order to use the correct object.
We consider two types of cues: \textit{language feedback} and \textit{color}. 
In both settings, the agent was trained on five different problems, and on the asocial version of the Doors problem (only one door and no peer present in the environment) - denoted by "train".
Agents were periodically evaluated on the social version of the Doors problem (two doors and a peer giving cues) - denoted by "test".
The figure compares the success rate (mean +/- std over 8 seeds) on the training environments with the evaluation on the testing environment.
The cross marks depict statistical significance ($p=0.05$).
In both cases the agents achieve much better performance on the training problems, but fail to generalize to a new problem - the agent is not able to infer the meaning of an utterance in a new context.
}
\label{fig:pointing_feedback_color_exp}
\end{figure*}

\begin{figure*}[htb]
\centering 
\includegraphics[width=0.45\textwidth]{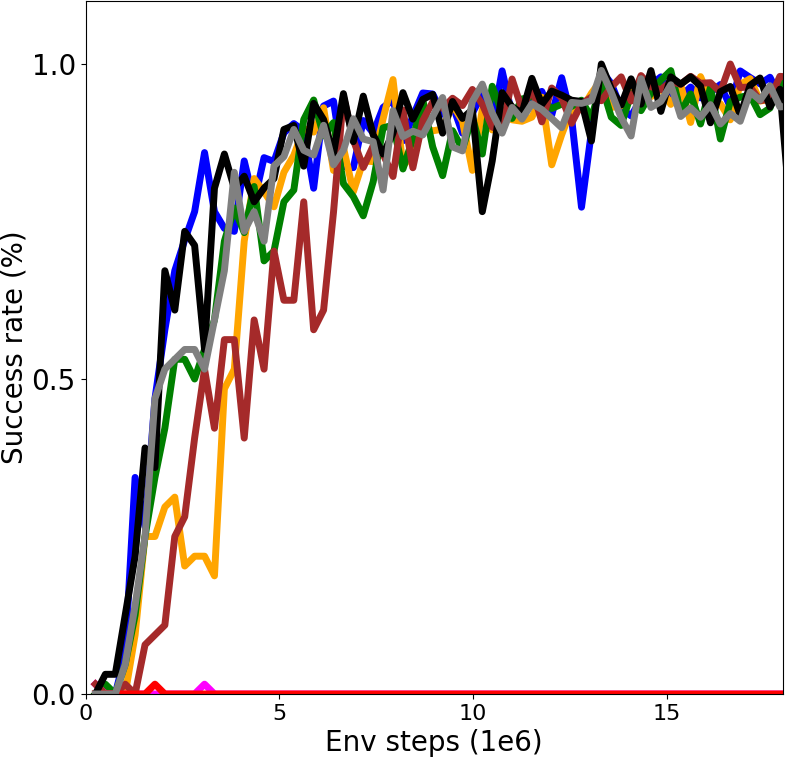}
\caption{\footnotesize Per-seed performance on the training environments of the agent from figure \ref{fig:feedback_exp} ("PPO\_CBL(train)").
The figure shows that the agent is able to solve the training tasks efficiently, but that there are two unstable seeds which result in the success rate of 0\%.}
\label{fig:feedback_per_seed}
\end{figure*}

Figures \ref{fig:feedback_exp} and \ref{fig:color_exp} show the performance of the agent with the linguistic count-based exploration bonus (denoted PPO\_CBL\_train).
We can see that the agent (PPO-CBL) solves these environments efficiently, reaching a final performance of $95.9\%$ and $71.4\%$ for \mytextsc{Color} and \mytextsc{Feedback} cue types, respectively.
We further analyse the performance of each separate seed for the agent trained on the \mytextsc{Feedback} cue type. This is shown in figure \ref{fig:feedback_per_seed} where it is visible that the agent is normally able to achieve high performance, but that there are two seeds which, due to their instability, reach a success rate of $0$.
This experiment shows that the agent is capable of learning to infer the meaning of simple utterances in familiar contexts.

\textbf{Can the agent generalize to new situations?}
A more interesting question is whether that agent can infer the meaning of the same word based on a new context.
Therefore, we evaluate the agent's generalization abilities in a new scenario - the \mytextsc{door} problem - following the same procedure as in section \ref{sec:exp_pointing}.
This kind of generalization is particularly interesting as communication depends on our ability to ground words in \textit{new} social contexts: inferring meaning by combining the convention associated to a word with the recursively inferred intention of the speaker.
For example, while "red" can mean "open the red box" in one context, it can mean "push the marble towards the red generator" in another.

Figures \ref{fig:color_exp} and \ref{fig:feedback_exp} show the performance of the same agent evaluated on the \mytextsc{doors} problem (denoted "PPO\_CBL\_test")
They show that neither of the agents is capable of such generalization, which is consistent with the experiments with the pointing gesture in section \ref{sec:exp_pointing}.

These results motivate future research on what kind of biases could be built into the agents (and in what way) so that they could infer the meaning of familiar words in new contexts.
For example, an interesting avenue of future work is to try to combine an agent with a large language models, and see if the knowledge contained in it could make the agent generalize better.

\subsection{Joint Attention}
\label{sec:exp_ja}        

Tomasello describes joint attention as consisting of two parts: triangulation and recursiveness \shortcite{tomasello2019becoming}.
He argues that joint attention plays a key role in the 9-month revolution by transforming dyadic interactions (e.g. mimicking facial expressions) to triadic (e.g. imitating an action on an object).
Joint attention was also required in the previous experiments (sections \ref{sec:exp_pointing} and \ref{sec:exp_language}).
The agent and the peer triangulated on an external referent, however, the agent could assume that the peer was participating in the interaction.

In this experiment, we aim to conduct a more thorough test of the second aspect of joint attention - \textit{recursiveness} (both participants being aware that they are both sharing attention).
To solve the task, the agent needs to infer if the peer is participating and is aware that the agent is participating too.
We create environments where the peer, in addition to giving regular cues inside joint attention, gives \textit{misleading} cues outside joint attention. 
These cues are implemented uttering a random cue, and are given before the agent completes the introductory sequence.
In other words, the agent should learn to discriminate between cues given for the agent during joint attention (after the introduction) and cues given regardless of the agent outside joint attention  (before the introduction).

We study the following question:
\begin{itemize}
\item Can RL agents learn to differentiate between cues given inside and outside joint attention, i.e. can they learn to infer whether the peer is participating in the interaction?
\end{itemize}

\paragraph{Environments} 
In this section, we extend the environment from section \ref{sec:exp_language} studying the \textsc{Color} cue type.
The environment is extended so that the agent must also recursively infer whether the cue is intended for the agent.
This misleading cue is given before the introductory sequence is completed, and takes the form of the peer uttering a color of a random one of the two objects. 
Apart from that, the experiments are conducted in the same way as in section \ref{sec:exp_language} (we train on the same problems for and use the same baseline).

\begin{figure*}[htb!]
\centering
\includegraphics[width=0.4\textwidth]{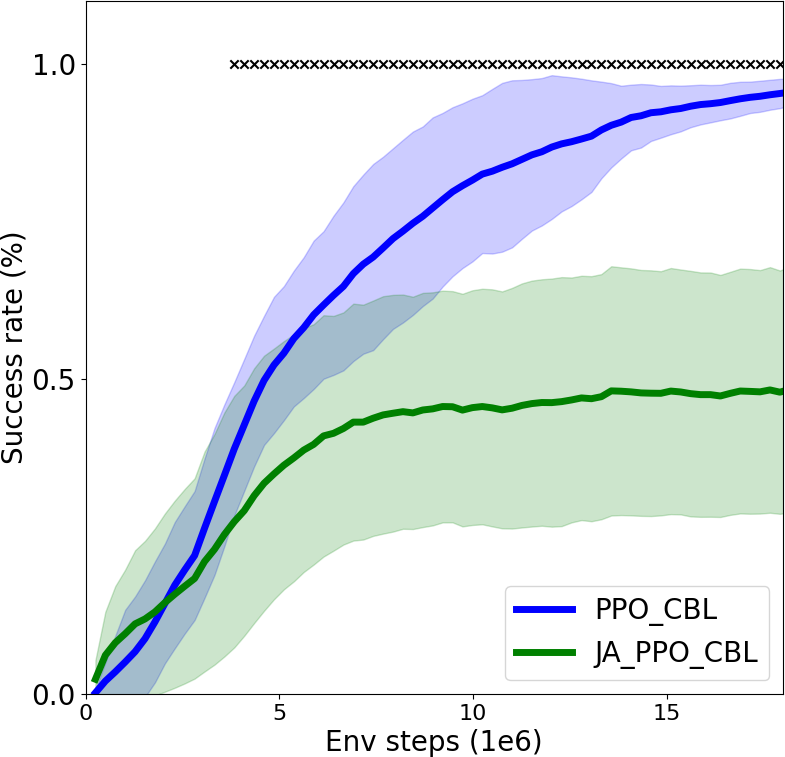}
\caption{\footnotesize \textbf{The joint attention experiment}. The environments feature a test for recursivness - infer if the peer knows that they are working together. The environments are same as the ones from figure \ref{fig:color_exp}, but with the addition of misleading cues - random cues given regardless of the agent (a random color).
The peer gives misleading cues outside of joint attention (before the introductory sequence). The agent should ignore these cues, and use only cues given inside joint attention.
The figure compares the success rate (mean +- std over 8 seeds) of the agent trained on the environments with both regular and misleading cues("JA\_PPO\_CBL"), to the agent trained on the environments with only regular cues ("PPO\_CBL(train)" from figure \ref{fig:color_exp}). 
The figure shows that the agent is unable to master the Joint Attention variant.
}
\label{fig:ja_exp}
\end{figure*}

\textbf{Results} Figure \ref{fig:ja_exp} compares the performance (success rate) of the agent trained on this extended environment (denoted by JA) with the agent trained on the regular environment (from the experiment in section \ref{sec:exp_language}).
These results show that the agent is not able to differentiate between cues given inside and outside of joint attention.
We believe that this is due to the cues being highly misleading in this environment. As the peer utters the color of a random object present in the environment 50\% of the a misleading cue will be the same as the helpful one. 

These results open many avenues for future research. 
One might study which kinds of biases can be integrated into the agent to make such cues less misleading. 
The generalization abilities of those agents should also be investigated.
For instance, we could study if an agent that learned to ignore misleading linguistic cues would ignore misleading pointing cues.

\subsection{Imitation learning}
\label{sec:exp_imitation}


In the following section, we study the ability of the agent to learn how to obtain the apple by imitating the peer.
This experiment is motivated by an experiment from \shortcite{carpenter1998_monographs} discussed in section \ref{sec:cog_sci_background_MT_soc_learning}.
In it, infants showed a steady increase in imitation learning abilities in the period between 9 and 15 months after birth.
We want to test the agent's ability to imitate an instrumental action on an object.

From an AI perspective, this can be seen as meta-imitation learning.
We want to see if an agent can obtain (through gradients) the imitation learning mechanism, which it could then use (during the episode) to learn how to use a new object.
In this section, we study the following question:
\begin{itemize}
\item Can RL agents learn (through gradients) an imitation mechanism? 
\end{itemize}
In these experiment, we use the agent with the visual count-based exploration bonus (CB), as we found it worked best in our pilot study (see appendix \ref{app:pilot}).
We compare three agents trained with the same exploration bonus scaled by different weights: 0.25, 0.5, and 1.

\textbf{Environment}
The Environment is an \mytextsc{Information seeking} type environment without a distractor.
After the introductory sequence (\mytextsc{Eye\_contact}), the peer will demonstrate using an object to obtain the apple.
For example, it will toggle a box or push a generator.
Then the peer will then eat the apple, and revert the environment to its initial state.
The agent should then imitate the peer - use the same action on the object - to obtain the apple for itself. 
If the agent uses the object it in the wrong way (e.g. pushes the box instead of toggling it) it will be blocked and the apple will not be obtainable in this episode.
The agents are evaluated on a new problem in which the agent encounters a new object for the first time.
This means that the agent must pay attention to how the peer uses the object, and use it in the same way. \footnote{The encoding of the peer includes the peer's previous timestep action.}

The agents are trained on five problems (all expect \mytextsc{Doors}).
Most importantly, compared to the experiments in sections \ref{sec:exp_pointing} and \ref{sec:exp_language}, these agents will not be trained on the asocial version of the \mytextsc{Doors} problem. 
That is because, in the generalization testing, we want to see if the agent can learn to use a completely new object.

\begin{figure*}[htb!]
\centering
\subfloat[\footnotesize Imitation experiments performance (success rate) on the training environments.]{
\includegraphics[width=0.30\textwidth]{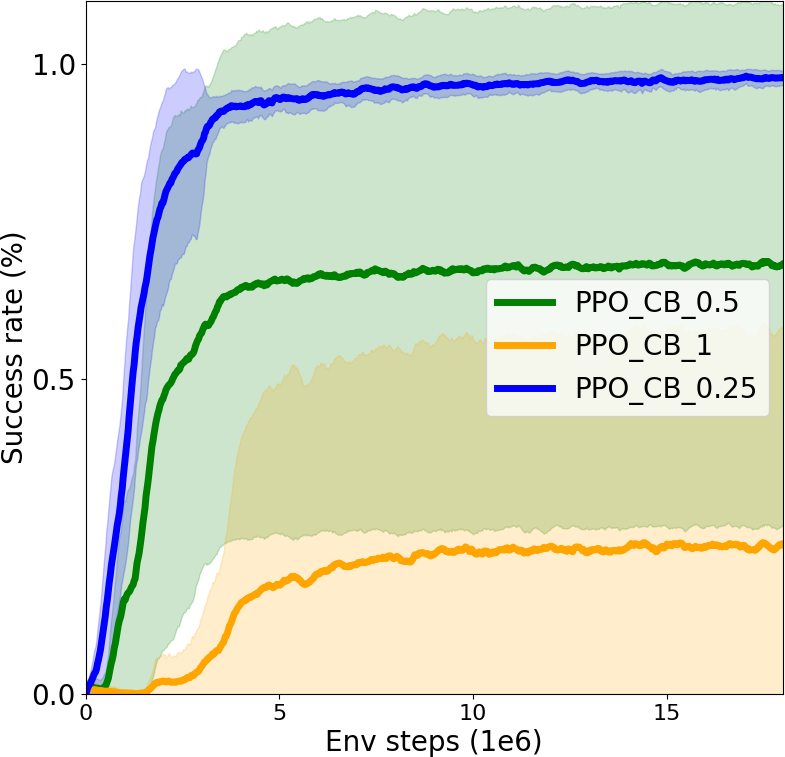}
\label{fig:imitation_train}
}
\hspace{1pt}
\subfloat[\footnotesize The percentage of successful introductory sequences on the training environments.]{
\includegraphics[width=0.30\textwidth]{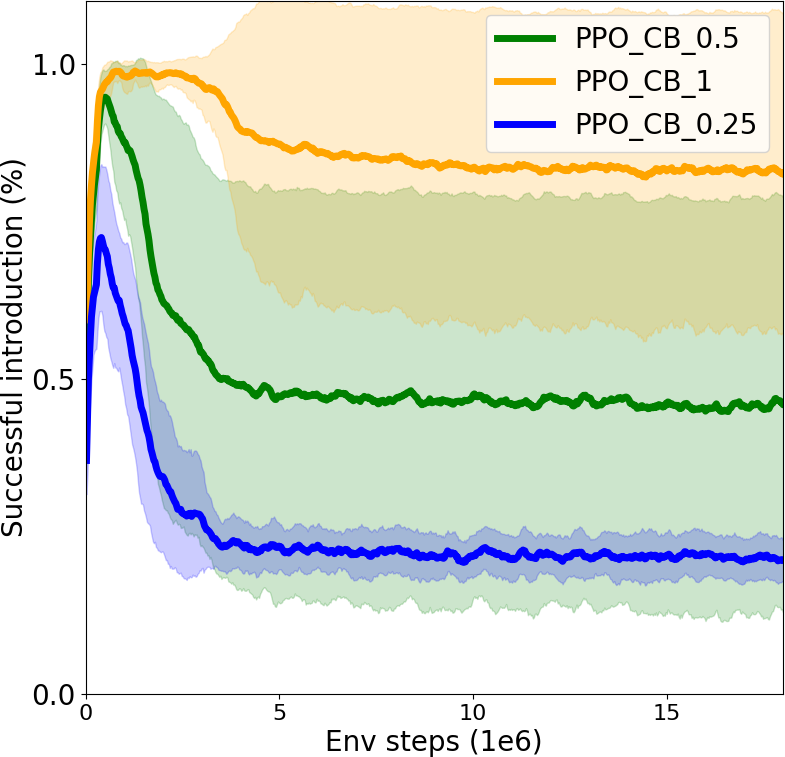}
\label{fig:imitation_train_intro}
}
\hspace{1pt}
\subfloat[\footnotesize Imitation experiments performance (success rate) on the testing environment.]{
\includegraphics[width=0.30\textwidth]{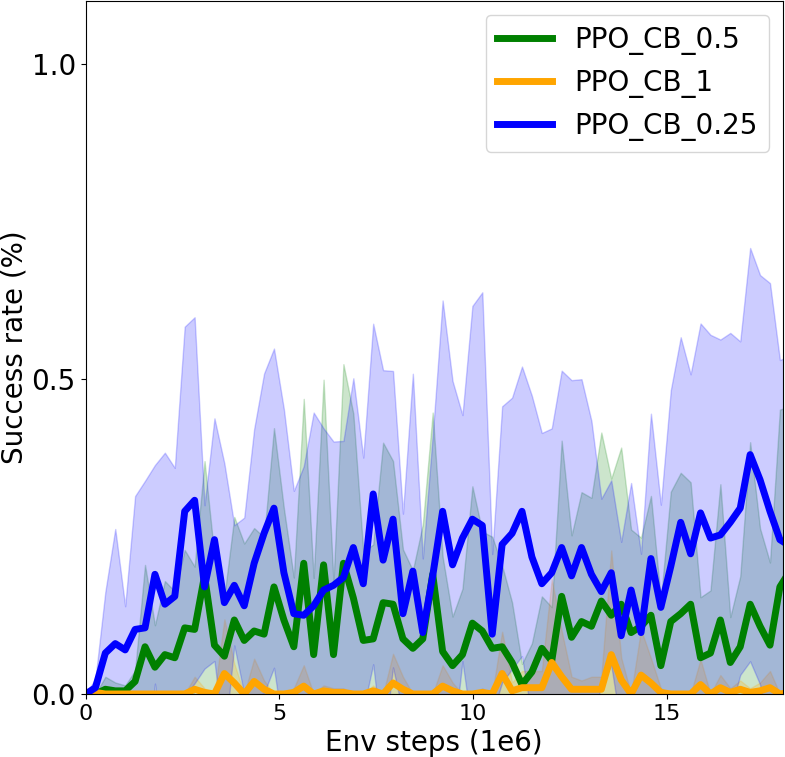}
\label{fig:imitation_test}
}
\caption{\footnotesize \textbf{Imitation learning experiments.}
The peer demonstrates how to use an object (after the agent succesfully introduces itself). The agent is trained on five different problems and evaluated on a new problem with a previously unobserved object (a door).
A socially proficient agent should be able to learn (by observing the demonstration) which action (toggle or push) to use on the new object.
The curves compare three agents trained with a different scaling factor for the visual count-based exploration bonus.
One can see that the agent with high exploration bonus ("PPO\_CB\_1") focuses too much on the peer, which results in ignoring the task. This is evidenced by high success in completing the introductory sequence (fig. \ref{fig:imitation_train_intro}), but low success rate on the task (fig. \ref{fig:imitation_train}).
On the other hand, using low exploration bonus ("PPO\_CB\_0.25") pushes the agent to solve the training task whilst ignoring the peer. Rather than observing the peer's demonstration, this agent learns how to use objects by themselves. This results in perfect performance on the training object, but it makes it impossible to generalize to a new object.
Neither of the agents is able to achieve high performance on the heldout testing environment. This implies that they are not able to learn (online) through imitation which action to use with a new object.}
\label{fig:imitation_exp}
\end{figure*}

\textbf{Results}
Figure \ref{fig:imitation_exp} shows the performance (success rate) of the agents on the training environments,  the percentage of succesful introduction with the peer, and the evaluation on the (unseen) \mytextsc{Doors} problem.

On figures \ref{fig:imitation_train} and \ref{fig:imitation_train_intro} we can see that the agent with a lot of exploration bonus (PPO\_CB\_1) is too focused on the peer and, and is unable to solve the task. This is implied by the high percentage of the successful introductory sequence, and low success rate on the training environments.
On the other hand the agent with smaller exploration bonus weight (PPO\_CB\_0.25) solves these environments without problems, however it does use the peer.
As such the agent can solve the training environments by ignoring the peer and discovering how to use each object by itself.
However, this agent is not able to generalize to a new object as the only way to know how to use that object is to observe the peer's demonstration (see figure \ref{fig:imitation_test}).
Figure \ref{fig:imitation_test} shows the performance of those agents on the testing environment.
The figure shows that neither of the three agents is capable of acquiring an meta-imitation learning mechanism that can generalize to a novel object.

These results are not surprising, as current exploration bonuses are not well suited to enable RL agents to meta-learn mechanisms.
These results imply that an interesting avenue of research is to study how to endow agents with such meta-imitation learning mechanisms that would enable them to learn a behavior in a new scenario.
An promising solution to this problem are large language models and other large transformer-based networks pretrained on many other tasks. It would be interesting to study if such agents already have an imitation learning mechanism which would enable such online imitation. This would open up countless avenues of research into various forms of online imitation and emulation learning.

\subsection{Inferring another's field of view}
\label{sec:exp_adversarial}

In this section, we study the ability of the agent to infer what the other observes.
This experiment is modeling the one in \shortciteA{hare2001chimpanzees}.
In it, apes were shown to be able to infer what another sees, as they only took the food the alpha male could not see.

In this section, we want to study the following question:
\begin{itemize}
\item Can agents learn to infer the other's field of view?
\end{itemize}

\textbf{Environment} In the following experiment, we are using the \mytextsc{AdversarialPeer} environment type, in which the agent has to eat the apple while not being seen by the peer. We study two version of this environment: with and without obstacles (for more details, refer to appendix \ref{app:adversarial_type}).
Obstacles make the problem of inferring the peer's field of view harder.

\textbf{Experiment}
We study how the agent infers the peer's field of view by training the agent on the AdversarialPeer task.
It is important to note that this agent can sometimes use other (asocial) information to achieve performance. 
For example, if the object is surrounded by occlusions the agent could guess that it is not observed by the peer, which is not necessarily the case. 
To better understand the performance of the agent we compare the agent with two baselines.
First, we assess to what extent the agent is making inferences based on the peer's location and gaze direction.
We train an agent ("invisible\_peer" ) that has the peer filtered from the its observations (it cannot observe the peer). This baselines estimates the maximum possible performance 
If the standard agent outperforms this baseline this implies that it is leveraging the social information in the environment. 
Second, to estimate the upper bound on the performance we train an agent in the environment without the peer present (this agent is reward every time it eats the apple).

\begin{figure*}[htb!]
\centering
\subfloat[\footnotesize No occlusions.]{
\includegraphics[width=0.4\textwidth]{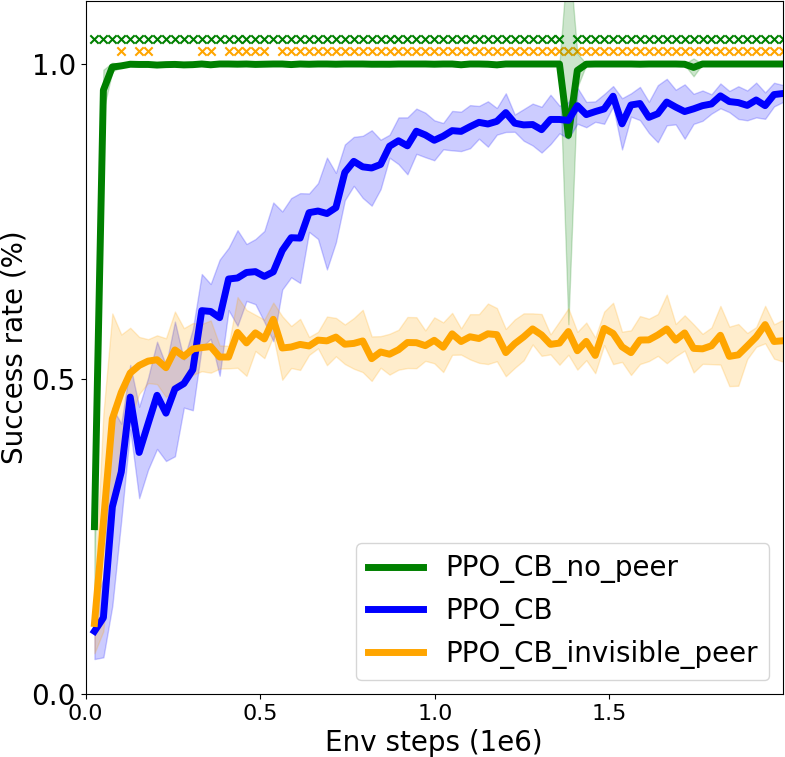}
\label{fig:adversarial}
}
\hfill
\subfloat[\footnotesize Occlusions]{
\includegraphics[width=0.4\textwidth]{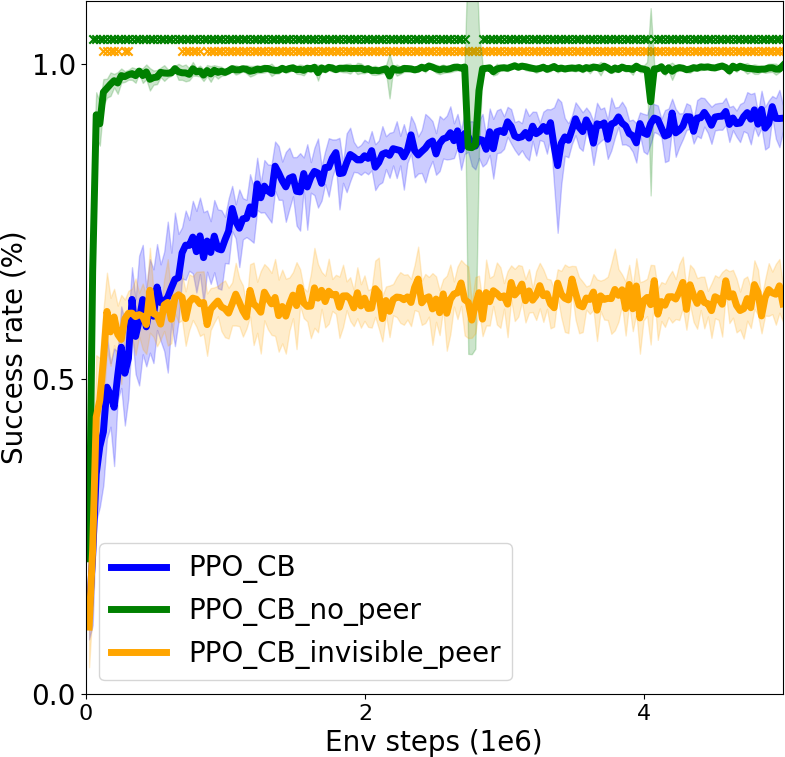}
\label{fig:adversarial_stumps}
}
\caption{\footnotesize Adversarial peer experiments. We compare three agents on two environments (depicted on figures \ref{fig:env_adversarial_stumps} and \ref{fig:env_adversarial}).
The "PPO\_CB" agent is trained on the regular environment (rewarded upon eating the apple while not being observed by the peer).
The "PPO\_CB\_no\_peer" agent is trained in the environment without the peer (the agent is rewarded every time it eats the apple). This represents the upper bound of the performance.
The "PPO\_CB\_invisible\_peer" agent is trained on the regular environment with the peer filtered from the agent's observations. This represents the performance of a completely asocial agent which ignores the peer.
Figures \ref{fig:adversarial} and \ref{fig:adversarial_stumps} compare the performance of these three agents (8 seeds +- std), the crosses depict a statistically significant difference (p<0.05) compared to the "PPO\_CB" agent.
The results show that the "PPO\_CB" agent is able to partially infer the peer's field of view (as it outperforms the "invisible\_peer" baseline), but is not able to reach perfect performance (as defined by the "PPO\_CB\_no\_peer" baseline).
}
\label{fig:adversarial_exp}
\end{figure*}

\textbf{Results}
Figure \ref{fig:adversarial_exp} shows the performances of those three agents.
It shows that the agent outperforms the agent with the peer filtered from its observations ("invisible\_peer"), which implies that the agent is using the peer's location and gaze direction to infer weather to eat the apple or not.
Furthermore, the agent is not able to match performance of the agent trained without the peer present in the environment ("no\_peer"). 
This results imply that, while the agent is able to leverage some social information in the environment, there still remains room for improvement.
Future research could focus on constructing novel types of exploration bonuses to bridge this gap.

\subsection{Formats}
\label{sec:exp_formats}
In the following experiment, we study the ability of the agent to learn formats (also referred to as pragmatic frames in \shortcite{vollmer_2016}).
Formats are a concept introduced by Jerome Bruner, which we discussed in more detail in section \ref{sec:cog_sci_background_JB}.
They can be regarded as protocols of social interactions.
We study the following question:
\begin{itemize}
\item To what extend can an exploration bonus help with the acquisition of a complex format.
\end{itemize}
We address this question by training two agents (one with an exploration bonus, and one without it).

\textbf{Environment} We use the \mytextsc{Information seeking} environment type with the \mytextsc{language feedback} cue type.
We train all agents on all six problems.
In contrast to section \ref{sec:exp_language}, where the introductory sequence was always set to \mytextsc{eye contact}, here it is set to \mytextsc{Ask eye contact} - the peer will give cues after the agent utters "Help, please" during eye contact.

\begin{figure*}[htb!]
\includegraphics[width=0.4\textwidth]{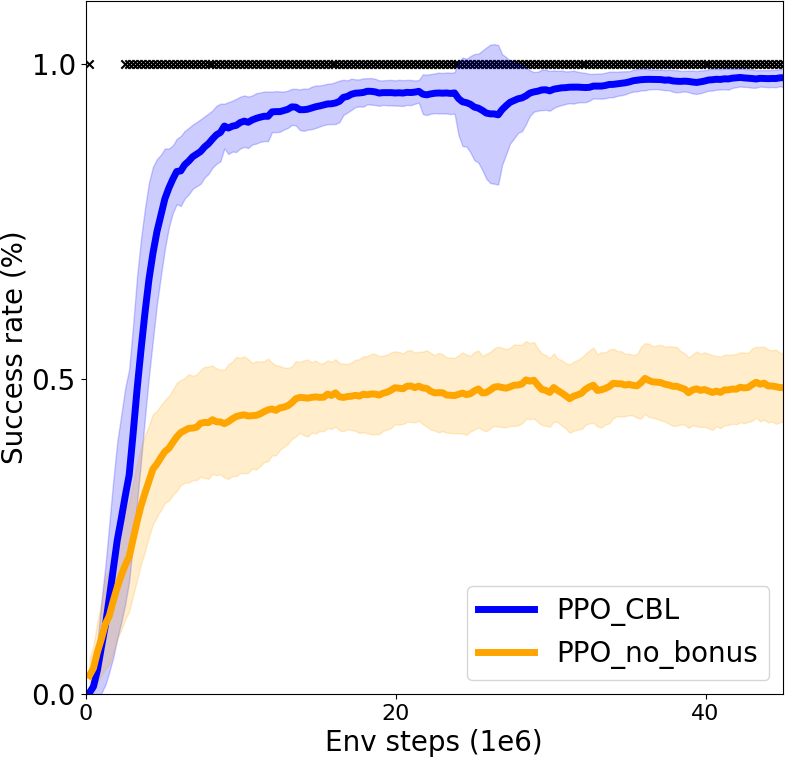}
\centering
\caption{\footnotesize Comparison of an agent with and without the exploration bonus on an environment with a more complex introductory sequence (format).
The task consists of the agent doing the introductory sequence by making eye contact and uttering "Help, please". The peer will then give linguistic cues regarding the proximity of the agent to the target object (e.g. Hot, Warm, Cold). Based on these cues, the agent should use the target object, instead of the distractor, to obtain the apple.
The figure shows that using the visual count-based exploration bonus enables the agent to learn a more complex introductory sequence and solve the task.}
\label{fig:formats_exp}
\end{figure*}

\textbf{Results} Figure \ref{fig:formats_exp} compares the performance of an agent that does not use any exploration bonus ("PPO\_no\_bonus") to an agent that uses the visual count-based exploration bonus ("PPO\_CBL"). The agent with the exploration bonus achieves high performance (97.9\% success rate) and greatly outperforms the agent without the exploration bonus.
These experiments show that, as expected, learning complex formats can be made easier with exploration bonuses.

This experiment can be interpreted in tandem with the experiment in section \ref{sec:exp_scaffolding} where we show how more complex formats can be learned by weaker agents (without an exploration bonus) when learning in a scaffolded environment.
Future work could explore how these two different approaches - modifying the agent and modifying the environment - can be used in tandem to learn even more complex formats.
Furthermore, one interesting research direction is to study which kinds of problems are better addressed by modifying the agent and which by modifying the environment.


\begin{table}
\centering
\caption{\footnotesize Template-based grammar used in all of the SocialAI environments. If the agent decided to speak it chooses a template and a noun to insert into the template.}
\centering
\begin{tabular}{@{}lll@{}}
\toprule
\textbf{Nouns} \\
\toprule
Action       & Template                & Noun          \\
\midrule    
0            & Where is <noun>         & please                \\
1            & Help <noun>             & the exit              \\
2            & Close <noun>            & the wall              \\
3            & How are <noun>          & you                   \\
4            &                         & the ceiling           \\
5            &                         & the window            \\
6            &                         & the entrance          \\
7            &                         & the closet            \\
8            &                         & the drawer            \\
9            &                         & the fridge            \\
10           &                         & the floor             \\
11           &                         & the lamp              \\
12           &                         & the trash can         \\
13           &                         & the chair             \\
14           &                         & the bed               \\
15           &                         & the sofa              \\
\bottomrule
\end{tabular}
\label{tab:grammar}
\end{table}

\begin{table}
\centering
\caption{\footnotesize Examples of  actions in the environment. Second and third dimension must both either be underfined or not. In practice, there is an additional binary output which defines if the agent will speak.}
\centering
\begin{tabular}{@{}lll@{}}
\toprule
Action & description \\ \midrule
(1, -, -) &  moves left without speaking   \\
(1, 1, 5) & moves left and utters "Help the window" \\
(-, 1, 5) & doesn't move but utters "Help the window" \\
(-, -, -) & nothing happens \\
\bottomrule
\end{tabular}
\label{tab:action_examples}
\end{table}

\begin{figure*}[ht]
\centering
\includegraphics[width=0.9\textwidth]{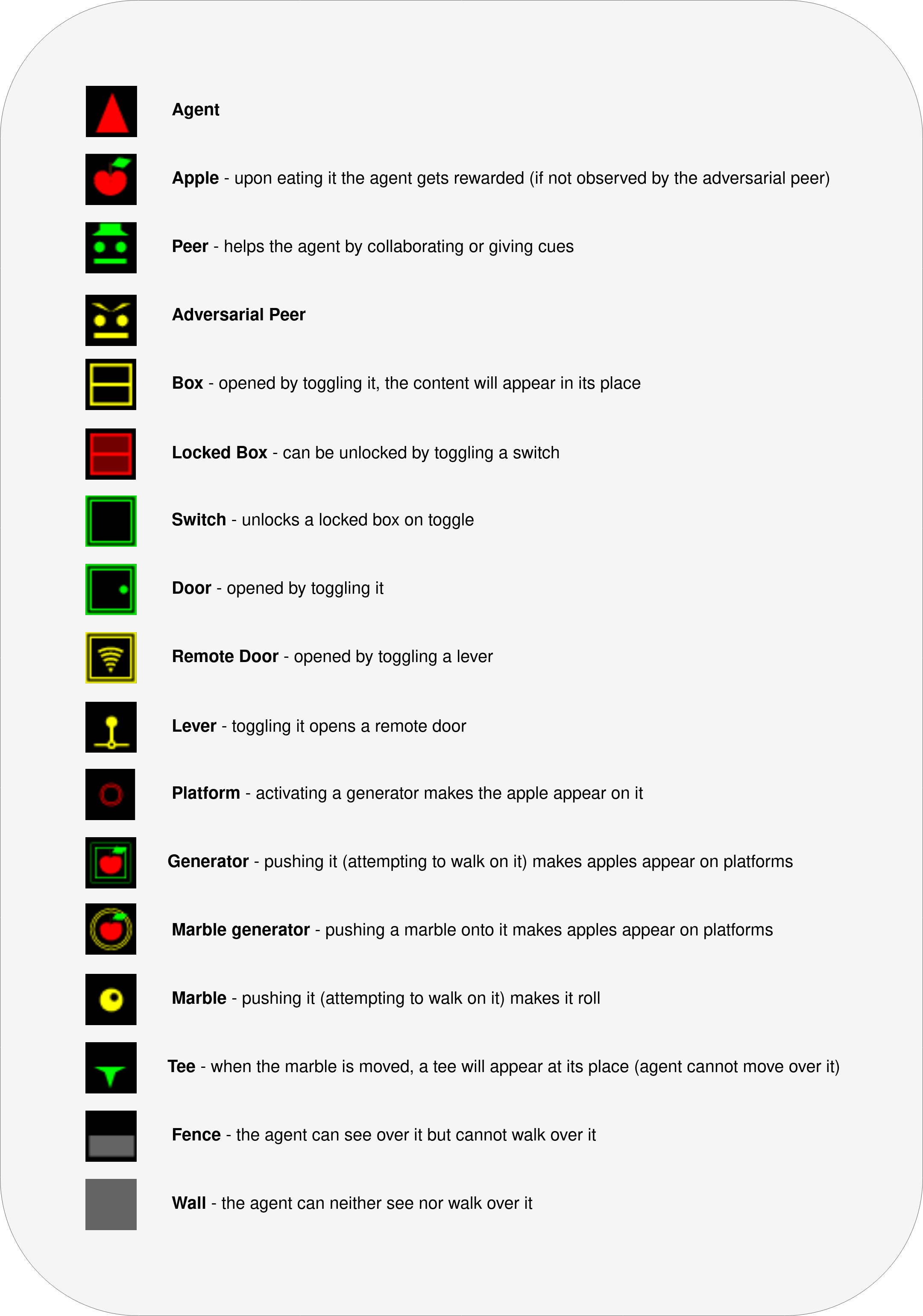}
\caption{\footnotesize Visualizations and descriptions for all objects featured in \socialai environments.}
\label{fig:objects}
\end{figure*}

\subsection{Additional information on the case study with large language models as interactive agents}
\label{app:additional_llm}

Figures \ref{fig:in_context_asocialapple}, \ref{fig:in_context_colorboxes}, and \ref{fig:in_context_colorboxes_gen} show the in-context examples provided to large language models in section \ref{sec:exp_llm} for the AsocialBox environment, ColorBoxes environment, and the generalization to the ColorBoxes environment, respectively. The in-context examples were created by hand.

\begin{figure}
\centering
\begin{minipage}[t]{0.48\textwidth}

\begin{lstlisting}
New episode.
Obs : 1 steps in front of you and 1 steps to the left there is a closed green lockablebox 
Act : move forward
Obs : Just to the left of you there is a closed green lockablebox 
Act : turn left
Obs : Right in front of you  there is a closed green lockablebox 
Act : toggle
Obs : Right in front of you  there is a red apple 
Act : toggle
Obs : Right in front of you  there is a yellow apple 
Success!
New episode.
Obs : 1 steps in front of you and 1 steps to the right there is a closed blue lockablebox 
Act : move forward
Obs : Just to the right of you there is a closed blue lockablebox 
Act : turn right
Obs : Right in front of you  there is a closed blue lockablebox 
Act : toggle
Obs : Right in front of you  there is a red apple 
Act : toggle
Obs : Right in front of you  there is a yellow apple 
Success!
New episode.
Obs : 3 steps in front of you  there is a closed green lockablebox 
Act : move forward
Obs : 2 steps in front of you  there is a closed green lockablebox 
Act : move forward
Obs : Right in front of you  there is a closed green lockablebox 
Act : toggle
Obs : Right in front of you  there is a red apple 
Act : toggle
Obs : Right in front of you  there is a yellow apple 
Success!
New episode.
Obs : 3 steps in front of you and 1 steps to the left there is a closed brown lockablebox 
Act : move forward
Obs : 2 steps in front of you and 1 steps to the left there is a closed brown lockablebox 
Act : move forward
Obs : 1 steps in front of you and 1 steps to the left there is a closed brown lockablebox 
Act : move forward
Obs : Just to the left of you there is a closed brown lockablebox 
Act : turn left
Obs : Right in front of you  there is a closed brown lockablebox 
Act : toggle
Obs : Right in front of you  there is a red apple 
Act : toggle
Obs : Right in front of you  there is a yellow apple 
Success!
\end{lstlisting}
\end{minipage}
\hfill\vline\hfill
\begin{minipage}[t]{0.48\textwidth}
\begin{lstlisting}
New episode.
Obs : 2 steps in front of you and 3 steps to the right there is a closed blue lockablebox 
Act : move forward
Obs : 1 steps in front of you and 3 steps to the right there is a closed blue lockablebox 
Act : move forward
Obs :  3 steps to the right there is a closed blue lockablebox 
Act : turn right
Obs : 3 steps in front of you  there is a closed blue lockablebox 
Act : move forward
Obs : 2 steps in front of you  there is a closed blue lockablebox 
Act : move forward
Obs : Right in front of you  there is a closed blue lockablebox 
Act : toggle
Obs : Right in front of you  there is a red apple 
Act : toggle
Obs : Right in front of you  there is a yellow apple 
Success!
New episode.
Obs : 
Act : turn right
Obs : 
Act : turn right
Obs : 1 steps in front of you and 1 steps to the right there is a closed brown lockablebox 
Act : move forward
Obs : Just to the right of you there is a closed brown lockablebox 
Act : turn right
Obs : Right in front of you  there is a closed brown lockablebox 
Act : toggle
Obs : Right in front of you  there is a red apple 
Act : toggle
Obs : Right in front of you  there is a yellow apple 
Success!
\end{lstlisting}
\end{minipage}
\caption{\footnotesize In context examples for the ASocialBox environment.}
\label{fig:in_context_asocialapple}
\end{figure}

\begin{figure}
\centering
\begin{minipage}[t]{0.48\textwidth}
\begin{lstlisting}
New episode.
Obs : 1 steps in front of you and 2 steps to the left there is a closed brown lockablebox 
Caretaker says:  blue 
Act : turn left
Obs : 3 steps in front of you and 3 steps to the left there is a caretaker
1 steps in front of you and 1 steps to the left there is a closed blue lockablebox 
2 steps in front of you and 1 steps to the right there is a closed brown lockablebox 
Caretaker says:  blue 
Act : move forward
Obs : 2 steps in front of you and 3 steps to the left there is a caretaker
Just to the left of you there is a closed blue lockablebox 
1 steps in front of you and 1 steps to the right there is a closed brown lockablebox 
Caretaker says:  blue 
Act : turn left
Obs : Right in front of you  there is a closed blue lockablebox 
3 steps in front of you and 2 steps to the right there is a caretaker
Caretaker says:  blue 
Act : toggle
Obs : Right in front of you  there is a red apple 
3 steps in front of you and 2 steps to the right there is a caretaker
Caretaker says:  blue 
Act : toggle
Obs : Right in front of you  there is a yellow apple 
3 steps in front of you and 2 steps to the right there is a caretaker
Caretaker says:  blue 
Success!
New episode.
Obs : 3 steps in front of you  there is a caretaker
Act : turn left
Obs :  2 steps to the left there is a closed blue lockablebox 
5 steps in front of you  there is a closed green lockablebox 
 3 steps to the right there is a caretaker
Caretaker says:  blue 
Act : turn left
Obs : 2 steps in front of you  there is a closed blue lockablebox 
Caretaker says:  blue 
Act : move forward
Obs : Right in front of you  there is a closed blue lockablebox 
Caretaker says:  blue 
Act : toggle
Obs : Right in front of you  there is a red apple 
Caretaker says:  blue 
Act : toggle
Obs : Right in front of you  there is a yellow apple 
Caretaker says:  blue 
Success!
New episode.
Obs : 2 steps in front of you and 2 steps to the left there is a closed blue lockablebox 
Caretaker says:  brown 
Act : turn left
Obs : 3 steps in front of you and 1 steps to the left there is a caretaker
2 steps in front of you and 2 steps to the right there is a closed blue lockablebox 
Caretaker says:  brown 
Act : turn left
Obs : 4 steps in front of you  there is a closed brown lockablebox 
1 steps in front of you and 3 steps to the right there is a caretaker
Caretaker says:  brown 
Act : move forward
Obs : 3 steps in front of you  there is a closed brown lockablebox 
 3 steps to the right there is a caretaker
Caretaker says:  brown 
Act : move forward
Obs : 2 steps in front of you  there is a closed brown lockablebox 
Caretaker says:  brown 
Act : move forward
Obs : Right in front of you  there is a closed brown lockablebox 
Caretaker says:  brown 
Act : toggle
\end{lstlisting}
\end{minipage}
\hfill\vline\hfill
\begin{minipage}[t]{0.48\textwidth}
\begin{lstlisting}
Obs : Right in front of you  there is a red apple 
Caretaker says:  brown 
Act : toggle
Obs : Right in front of you  there is a yellow apple 
Caretaker says:  brown 
Success!
New episode.
Obs : Just to the right of you there is a closed green lockablebox 
Caretaker says:  brown 
Act : turn left
Obs : 4 steps in front of you and 3 steps to the right there is a closed brown lockablebox 
Caretaker says:  brown 
Act : move forward
Obs : 3 steps in front of you and 3 steps to the right there is a closed brown lockablebox 
Caretaker says:  brown 
Act : move forward
Obs : 2 steps in front of you and 3 steps to the right there is a closed brown lockablebox 
Caretaker says:  brown 
Act : move forward
Obs : 1 steps in front of you and 3 steps to the right there is a closed brown lockablebox 
Caretaker says:  brown 
Act : move forward
Obs :  3 steps to the right there is a closed brown lockablebox 
Caretaker says:  brown 
Act : turn right
Obs : 5 steps in front of you  there is a caretaker
3 steps in front of you  there is a closed brown lockablebox 
Caretaker says:  brown 
Act : move forward
Obs : 4 steps in front of you  there is a caretaker
2 steps in front of you  there is a closed brown lockablebox 
Caretaker says:  brown 
Act : move forward
Obs : 3 steps in front of you  there is a caretaker
Right in front of you  there is a closed brown lockablebox 
Caretaker says:  brown 
Act : toggle
Obs : 3 steps in front of you  there is a caretaker
Right in front of you  there is a red apple 
Caretaker says:  brown 
Act : toggle
Obs : 3 steps in front of you  there is a caretaker
Right in front of you  there is a yellow apple 
Caretaker says:  brown 
Success!
New episode.
Obs : 
Caretaker says:  blue 
Act : turn left
Obs :  3 steps to the left there is a caretaker
Caretaker says:  blue 
Act : turn left
Obs : 1 steps in front of you and 1 steps to the left there is a closed blue lockablebox 
3 steps in front of you  there is a caretaker
4 steps in front of you and 2 steps to the right there is a closed brown lockablebox 
Caretaker says:  blue 
Act : move forward
Obs : Just to the left of you there is a closed blue lockablebox 
2 steps in front of you  there is a caretaker
3 steps in front of you and 2 steps to the right there is a closed brown lockablebox 
Caretaker says:  blue 
Act : turn left
Obs : Right in front of you  there is a closed blue lockablebox 
 2 steps to the right there is a caretaker
Caretaker says:  blue 
Act : toggle
Obs : Right in front of you  there is a red apple 
 2 steps to the right there is a caretaker
Caretaker says:  blue 
Act : toggle
Obs : Right in front of you  there is a yellow apple 
 2 steps to the right there is a caretaker
Caretaker says:  blue 
Success!
\end{lstlisting}
\end{minipage}
\caption{\footnotesize  In context examples for the ColorBoxes environment.}
\label{fig:in_context_colorboxes}
\end{figure}

\begin{figure}
\centering
\begin{minipage}[t]{0.48\textwidth}
\begin{lstlisting}
New episode.
Obs : 2 steps in front of you and 1 steps to the left there is a unactivated green lever 
2 steps in front of you and 1 steps to the right there is a unactivated blue lever 
Caretaker says:  green 
Act : move forward
Obs : 1 steps in front of you and 1 steps to the left there is a unactivated green lever 
1 steps in front of you and 1 steps to the right there is a unactivated blue lever 
Caretaker says:  green 
Act : move forward
Obs : Just to the left of you there is a unactivated green lever 
Just to the right of you there is a unactivated blue lever 
Caretaker says:  green 
Act : turn left
Obs : Right in front of you  there is a unactivated green lever 
Caretaker says:  green 
Act : toggle
Obs : Right in front of you  there is a activated green lever 
Caretaker says:  green 
Act : turn left
Obs : 6 steps in front of you and 2 steps to the left there is a open green remotedoor 
Just to the left of you there is a unactivated blue lever 
Just to the right of you there is a activated green lever 
5 steps in front of you and 2 steps to the right there is a caretaker
Caretaker says:  green 
Act : move forward
Obs : 5 steps in front of you and 2 steps to the left there is a open green remotedoor 
4 steps in front of you and 2 steps to the right there is a caretaker
Caretaker says:  green 
Act : turn left
Obs : 1 steps in front of you and 1 steps to the left there is a unactivated blue lever 
Caretaker says:  green 
Act : move forward
Obs : Just to the left of you there is a unactivated blue lever 
Caretaker says:  green 
Act : move forward
Obs : 
Caretaker says:  green 
Act : turn right
Obs : 6 steps in front of you  there is a red apple 
5 steps in front of you  there is a open green remotedoor 
Caretaker says:  green 
Act : move forward
Obs : 5 steps in front of you  there is a red apple 
4 steps in front of you  there is a open green remotedoor 
Caretaker says:  green 
Act : move forward
Obs : 4 steps in front of you  there is a red apple 
3 steps in front of you  there is a open green remotedoor 
Caretaker says:  green 
Act : move forward
Obs : 3 steps in front of you  there is a red apple 
2 steps in front of you  there is a open green remotedoor 
Caretaker says:  green 
Act : move forward
Obs : 2 steps in front of you  there is a red apple 
Right in front of you  there is a open green remotedoor 
Caretaker says:  green 
Act : move forward
Obs : Right in front of you  there is a red apple 
Caretaker says:  green 
Act : toggle
Obs : Right in front of you  there is a yellow apple 
Caretaker says:  green 
Success!
New episode.
Obs : 2 steps in front of you and 1 steps to the right there is a closed green lockablebox 
Act : move forward
Obs : 1 steps in front of you and 1 steps to the right there is a closed green lockablebox 
Act : move forward
Obs : Just to the right of you there is a closed green lockablebox 
Act : turn right
\end{lstlisting}
\end{minipage}
\hfill\vline\hfill
\begin{minipage}[t]{0.48\textwidth}
\begin{lstlisting}
Obs : Right in front of you  there is a closed green lockablebox 
Act : toggle
Obs : Right in front of you  there is a red apple 
...
...
...
...
...
...
Obs : 2 steps in front of you  there is a red apple 
2 steps in front of you and 1 steps to the right there is a round green applegenerator 
Just to the right of you there is a green marble 
 2 steps to the right there is a round brown applegenerator 
Caretaker says:  brown 
Act : move forward
Obs : Right in front of you  there is a red apple 
1 steps in front of you and 1 steps to the right there is a round green applegenerator 
Caretaker says:  brown 
Act : toggle
Obs : Right in front of you  there is a yellow apple 
1 steps in front of you and 1 steps to the right there is a round green applegenerator 
Caretaker says:  brown 
Success!
New episode.
Obs : 2 steps in front of you and 2 steps to the left there is a locked green lockablebox 
 2 steps to the left there is a green switch 
5 steps in front of you and 3 steps to the right there is a caretaker
Caretaker says:  green 
Act : turn left
Obs : 1 steps in front of you and 1 steps to the left there is a blue switch 
2 steps in front of you  there is a green switch 
2 steps in front of you and 2 steps to the right there is a locked green lockablebox 
Caretaker says:  green 
Act : move forward
Obs : Just to the left of you there is a blue switch 
Right in front of you  there is a green switch 
1 steps in front of you and 2 steps to the right there is a locked green lockablebox 
Caretaker says:  green 
Act : toggle
Obs : Just to the left of you there is a blue switch 
Right in front of you  there is a green switch 
1 steps in front of you and 2 steps to the right there is a closed green lockablebox 
Caretaker says:  green 
Act : turn right
Obs : 2 steps in front of you and 1 steps to the left there is a closed green lockablebox 
Just to the left of you there is a green switch 
Caretaker says:  green 
Act : move forward
Obs : 1 steps in front of you and 1 steps to the left there is a closed green lockablebox 
Caretaker says:  green 
Act : move forward
Obs : Just to the left of you there is a closed green lockablebox 
Caretaker says:  green 
Act : turn left
Obs :  3 steps to the left there is a blue switch 
1 steps in front of you and 2 steps to the left there is a green switch 
Right in front of you  there is a closed green lockablebox 
Caretaker says:  green 
Act : toggle
Obs :  3 steps to the left there is a blue switch 
1 steps in front of you and 2 steps to the left there is a green switch 
Right in front of you  there is a red apple 
Caretaker says:  green 
Act : toggle
Obs :  3 steps to the left there is a blue switch 
1 steps in front of you and 2 steps to the left there is a green switch 
Right in front of you  there is a yellow apple 
Caretaker says:  green 
Success!
\end{lstlisting}
\end{minipage}
\caption{\footnotesize In context examples for the ColorBoxes environment when tested for generalization.}
\label{fig:in_context_colorboxes_gen}
\end{figure}

\begin{figure*}[htb!]
\centering
\subfloat[\footnotesize Asocial Apple]{
\includegraphics[height=0.25\textheight]{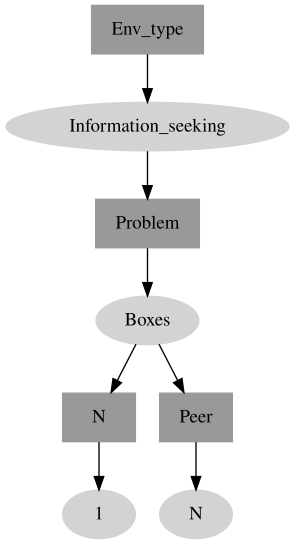}
\label{fig:llms_tree_train}
}
\subfloat[\footnotesize Color boxes]{
\includegraphics[height=0.25\textheight]{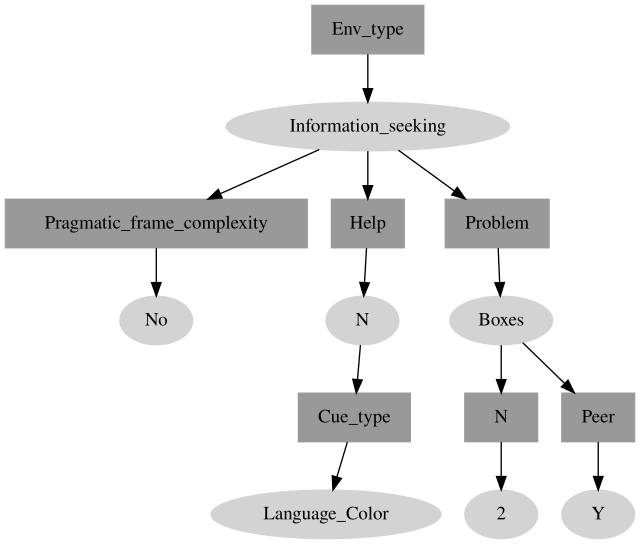}
\label{fig:llms_tree_test}
}
\caption{\footnotesize Sampling trees used for evaluation in the experiments with LLM-based interactive agents (section \ref{sec:exp_llm})}
\label{fig:llms_tree}
\end{figure*}

\vskip 0.2in
\clearpage
\bibliography{sample}
\bibliographystyle{theapa}

\end{document}